\def\year{2017}\relax
\documentclass[letterpaper]{article}
\usepackage{aaai17}
\usepackage{times}
\usepackage{helvet}
\usepackage{courier}
\usepackage[named]{algo}

\usepackage{epsfig}
\usepackage{amssymb}
\usepackage{amsmath}
\usepackage{amsfonts}
\usepackage{ifthen}

\usepackage{graphicx}
\usepackage{caption}
\usepackage{subcaption}
\usepackage{epstopdf}
\usepackage{wrapfig}
\usepackage{bm}
\usepackage{booktabs}       % professional-quality tables
\usepackage{nicefrac}       % compact symbols for 1/2, etc.
\usepackage{array}			% array
\usepackage{url}

\newtheorem{theorem}{Theorem}
\newtheorem{proposition}{Proposition}
\newtheorem{lemma}{Lemma}

\newtheorem{definition}{Definition}

\newcommand{\squishlist}{
 \begin{list}{$\bullet$}
  { \setlength{\itemsep}{5pt}
     \setlength{\parsep}{0pt}
     \setlength{\topsep}{0pt}
     \setlength{\partopsep}{0pt}
     \setlength{\leftmargin}{0.7em}
     \setlength{\labelwidth}{0.5em}
     \setlength{\labelsep}{0.2em} } }

\newcommand{\squishlisttwo}{
 \begin{list}{$\bullet$}
  { \setlength{\itemsep}{2pt}
     \setlength{\parsep}{0pt}
    \setlength{\topsep}{2pt}
    \setlength{\partopsep}{0pt}
    \setlength{\leftmargin}{1em}
    \setlength{\labelwidth}{1.5em}
    \setlength{\labelsep}{0.5em} } }

\newcommand{\squishend}{
  \end{list}  }

\def\myproof{1} 
\begin{document}
\title{A Generalized Stochastic Variational Bayesian Hyperparameter Learning Framework for Sparse Spectrum Gaussian Process Regression}
%\title{Variational Bayesian Gaussian Processes for Efficient Learning with Big Data}
\author{
Quang Minh Hoang$^{\dag}$ \and Trong Nghia Hoang$^{\ast}$ \and Kian Hsiang Low$^{\dag}$\\
Department of Computer Science$^{\dag}$, Interactive Digital Media Institute$^{\ast}$\\ 
National University of Singapore, Republic of Singapore\\
\{hqminh, lowkh\}@comp.nus.edu.sg$^{\dag}$, nghiaht@nus.edu.sg$^{\ast}$ 
%Paper ID: \#1395\\
%Keywords: Gaussian process, stochastic variational inference, scalability\\
%Topics: ML: Bayesian Learning, ML: Big Data / Scalability, ML: Kernel Methods
%Quang Minh Hoang, Trong Nghia Hoang \and Kian Hsiang Low\\
%Department of Computer Science, National University of Singapore, Republic of Singapore\\
%\{nghiaht,hqminh,lowkh\}@comp.nus.edu.sg
}

\maketitle
\begin{abstract}\vspace{-1mm}
%This paper considers a less 
While much research effort has been dedicated to scaling up sparse \emph{Gaussian process} (GP) models based on inducing variables for big data, little attention is afforded to the other less explored class of low-rank GP approximations that exploit the sparse spectral representation of a GP kernel.
This paper presents such an effort to advance the state of the art of sparse spectrum GP models to achieve competitive predictive performance for massive datasets. Our  generalized framework of \emph{stochastic variational Bayesian sparse spectrum GP} ($s$VBSSGP) models 
addresses their shortcomings by adopting a Bayesian treatment of the spectral frequencies to avoid overfitting, modeling these frequencies jointly in its variational distribution to enable their interaction \emph{a posteriori}, and exploiting local data for boosting the predictive performance. However, such structural improvements result in a variational lower bound that is intractable to be optimized. To resolve this, we exploit a variational parameterization trick to make it amenable to stochastic optimization.
Interestingly, the resulting stochastic gradient has a linearly decomposable structure that can be exploited to refine our stochastic optimization method to incur constant time per iteration while preserving its property of being an unbiased estimator of the exact gradient of the variational lower bound. Empirical evaluation on real-world datasets shows that $s$VBSSGP outperforms state-of-the-art stochastic implementations of sparse GP models.\vspace{-2.3mm}
%Inference with \emph{Gaussian processes} (GP) has been radically scaled up to massive, million-sized datasets in the past few years. Learning its hyperparameters on the same scale of data, however, has not been as successful. In fact, most existing GP models suffer a cubic complexity when it comes to hyperparameter learning. Even among the few that scale well with large data, hyperparameters are often learned via an iterative point-based estimation procedure which fails to account for the estimation uncertainty and consequently compromises their predictive performances. To tackle this challenge, this paper exploits the sparse spectral representation of GP to devise a novel Variational GP framework with a rigorous Bayesian treatment for hyperparameter learning whose processing cost is independent of the data size. This entirely mitigates both GP's existing computational bottleneck and its inability to embrace hyperparameter uncertainty. Our empirical results show that the proposed framework also outperforms state-of-the-art GP models on several real-world datasets, one of which contains millions of data points.
\end{abstract}

\section{Introduction}
\label{intro}%\vspace{-0.5mm}
The machine learning community has recently witnessed the \emph{Gaussian process} (GP) models gaining considerable traction in the research on kernel methods due to its expressive power and capability of performing probabilistic non-linear regression.
%captured a rapidly growing interest among the machine learning (ML) community as one of the most effective probabilistic ML models. 
However, a full-rank GP regression model 
%\cite{Rasmussen06} 
incurs cubic time in the size of the data to compute its predictive distribution, hence limiting its use to small datasets. 
To lift this computational curse, a vast literature of sparse GP regression models \cite{Candela05,Titsias09} have exploited a structural assumption of conditional independence  based on the notion of \emph{inducing variables} for achieving linear time in the data size. 
%Tresp02,Seeger03,Smola01,Snelson06,Snelson07a,Titsias13
%To scale up to massive datasets of millions in size, 
To scale up these sparse GP regression models further for performing real-time predictions necessary in many time-critical applications and decision support systems (e.g., environmental sensing \cite{LowAAMAS13,LowSPIE09,LowAAAI16b,LowAAMAS08,LowICAPS09,LowAAMAS11,LowAAMAS12,LowAeroconf10,LowAAAI16a}, traffic monitoring \cite{LowUAI12,LowRSS13,TASE15,LowECML14b,NghiaICML14,LowDyDESS14,LowECML14a,LowAAMAS14,LowAAAI14,LowIAT12}), 
(a) distributed \cite{Chen13,HoangICML16,LowAAAI15} and (b) stochastic  \cite{Hensman13,NghiaICML15} implementations of such models have been developed to, respectively, (a) reduce their time to train with all the data by a factor close to the number of machines and (b) train with a small, randomly sampled subset of data in constant time per iteration of stochastic gradient ascent update and achieve asymptotic convergence to their predictive distributions.
%Yarin14,there has been a vast literature of inducing point methods , which exploits the notion of a small set of inducing outputs to construct a sparse approximation of the predictive distribution in linear time. When parallel computing resources are abundant, these works can be further coupled with a distributed framework \cite{Chen13,Yarin14,LowAAAI15,HoangICML16} to achieve even better scalability on big data. In contrast, when such resources are not available, a more affordable alternative involves iteratively training the GP model with randomly sampled small subsets of data \cite{Hensman13}, which has subsequently been generalized into a unifying framework of anytime GPs \cite{NghiaICML15}. The most compelling factor of this anytime paradigm revolves around its ability to produce low-cost inferences (i.e., its processing cost is constant with respect to the data size) with guaranteed convergence to those of the existing GP approximations spanned by the unifying view of \citeauthor{Candela05}~\shortcite{Candela05}. 

On the other hand, there is a less well-explored, alternative class of low-rank GP approximations that exploit sparsity in the spectral representation of a GP kernel \cite{Yarin15,Miguel10} for gaining time efficiency and have empirically demonstrated competitive predictive performance for datasets of up to tens of thousands in size but, surprisingly, not received as much attention and research effort.
In contrast to the above literature, such sparse spectrum GP regression models do not need to introduce an additional set of inducing inputs which is computationally challenging to be jointly optimized, especially with a large number of them that is necessary for accurate predictions.
Unfortunately, the \emph{sparse spectrum GP} (SSGP) model of~\citeauthor{Miguel10}~\shortcite{Miguel10} does not scale well to massive datasets due to its linear time in the data size and also finds a point estimate of the spectral frequencies of its kernel 
%(i.e., equivalent to learning the kernel hyperparameters (Section~\ref{hgp})) 
that risks overfitting.
The recent \emph{variational SSGP} (VSSGP) model of \citeauthor{Yarin15}~\shortcite{Yarin15} has attempted to address both shortcomings of SSGP with a stochastic implementation and a Bayesian treatment of the spectral frequencies, respectively. However, VSSGP has imposed the following highly restrictive structural assumptions to facilitate analytical derivations but potentially compromise its predictive performance:
%undermine, detrimental
% More recently, the work of \citeauthor{Yarin15}~\shortcite{Yarin15} has further adopted a Bayesian treatment of these spectral frequencies, hence also resolving the first issue of estimation uncertainty. Their proposed anytime GP model (named VSSGP), however, still suffers from the following non-trivial shortcomings: 
(a) The spectral frequencies are assumed to be fully independent \emph{a posteriori} in its variational distribution, and (b) every test output assumes a deterministic relationship with the spectral frequencies in its test conditional and is thus conditionally independent of the training data, including the local data 
%\footnote{The use of local data has been empirically shown to be very effective in improving the prediction quality of sparse GP models utilizing inducing variables \cite{Chen13,NghiaICML15,HoangICML16,LowAAAI15}.} 
``close'' to it (in the correlation sense).
%, thus failing to utilize local information Snelson07a
%the latter of which is empirically very effective in  .
As such, open questions remain whether VSSGP can still perform competitively well with these highly restrictive structural assumptions for massive, million-sized datasets 
%with multiple input dimensions\footnote{In the work of \citeauthor{Yarin15}~\shortcite{Yarin15}, VSSGP is empirically evaluated on datasets of only $1$ input dimension.} 
and, more importantly, whether these assumptions can be relaxed to improve the predictive performance while preserving scalability to big data.
 %
% VSSGP operate on small datasets and 1D input dimension, don't know how effective it is in terms of predictive performance on massive datasets. With these limitations,  it is not known whether it can still perform competitively well, which is the focus of our work in this paper. 
%
% Little effort has been made to scale up to massive datasets and
%
%Despite the apparent speedup over FGPR inference on big data, the above literature of inducing-point anytime GPs has not satisfactorily addressed the issue of hyperparameter learning. Among the aforementioned works, hyperparameters are often assumed to be known in advance \cite{NghiaICML15} or estimated using point-based approaches \cite{Hensman13} which fail to convey the estimation uncertainty. Furthermore, these models also incur an extra set of variables to be optimised by introducing a set of inducing inputs, thus shifting the computational bottleneck from inference to hyperparameter learning. The latter can be resolved by adopting a less well-explored alternative, which exploits the spectral representation of the GP kernel \cite{Miguel10} to construct a sparse approximation of the predictive distribution without having to introduce (hence, optimize) an extra set of inducing inputs. In this paradigm, learning the hyperparameters is equivalently cast as optimising the spectral frequencies. 
%
%This risks overfitting, especially when the number of hyperparameters is all but small as previously criticised by \citeauthor{Titsias13}~\shortcite{Titsias13}. 

To tackle these questions, this paper presents a novel generalized framework of  \emph{stochastic variational Bayesian sparse spectrum GP} ($s$VBSSGP) regression models which (a) enables the spectral frequencies to interact \emph{a posteriori} by modeling them jointly in its variational distribution, and (b) spans a spectrum of test conditionals that can trade off between the contributions of the degenerate test conditional of VSSGP vs. the \emph{local} SSGP model trained with local data (Section~\ref{hgp}). 
However, such proposed structural improvements over VSSGP and SSGP to boost the predictive performance result in a variational lower bound that is intractable to be optimized.
%come at a cost of an intractable optimization of the .
To overcome this computational difficulty, we exploit a variational parameterization trick to make the variational lower bound amenable to stochastic optimization, which still incurs linear time in the data size per iteration of stochastic gradient ascent update (Section~\ref{vbgpr}).
% in linear time per iteration. 
Interestingly, we can exploit the linearly decomposable structure of this stochastic gradient to refine our stochastic optimization method to incur only constant time per iteration while preserving its property of being an unbiased estimator of the exact gradient of the variational lower bound (Section~\ref{anytime}). 
%
%Though our generalized framework relaxes the highly restrictive structural assumptions of VSSGP, the variational lower bound remains amenable to stochastic optimization by jointly exploiting  and of its resulting stochastic gradient incurring linear time to derive a new stochastic gradient incurring 
%
% which generalises VSSGP \cite{Yarin15} to facilitate anytime Bayesian learning of the hyperparameters while incidentally lifting its restrictive assumptions of VSSGP.
% 
% This is achieved by generalising the concept of \emph{nuisance} random variables first introduced in VSSGP (Proposition~\ref{lem:l1})  to exploit varying amounts of local information to improve the prediction quality of our framework (Section 2). 
%
%In addition, our generalized concept also leads to a decomposable structure (Section 3) that can be embedded within the log likelihood of data which allows the hyperparameters to be learned in an anytime (Bayesian) fashion (thus preserving scalability) without having to impose the restrictive assumptions of VSSGP (Section 4).
We empirically evaluate the predictive performance and time efficiency of $s$VBSSGP on three real-world datasets, one of which is millions in size (Section~\ref{experiment}).\vspace{-0.3mm} 
%Our experiments expectedly show that VBGPR not only preserves scalability but also produces better predictive performance than the state-of-the-art anytime GPss.
%
%Our work here generalizes SSGP to a variational Bayesian formulation as well as VSSGP by addressing its limitations of ...
%
%This is achieved by introducing an auxiliary set of random variables which, together with the hyperparameters, naturally embed a decomposable structure within the log likelihood of data, that underlines the core of our proposed anytime algorithm. Finally, we empirically evaluate the predictive performance and scalability of VBGPR on two large real-world datasets (Section 5). The experiment results expectedly show that VBGPR not only preserves scalability but also produces better predictive performance than the existing state-of-the-art anytime GPs.
%
\section{A Generalized Bayesian Sparse Spectrum Gaussian Process Regression Framework}
\label{hgp}%\vspace{-0mm} 
Let $\mathcal{X}$ be a
%The original \emph{full-rank Gaussian process} (FGPR) model is defined on a 
$d$-dimensional input space such that each input vector $\mathbf{x} \in \mathcal{X}$ is associated with a latent output $f_\mathbf{x}$ and a noisy output $y_\mathbf{x} \triangleq f_\mathbf{x} + \epsilon$ generated by perturbing $f_\mathbf{x}$ with a random noise $\epsilon\sim \mathcal{N}(0, \sigma_n^2)$ where $\sigma_n^2$ is the noise variance.
% to the latent output $f_\mathbf{x}$. 
Let $\{f_\mathbf{x}\}_{\mathbf{x} \in \mathcal{X}}$ denote a \emph{Gaussian process} (GP), that is, every finite subset of $\{f_\mathbf{x}\}_{\mathbf{x} \in \mathcal{X}}$ follows a multivariate Gaussian distribution.
Then, the GP is fully specified by its prior mean $\mathbb{E}[f_\mathbf{x}]$ (i.e., assumed to be $0$ for notational simplicity) and covariance $k(\mathbf{x},\mathbf{x}')\triangleq\text{cov}[f_\mathbf{x},f_{\mathbf{x}'}]$ for all $\mathbf{x}, \mathbf{x}'\in\mathcal{X}$, the latter of which can be defined by the commonly-used squared exponential kernel $k(\mathbf{x},\mathbf{x}') \triangleq \sigma^2_s\mathrm{exp}(-0.5(\mathbf{x}-\mathbf{x'})^\top\boldsymbol{\Delta}^{-1}(\mathbf{x}-\mathbf{x'}))$ 
where $\boldsymbol{\Delta} \triangleq \mathrm{diag}[\ell_1^2, \ell_2^2, \ldots, \ell_d^2]$ and $\sigma^2_s$ are its squared length-scale and signal variance hyperparameters, respectively. Such a kernel can be expressed as the Fourier transform of a density function $p(\mathbf{r})$ over the domain of frequency vector $\mathbf{r}$ whose coefficients form a set of trigonometric basis functions \cite{Miguel10}:%\vspace{-0.5mm}
\begin{equation}
\label{a01}
k(\mathbf{x},\mathbf{x}') = \mathbb{E}_{\mathbf{r} \sim p(\mathbf{r})}[\ \sigma^2_s \mathrm{cos}(2\pi \mathbf{r}^\top(\mathbf{x} - \mathbf{x}'))\ ]%\vspace{-0.5mm}
\end{equation}
where $p(\mathbf{r}) \triangleq \mathcal{N}\hspace{-1mm}\left(\mathbf{0}, (4\pi^2\boldsymbol{\Delta})^{-1}\right)$. In the same spirit as that of \citeauthor{Miguel10}~\shortcite{Miguel10}, we approximate the kernel in~\eqref{a01} by its unbiased estimator constructed from $m$ i.i.d. sampled spectral frequencies $\mathbf{r}_i$ for $i=1,\ldots,m$:\vspace{-0.5mm}
\begin{equation}
\hspace{-1.7mm}
\begin{array}{l}\displaystyle
k(\mathbf{x},\mathbf{x}')\hspace{-0.5mm}\simeq\hspace{-0.5mm} \frac{\sigma^2_s}{m} \sum^m_{i=1}\mathrm{cos}(2\pi \mathbf{r}^\top_i(\mathbf{x} \hspace{-0.5mm}- \hspace{-0.5mm}\mathbf{x}'))\hspace{-0.5mm} =\hspace{-0.5mm} \boldsymbol{\phi}^\top_{{\boldsymbol{\theta}}}(\mathbf{x}) \boldsymbol{\Lambda} \boldsymbol{\phi}_{{\boldsymbol{\theta}}}(\mathbf{x'})\hspace{-4.4mm}\vspace{-0mm}
\end{array}
\label{eq:kernel}\vspace{-1mm}
\end{equation}
where $\boldsymbol{\phi}_{\boldsymbol{\theta}}(\mathbf{x}) \triangleq [\phi^1_{\boldsymbol{\theta}}(\mathbf{x}), \phi^2_{\boldsymbol{\theta}}(\mathbf{x}), \ldots, \phi^{2m}_{\boldsymbol{\theta}}(\mathbf{x})]^\top$ denotes a vector of basis functions  
%(i.e., of input $\mathbf{x}$) 
$\phi^{2i-1}_{\boldsymbol{\theta}}(\mathbf{x}) \triangleq \cos(2\pi\mathbf{r}_i^\top\mathbf{x})$ and $\phi^{2i}_{\boldsymbol{\theta}}(\mathbf{x}) \triangleq \sin(2\pi\mathbf{r}_i^\top\mathbf{x})$ for $i=1,\ldots,m$, $\boldsymbol{\Lambda} \triangleq (\sigma^2_s/m)\mathbf{I}$, and ${\boldsymbol{\theta}} \triangleq \mathrm{vec}(\mathbf{r}_1, \mathbf{r}_2, \ldots, \mathbf{r}_m)$. 
Learning the length-scales in $\boldsymbol{\Delta}$ of the original kernel is thus cast as optimizing ${\boldsymbol{\theta}}$ in this alternative representation~\eqref{eq:kernel} of the kernel. 
Then, the induced covariance matrix $\mathbf{K}(\mathbf{X},\mathbf{X})$ for any finite subset $\mathbf{X} \triangleq \{\mathbf{x}_i\}_{i=1}^n$ of training inputs can be written as $\mathbf{K}(\mathbf{X},\mathbf{X}) \triangleq \mathbf{\Phi}_{\boldsymbol{\theta}}^\top(\mathbf{X}) \mathbf{\Lambda} \mathbf{\Phi}_{\boldsymbol{\theta}}(\mathbf{X})$
%\begin{eqnarray}
%\mathbf{K}(\mathbf{X},\mathbf{X}) &=& \mathbf{\Phi}_{\boldsymbol{\theta}}^\top\left(\mathbf{X}\right) \mathbf{\Lambda} \mathbf{\Phi}_{\boldsymbol{\theta}}\left(\mathbf{X}\right) \ , \label{eq:covariance}
%\end{eqnarray}     
where $\mathbf{\Phi}_{\boldsymbol{\theta}}(\mathbf{X}) \triangleq [\boldsymbol{\phi}_{\boldsymbol{\theta}}(\mathbf{x}_1)  \ldots \boldsymbol{\phi}_{\boldsymbol{\theta}}(\mathbf{x}_n)]$. 

To learn the optimal parameters of the distribution over ${\boldsymbol{\theta}}$ (i.e., not known in advance) given by $\mathcal{N}(\mathbf{0}, (4\pi^2\boldsymbol{\Delta}_{\boldsymbol{\theta}})^{-1})$ (i.e., derived from $p(\mathbf{r})$ below~\eqref{a01}) where $\boldsymbol{\Delta}_{\boldsymbol{\theta}} \triangleq \mathbf{I} \otimes \boldsymbol{\Delta}$, we adopt the standard Bayesian treatment for ${\boldsymbol{\theta}}$ by first imposing a prior distribution ${\boldsymbol{\theta}} \sim \mathcal{N}(\mathbf{0}, {\boldsymbol{\Theta}})$ for some covariance ${\boldsymbol{\Theta}}$ designed \emph{a priori} to reflect our knowledge about ${\boldsymbol{\theta}}$ and then using training data to infer its posterior  which is expected to closely approximate the optimal distribution. This signifies a key difference between our generalized framework and the \emph{sparse spectrum GP} (SSGP) model of \citeauthor{Miguel10}~\shortcite{Miguel10}, the latter of which finds a point estimate of ${\boldsymbol{\theta}}$ via maximum likelihood estimation that risks overfitting. 

As shall be elucidated later in this paper,  
%later in Sections~\ref{vbgpr} and~\ref{anytime}, 
the finite trigonometric representation of the kernel~\eqref{eq:kernel} can be used to efficiently and scalably compute the predictive distribution of our framework by exploiting some mild structural assumptions. Specifically, we assume that for any finite subset $\mathbf{X} \subset \mathcal{X}$, a vector $\mathbf{s}$ of \emph{nuisance} variables exists for which the joint distribution of $\mathbf{f} \triangleq [f_\mathbf{x}]^{\top}_{\mathbf{x}\in\mathbf{X}}$ and $\mathbf{s}$ conditioned on ${\boldsymbol{\theta}}$ is%\vspace{-1.6mm}
\begin{equation}
\hspace{-0.7mm}
\left[\hspace{-1mm}
\begin{array}{l}
\mathbf{s} \\
\mathbf{f}
\end{array}\hspace{-1mm}
\right]
\hspace{0.25mm}\sim\hspace{0.25mm}
\mathcal{N}\hspace{-0.7mm}\left(
\left[
\hspace{-1mm}\begin{array}{l}
\mathbf{0}\\
\mathbf{0}
\end{array}\hspace{-1mm}
\right],
\left[\hspace{-1mm}
\begin{array}{cc}
\boldsymbol{\Lambda} & \boldsymbol{\Lambda}\boldsymbol{\Phi}_{\boldsymbol{\theta}}(\mathbf{X}) \\
\boldsymbol{\Phi}^\top_{\boldsymbol{\theta}}(\mathbf{X})\boldsymbol{\Lambda} & \boldsymbol{\Phi}^\top_{\boldsymbol{\theta}}(\mathbf{X})\boldsymbol{\Lambda}\boldsymbol{\Phi}_{\boldsymbol{\theta}}(\mathbf{X}) 
\end{array}\hspace{-1mm}
\right]
\right).\hspace{-1.4mm}
\label{eq:imagine}%\vspace{-1mm} 
\end{equation}
Intuitively, $\mathbf{s}$ can be interpreted as the latent outputs of some imaginary inputs $\mathbf{U} \triangleq \{\mathbf{u}_j\}_{j=1}^{2m}$ such that $\phi^i_{\boldsymbol{\theta}}(\mathbf{u}_j)\triangleq \mathbb{I}(i = j)$. Using~\eqref{eq:kernel}, it follows immediately that $\mathbf{K}(\mathbf{U}, \mathbf{U}) = \boldsymbol{\Lambda}$, $\mathbf{K}(\mathbf{U}, \mathbf{X}) = \boldsymbol{\Lambda} \boldsymbol{\Phi}_{\boldsymbol{\theta}}(\mathbf{X})$, and $\mathbf{K}(\mathbf{X}, \mathbf{U}) = \boldsymbol{\Phi}^\top_{\boldsymbol{\theta}}(\mathbf{X}) \boldsymbol{\Lambda}$ which reproduce the covariance matrix in~\eqref{eq:imagine}. Secondly, we assume that given $\boldsymbol{\alpha} \triangleq \mathrm{vec}({\boldsymbol{\theta}},\mathbf{s})$, any latent test output depends on only a small subset of local training data of fixed size: Supposing the input space is partitioned into $p$ disjoint sub-spaces (i.e., $\mathcal{X} = \bigcup_{i=1}^p \mathcal{X}_i$) which directly induce a partition on the training inputs $\mathbf{X} = \bigcup_{i=1}^p \mathbf{X}_i$ such that $\mathbf{X}_i \subset \mathcal{X}_i$, $p(f_{\mathbf{x}_\ast} | \mathbf{y}, \boldsymbol{\alpha}) = p(f_{\mathbf{x}_\ast} | \mathbf{y}_k, \boldsymbol{\alpha})$ for any test input $\mathbf{x}_\ast \in \mathcal{X}_k$ and $k=1,\ldots,p$ where $\mathbf{y} \triangleq [y_\mathbf{x}]^{\top}_{\mathbf{x}\in\mathbf{X}}$ and $\mathbf{y}_k \triangleq[y_{\mathbf{x}}]^{\top}_{\mathbf{x}\in\mathbf{X}_k}$. Then, the predictive distribution can be computed using%\vspace{-0.5mm} 
\begin{equation}
p(f_{\mathbf{x}_*} | \mathbf{y}) = \mathbb{E}_{\boldsymbol{\alpha}\sim p(\boldsymbol{\alpha}|\mathbf{y})}[\ p(f_{\mathbf{x}_*} | \mathbf{y}_k, \boldsymbol{\alpha})\ ] 
\label{eq:decompose}%\vspace{-0.5mm}
%\int_{\boldsymbol{\alpha}} p(f_{\mathbf{x}_*} | \mathbf{y}_k, \boldsymbol{\alpha})\ p(\boldsymbol{\alpha} | \mathbf{y})\ \mathrm{d}\boldsymbol{\alpha}
\end{equation}
 %\eqref{eq:decompose} 
which reveals that it can be evaluated by deriving posterior distribution $p(\boldsymbol{\alpha} | \mathbf{y})$ described later in Section~\ref{vbgpr}, and the test conditional $p(f_{\mathbf{x}_*} | \mathbf{y}_k, \boldsymbol{\alpha})$  \emph{consistent} with the above structural assumption of $\mathbf{s}$: Marginalizing out  nuisance variables $\mathbf{s}$ from such a test conditional should yield $p(f_\mathbf{x_\ast}| \mathbf{y}_k, {\boldsymbol{\theta}})$, i.e., $p(f_\mathbf{x_\ast}| \mathbf{y}_k, {\boldsymbol{\theta}}) = \int_{\mathbf{s}}\ p(f_{\mathbf{x}_*} | \mathbf{y}_k, \boldsymbol{\alpha})\ p(\mathbf{s} | \mathbf{y}_k, {\boldsymbol{\theta}})\ \mathrm{d}\mathbf{s}$ where both $p(f_\mathbf{x_\ast}| \mathbf{y}_k, {\boldsymbol{\theta}})$ and $p(\mathbf{s} | \mathbf{y}_k, {\boldsymbol{\theta}})$ can be derived from~\eqref{eq:imagine}, as shown in\if\myproof1 Appendix~\ref{app:a}\fi\if\myproof0 \cite{MinhAAAI17}\fi.
% Appendix~. 
%To overcome this limitation, 

Our first result below derives a spectrum of consistent test conditionals in our generalized framework that trade off between the use of global information $\boldsymbol{\alpha}$ vs. local data $(\mathbf{X}_k,\mathbf{y}_k)$, albeit to varying degrees controlled by $\gamma$:
\begin{proposition}
\label{lem:l1}	
%Let $\mathbf{s}$ be defined by~\eqref{eq:imagine} above with $p(\mathbf{s}) \triangleq \mathcal{N}(\mathbf{s}| 0, \boldsymbol{\Lambda})$, then 
For all $\mathbf{x_\ast} \in \mathcal{X}_k$ and $|\gamma| \leq 1$, define the test conditional 
$p(f_\mathbf{x_\ast} | \mathbf{y}_k, \boldsymbol{\alpha}) \triangleq \mathcal{N}(\mu_{\mathbf{x}_\ast}\hspace{-0.7mm}(\boldsymbol{\alpha}), \sigma^2_{\mathbf{x}_\ast}\hspace{-0.7mm}(\boldsymbol{\alpha}))$ where
%\vspace{-0.3mm}
$$
\hspace{-1.7mm}
\begin{array}{rcl}
\mu_{\mathbf{x}_\ast}\hspace{-0.7mm}(\boldsymbol{\alpha}) &\hspace{-2.4mm}\triangleq&\hspace{-2.4mm}\displaystyle \gamma\boldsymbol{\phi}^\top_{\boldsymbol{\theta}}(\mathbf{x_\ast})\mathbf{s} + (1 - \gamma)\boldsymbol{\phi}^\top_{\boldsymbol{\theta}}(\mathbf{x_\ast})\boldsymbol{\Gamma}^{-1}_k \boldsymbol{\Phi}_{\boldsymbol{\theta}}(\mathbf{X}_k) \mathbf{y}_k\ ,  \vspace{0.5mm}\\
\sigma^2_{\mathbf{x}_\ast}\hspace{-0.7mm}(\boldsymbol{\alpha}) &\hspace{-2.4mm}\triangleq&\hspace{-2.4mm}\displaystyle (1 - \gamma^2)\sigma^2_n\boldsymbol{\phi}^\top_{\boldsymbol{\theta}}(\mathbf{x_\ast})
 \boldsymbol{\Gamma}^{-1}_k\boldsymbol{\phi}_{\boldsymbol{\theta}}(\mathbf{x_\ast})\ , 
%\vspace{-0.5mm}
\end{array}
$$ 
and  
$\boldsymbol{\Gamma}_k \triangleq\boldsymbol{\Phi}_{\boldsymbol{\theta}}(\mathbf{X}_k)\boldsymbol{\Phi}_{\boldsymbol{\theta}}^\top(\mathbf{X}_k)+\sigma_n^{2}\boldsymbol{\Lambda}^{-1}$.
%$\mathbf{\Gamma}_k \triangleq \left(\mathbf{K}(\mathbf{X}_k,\mathbf{X}_k) + \sigma^2_n\mathbf{I}\right)^{-1}$. 
Then, $p(f_\mathbf{x_\ast} | \mathbf{y}_k, \boldsymbol{\alpha})$ is consistent with the structural assumption of $\mathbf{s}$ in~\eqref{eq:imagine}. %See  of \cite{MinhAAAI17Supp} for a detailed proof.
%with $h(\mathbf{x_\ast}) \triangleq \left(\mathbf{K}(\mathbf{X}_k,\mathbf{X}_k) + \sigma^2_n\mathbf{I}\right)^{-1}\mathbf{y}_k$ and $q(\mathbf{x_\ast}, \mathbf{x_\ast}) \triangleq \mathbf{K}(\mathbf{x_\ast}, \mathbf{X}_k)\left(\mathbf{K}(\mathbf{X}_k,\mathbf{X}_k) + \sigma^2_n\mathbf{I}\right)^{-1}\mathbf{K}(\mathbf{X}_k,\mathbf{x_\ast})$ is consistent with our assumption in Eq.~\eqref{eq:imagine}. See Appendix A for a detailed proof. 
\end{proposition}
Its proof is in\if\myproof1 Appendix~\ref{app:a}\fi\if\myproof0 \cite{MinhAAAI17}\fi.
\vspace{1mm}

\iffalse
{\noindent \bf Remark 1} Note that the condition $|\gamma| \leq 1$ is necessary for $\sigma^2_\ast(\boldsymbol{\alpha})$ to be non-negative. Subsequently, varying $\gamma$ within this range allows us to control the influence of local information $\{\mathbf{X}_k, \mathbf{y}_k\}$ on our prediction. Particularly, when $\gamma \neq 0$, the prediction $\mu_\ast(\boldsymbol{\alpha})$ is given as a combination of both the global information summarised by $\boldsymbol{\alpha}$ and local information $\{\mathbf{X}_k, \mathbf{y}_k\}$. Otherwise, when $\gamma = 0$, the first term in the expression of $\mu_\ast(\boldsymbol{\alpha})$ becomes zero, leaving $\mu_\ast(\boldsymbol{\alpha})$ to be independent of $\mathbf{s}$. Interestingly, while $\mathbf{s}$ is essential to enable the anytime learning of ${\boldsymbol{\theta}}$ (Section 4), Section 5 empirically shows that given the learned ${\boldsymbol{\theta}}$, the predictive performance is maximized when there is no influence of $\mathbf{s}$ (i.e., $\gamma = 0$), which implies its nuisance nature during inference.
\fi

\noindent 
{\bf Remark 1} 
The special case of $\gamma = 1$ recovers the degenerate test conditional 
$p(f_{\mathbf{x}_\ast}|\boldsymbol{\alpha}) = \mathcal{N}(\boldsymbol{\phi}_{\boldsymbol{\theta}}^\top(\mathbf{x}_\ast)\mathbf{s}, 0)$ 
%$f_{\mathbf{x}_\ast} \sim \mathcal{N}(\boldsymbol{\Phi}_{\boldsymbol{\theta}}^\top(\mathbf{x}_\ast)\mathbf{s}, 0)$ 
induced by the \emph{variational SSGP} (VSSGP) model of~\citeauthor{Yarin15}~\shortcite{Yarin15} (see equation $4$ and Section~$3$ therein) which reveals that it imposes a highly restrictive deterministic relationship between $f_{\mathbf{x}_\ast}$ and $\boldsymbol{\alpha}$ and also  
%While the achieved testing conditional is consistent with the structural assumption in~\eqref{eq:imagine}, this formulation 
fails to exploit the local data $(\mathbf{X}_k,\mathbf{y}_k)$ (i.e., due to conditional independence between $f_{\mathbf{x}_\ast}$ and $\mathbf{y}_k$ given $\boldsymbol{\alpha}$) that can potentially improve the predictive performance. 
Unfortunately, VSSGP cannot be trivially extended to span the entire spectrum since it relies on the induced deterministic relationship between $f_{\mathbf{x}_\ast}$ and $\boldsymbol{\alpha}$ to analytically derive its predictive distribution, which does not hold  for $\gamma \neq 1$. 
On the other hand, when $\gamma = 0$, the test conditional in Proposition~\ref{lem:l1} becomes independent of the nuisance variables $\mathbf{s}$ and reduces to the predictive distribution $p(f_\mathbf{x_\ast} | \mathbf{y}_k, \boldsymbol{\theta})$ of the SSGP model of~\citeauthor{Miguel10}~\shortcite{Miguel10} (see equation $7$ therein) given its point estimate of the spectral frequencies ${\boldsymbol{\theta}}$ but restricted to the local data $(\mathbf{X}_k,\mathbf{y}_k)$ corresponding to input subspace $\mathcal{X}_k$. 
Hence, $\gamma$ can also be perceived as a controlling parameter that trades off between the contributions of the degenerate test conditional of VSSGP vs. the \emph{local} SSGP model to constructing a test conditional in our generalized framework. 
%More importantly, unlike our VBGPR, the work of \citeauthor{Yarin15}~\shortcite{Yarin15} assumes the test outputs are conditionally independent with the training data given the learned parameters (see Eq. 4 of \citeauthor{Yarin15}~\shortcite{Yarin15}), thus impairing their predictive performance as empirically shown in our experiments (Section 5).
To investigate this trade-off, our experiments in Section~\ref{experiment} show that given the learned ${\boldsymbol{\theta}}$, the predictive performance is maximized at $\gamma = 0$ when 
the test conditional $p(f_\mathbf{x_\ast} | \mathbf{y}_k, \boldsymbol{\alpha})$ depends on local data $(\mathbf{X}_k,\mathbf{y}_k)$ but not $\mathbf{s}$, which justifies viewing $\mathbf{s}$ as a ``nuisance'' to prediction despite its crucial role in scalable learning of ${\boldsymbol{\theta}}$ via stochastic optimization (Section~\ref{anytime}).\vspace{1mm}
%
%More interestingly, when $\gamma = 0$, the testing conditional in Lemma~\ref{lem:l1} reproduces that of SSGP restricted to the local partition $(\mathbf{X}_k,\mathbf{y}_k)$
%The above spectrum of testing conditionals introduced by Lemma~\ref{lem:l1} interestingly recovers the conditional of \cite{Yarin15} when $\gamma = 1$.

In the sections to follow, we will propose a novel stochastic optimization method for deriving a \emph{variationally optimal} approximation to the posterior distribution $p(\boldsymbol{\alpha} | \mathbf{y})$ that is used to marginalize out the global information $\boldsymbol{\alpha}$ from any test conditional $p(f_\mathbf{x_\ast} | \mathbf{y}_k, \boldsymbol{\alpha})$ in Proposition~\ref{lem:l1} to yield an asymptotic approximation to the predictive distribution $p(f_\mathbf{x_\ast} | \mathbf{y})$~\eqref{eq:decompose} regardless of the value of $\gamma\in[-1,1]$.\vspace{-0mm}  
%
%So, using~\eqref{eq:decompose} and Lemma~\ref{lem:l1}, we instead propose a different solution paradigm which exploits the re-parameterisation technique of \cite{Titsias14} to facilitate an asymptotic approximation of the predictive distribution regardless of $\gamma$. 
%
% our prediction task ultimately reduces to obtaining/approximating the posterior distribution of the global information $p(\boldsymbol{\alpha} | \mathbf{y})$. To achieve this, Section 3 introduces a novel algorithm to construct a \emph{variationally optimal} approximation for $p(\boldsymbol{\alpha} | \mathbf{y})$ which can be exploited to establish our anytime algorithm in Section 4.
%
\section{Variational Inference for Bayesian Sparse Spectrum Gaussian Process Regression}
\label{vbgpr}
This section presents a variational approximation $q(\boldsymbol{\alpha})$ of the posterior distribution $p(\boldsymbol{\alpha}|\mathbf{y})$ achieved by using variational inference 
%\cite{Bishop06} 
which  involves choosing a parameterization for $q(\boldsymbol{\alpha})$ (Section~\ref{parameterisation}) and optimizing its defining parameters (Section~\ref{optimisation}) to minimize its \emph{Kullback-Leibler} (KL) distance ${D}_{\mathrm{KL}}(q) \triangleq \mathrm{KL}(q(\boldsymbol{\alpha}) \| p(\boldsymbol{\alpha}|\mathbf{y}))$ to $p(\boldsymbol{\alpha}|\mathbf{y})$. The optimized $q(\boldsymbol{\alpha})$ can then be used as a tractably cheap surrogate of $p(\boldsymbol{\alpha}|\mathbf{y})$ for marginalizing out $\boldsymbol{\alpha}$ from  test conditional $p(f_\mathbf{x_\ast} | \mathbf{y}_k, \boldsymbol{\alpha})$ in Proposition~\ref{lem:l1} to derive an approximation to  predictive distribution $p(f_\mathbf{x_\ast} | \mathbf{y})$~\eqref{eq:decompose} efficiently (Section~\ref{anytime}).
\subsection{Variational Parameterization}
\vspace{-0.4mm}
\label{parameterisation}
Specifically, we parameterize $\boldsymbol{\alpha} = \mathrm{vec}({\boldsymbol{\theta}},\mathbf{s}) \triangleq \mathbf{Mz + b}$ where the variational parameters $\boldsymbol{\eta} \triangleq \mathrm{vec}(\mathbf{M},\mathbf{b})$ are independent of $({\boldsymbol{\theta}} ,\mathbf{s})$, and $\mathbf{z}$ is distributed by an analytically  tractable user-specified distribution $\psi(\mathbf{z}) \simeq p(\mathbf{z} | \mathbf{y})$ that is straightforward to draw samples from (i.e., $\mathbf{z} \sim \psi(\mathbf{z})$). We assume that $\psi(\mathbf{z})$ is analytically differentiable with respect to $\mathbf{z}$ and the affine matrix $\mathbf{M}$ is invertible such that $\mathbf{z} = \mathbf{M}^{-1}(\boldsymbol{\alpha} - \mathbf{b})$ exists. Then, $q({\boldsymbol{\theta}},\mathbf{s})$ can be expressed in terms of $\mathbf{M}$, $\mathbf{b}$, and $\psi(\mathbf{z})$:
\begin{lemma}
\label{lem:l2}
%Let $\mathbf{z} \sim \psi(\mathbf{z})$ and $\boldsymbol{\alpha} = \mathbf{Mz + b}$. Then,
The variational distribution $q({\boldsymbol{\theta}}, \mathbf{s})$ can be indirectly parameterized via $\psi(\mathbf{z})$ by $q({\boldsymbol{\theta}},\mathbf{s}) = q(\boldsymbol{\alpha}) =  \psi(\mathbf{M}^{-1}(\boldsymbol{\alpha} - \mathbf{b}))/|\mathbf{M}| = \psi(\mathbf{z})/|\mathbf{M}|$
%\begin{eqnarray}
%\label{eqn:a3}
 %q({\boldsymbol{\theta}},\mathbf{s}) \ \triangleq\ q(\boldsymbol{\alpha}) \ =\  \frac{1}{|\mathbf{M}|}\psi\left(\mathbf{M}^{-1}(\boldsymbol{\alpha} - \mathbf{b})\right) \ =\ \frac{1}{|\mathbf{M}|}\psi\left(\mathbf{z}\right)
%\end{eqnarray}
where $|\mathbf{M}|$ denotes the absolute value of $\mathrm{det}(\mathbf{M})$.
\end{lemma} 
%\vspace{-2mm}
Lemma~\ref{lem:l2} follows directly from the change of variables theorem. 
%a result of \citeauthor{Titsias14}~\shortcite{Titsias14} (see Section~$2$ therein).
Using the above parameterization of $\boldsymbol{\alpha}$, we can also express the prior $p({\boldsymbol{\theta}},\mathbf{s})$ in terms of variational parameters $\mathbf{M}$ and $\mathbf{b}$. To see this, recall that $p({\boldsymbol{\theta}}) = \mathcal{N}(\mathbf{0}, {\boldsymbol{\Theta}})$ and $p(\mathbf{s}) = \mathcal{N}(\mathbf{0}, \boldsymbol{\Lambda})$ (see Section~\ref{hgp} and~\eqref{eq:imagine}). This  implies $p(\boldsymbol{\alpha}) = p(\mathrm{vec}({\boldsymbol{\theta}},\mathbf{s})) = p({\boldsymbol{\theta}})\ p(\mathbf{s}) = \mathcal{N}(\boldsymbol{\alpha} | \mathbf{0}, \mathrm{blkdiag}[{\boldsymbol{\Theta}},\boldsymbol{\Lambda}])$. Plugging in the expression of $\boldsymbol{\alpha}$ yields $p(\boldsymbol{\alpha}) = \mathcal{N}(\mathbf{M}\mathbf{z} + \mathbf{b} | \mathbf{0}, \mathrm{blkdiag}[{\boldsymbol{\Theta}},\boldsymbol{\Lambda}])$\footnote{
%This result does not conflict with our earlier assumption of $\mathbf{z} \sim \psi(\mathbf{z})$ 
Note that $\psi(\mathbf{z})$ is meant to approximate the posterior distribution of $\mathbf{z}$ given $\mathbf{y}$ instead of its prior distribution. So, it cannot be used to derive $p(\boldsymbol{\alpha})$ by applying the affine transformation on $\mathbf{z}$.}. 
At this time, the need to parameterize $q(\boldsymbol{\alpha})$ in Lemma~\ref{lem:l2} may not be obvious to a reader because it is motivated by a technical necessity to guarantee the asymptotic convergence of our stochastic optimization method (Remark $4$) rather than a conceptual intuition.\vspace{1mm}

\noindent {\bf Remark 2} Our generalized framework enables both the spectral frequencies ${\boldsymbol{\theta}}$ and the nuisance variables $\mathbf{s}$ to interact \emph{a posteriori} by modeling them jointly in the variational distribution $q({\boldsymbol{\theta}},\mathbf{s})$, as detailed in Lemma~\ref{lem:l2}, and still preserves scalability (Section~\ref{anytime}).
In contrast, the VSSGP model of \citeauthor{Yarin15}~\shortcite{Yarin15} assumes    
%${\boldsymbol{\theta}}$ 
$\mathbf{r}_1, \ldots, \mathbf{r}_m, \text{and}\ \mathbf{s}$ to be statistically independent \emph{a posteriori} in its variational distribution (see section $4$ therein).
%The idea of adopting a Bayesian treatment for learning both the spectral frequencies ${\boldsymbol{\theta}} = \{\mathbf{r}_1, \mathbf{r}_2, \ldots \mathbf{r}_{m}\}$ and the nuisance variables $\mathbf{s}$ has recently been employed in \citeauthor{Yarin15}~\shortcite{Yarin15}. However, while the VSSGP framework proposed by \citeauthor{Yarin15}~\shortcite{Yarin15} assumes that $\mathbf{s}$ and ${\boldsymbol{\theta}}$ are statistically independent a posteriori (see Eq. (10) of \citeauthor{Yarin15}~\shortcite{Yarin15}). 
Relaxing this assumption will cause 
%the existing formulation of 
VSSGP to lose its scalability as its induced variational lower bound will inevitably become intractable. 
\vspace{1mm}

%as demonstrated in Section 5, having this relaxation indeed contributes towards a better performance overall as compared to VSSGP when tested on the same datasets. 

\noindent {\bf Remark 3} Though the model of \citeauthor{Titsias14}~\shortcite{Titsias14} has adopted a similar parameterization but only for the original GP hyperparameters, it incurs cubic time in the data size per iteration of gradient ascent update, as shown in its supplementary materials and experiments. A factorized marginal likelihood has to be further assumed for this parameterization to achieve scalability. 
%Independently from our work, a similar parameterisation has been previously adopted by \citeauthor{Titsias14}~\shortcite{Titsias14}, which, however, only involves the hyperparameters ${\boldsymbol{\theta}}$. Furthermore, the work of \citeauthor{Titsias14}~\shortcite{Titsias14} has to further assume a factorised \emph{marginal} likelihood $p(\mathbf{y}|{\boldsymbol{\theta}})$ for this parameterization to achieve scalability. As a matter of fact, when this assumption is not met, their proposed algorithm scales cubically in the size of data as shown in one of their experiments on hyperparameter learning for FGPR models. 
Our framework does not require such a strong assumption and still scales well to million-sized datasets (Section~\ref{experiment}). 
This is, perhaps surprisingly, achieved through introducing the nuisance variables $\mathbf{s}$ (Section~\ref{hgp}), which interestingly embeds a linearly decomposable structure within the log-likelihood of data $\log p(\mathbf{y}|\boldsymbol{\alpha})$ instead of $\log p(\mathbf{y}|{\boldsymbol{\theta}})$. As shown later in Theorem~\ref{theo:t1}, such a structure is the key ingredient for developing our stochastic optimization method (Section~\ref{anytime}).\vspace{-1mm}
%\vspace{-1mm}
\subsection{Variational Optimization}
%\vspace{-1mm}
\label{optimisation}\vspace{-0.4mm}
To optimize $q(\boldsymbol{\alpha})$ (i.e., by minimizing ${D}_{\mathrm{KL}}(q)$),
% with $\boldsymbol{\alpha} \triangleq \mathrm{vec}({\boldsymbol{\theta}}, \mathbf{s})$, 
we first show that the log-marginal likelihood $\log p(\mathbf{y})$ can be decomposed into a sum of a lower-bound functional ${L}(q)$ and ${D}_{\mathrm{KL}}(q)$: $\log p(\mathbf{y}) = {L}(q) + {D}_\mathrm{KL}(q)$
%\begin{eqnarray}
%\log p\left(\mathbf{y}\right) &=& {L}\left(q\right) \ +\  {D}_\mathrm{KL}\left(q\right) \ ,\label{eqn:a4}
%\end{eqnarray}
where ${L}(q) \triangleq \mathbb{E}_{\boldsymbol{\alpha}\sim q(\boldsymbol{\alpha})}[\log p(\mathbf{y}|\boldsymbol{\alpha}) -\log ({q(\boldsymbol{\alpha})}/{p(\boldsymbol{\alpha})})]$, as detailed in\if\myproof1 Appendix~\ref{app:b}\fi\if\myproof0 \cite{MinhAAAI17}\fi.
%$\boldsymbol{\eta}_{t+1} = \boldsymbol{\eta}_{t} + \rho_t \partial\widehat{{L}}(\boldsymbol{\eta}_t)/\partial\boldsymbol{\eta}$
So, minimizing ${D}_\mathrm{KL}(q)$ is equivalent to maximizing ${L}(q)$ with respect to the variational parameter $\boldsymbol{\eta} = \mathrm{vec}(\mathbf{M},\mathbf{b})$ of $q(\boldsymbol{\alpha})$ (Lemma~\ref{lem:l2}) since $\log p(\mathbf{y})$ is a constant (i.e., independent of $\boldsymbol{\eta}$). 
In practice, this is usually achieved by setting the derivative $\partial{L}/\partial\boldsymbol{\eta} = 0$ and solving for $\boldsymbol{\eta}$, which is unfortunately  intractable since its exact analytic expression is not known. 

To sidestep this intractability issue, we instead adopt a stochastic optimization method that is capable of maximizing ${L}(q)$ via iterative stochastic gradient ascent updates. %\cite{Monro1951}. 
However, this also requires the stochastic gradient $\partial\widehat{{L}}/\partial\boldsymbol{\eta}$ \eqref{eqn:a5c} to be an \emph{analytically tractable} and \emph{unbiased} estimator of the exact gradient $\partial{L}/\partial\boldsymbol{\eta}$. 
To derive this, we first exploit Lemma~\ref{lem:l2} to re-express ${L}(q)$ as an expectation with respect to $\mathbf{z}\sim\psi(\mathbf{z})$:  
${L}(q) = \mathbb{E}_{\mathbf{z}\sim\psi(\mathbf{z})}[\log p(\mathbf{y}|\boldsymbol{\alpha}) - \log({q(\boldsymbol{\alpha})}/{p(\boldsymbol{\alpha})})]$ 
where $\boldsymbol{\alpha} = \mathbf{M}\mathbf{z} + \mathbf{b}$, as shown in\if\myproof1 Appendix~\ref{app:c}\fi\if\myproof0 \cite{MinhAAAI17}\fi.
Then, taking the derivatives with respect to $\boldsymbol{\eta}$ on both sides of the above equation yields
\begin{equation}
\partial{L}/\partial\boldsymbol{\eta} =
\mathbb{E}_{\mathbf{z}\sim\psi(\mathbf{z})}[\ \partial(\log p(\mathbf{y}|\boldsymbol{\alpha}) - \log (q(\boldsymbol{\alpha})/p(\boldsymbol{\alpha})))/\partial\boldsymbol{\eta}\ ] 
\label{eqn:a5b}
\end{equation}
which reveals a simple, analytically tractable (proven in\if\myproof1 Appendix~\ref{app:d} \fi\if\myproof0 \cite{MinhAAAI17} \fi 
using Lemma~\ref{lem:l2}) choice for our stochastic gradient 
\begin{equation}
\partial\widehat{{L}}/\partial\boldsymbol{\eta} \triangleq
\partial(\log p(\mathbf{y}|\boldsymbol{\alpha}) - \log (q(\boldsymbol{\alpha})/p(\boldsymbol{\alpha})))/\partial\boldsymbol{\eta}\ . 
\label{eqn:a5c}
\end{equation}
%$$
%{\partial\widehat{{L}}}/{\partial\boldsymbol{\eta}} \triangleq {\partial}/{\partial\boldsymbol{\eta}}(\log p(\mathbf{y}|\boldsymbol{\alpha}) - \log({q(\boldsymbol{\alpha})}/{p(\boldsymbol{\alpha})}))
%$$ 
%where $\boldsymbol{\alpha} = \mathbf{M}\mathbf{z} + \mathbf{b}$ and $\mathbf{z} \sim \psi(\mathbf{z})$. 
Thus, $\mathbb{E}_{\mathbf{z}\sim\psi(\mathbf{z})}[\partial\widehat{{L}}/\partial\boldsymbol{\eta}] = \partial{L}/\partial\boldsymbol{\eta}$ guarantees that $\partial\widehat{{L}}/\partial\boldsymbol{\eta}$ is indeed an unbiased estimator of $\partial{L}/\partial\boldsymbol{\eta}$.\vspace{1mm}

\noindent 
{\bf Remark 4}~\eqref{eqn:a5b} reveals that parameterizing $q(\boldsymbol{\alpha})$ indirectly through $\psi(\mathbf{z})$ in Lemma~\ref{lem:l2} is essential to enabling    stochastic optimization in our generalized framework: Since $\psi(\mathbf{z})$ does not depend on the variational parameters $\boldsymbol{\eta} = \mathrm{vec}(\mathbf{M},\mathbf{b})$, the derivative operator in \eqref{eqn:a5b} can be moved inside the expectation, which trivially reveals an unbiased stochastic estimate \eqref{eqn:a5c} of the exact gradient via sampling $\mathbf{z}$. Otherwise, suppose that we attempt to derive a stochastic gradient for ${L}(q)$ using the original expression ${L}(q) = \mathbb{E}_{\boldsymbol{\alpha}\sim q(\boldsymbol{\alpha})}[\log p(\mathbf{y}|\boldsymbol{\alpha}) -\log ({q(\boldsymbol{\alpha})}/{p(\boldsymbol{\alpha})})]$ instead of~\eqref{eqn:a5b} in the same manner with a direct parameterization of  $q(\boldsymbol{\alpha})$ not exploiting $\psi(\mathbf{z})$. Then, after differentiating both sides of the above expression with respect to $\boldsymbol{\eta}$, the derivative operator on the RHS cannot be moved inside the expectation over $\boldsymbol{\alpha}\sim q(\boldsymbol{\alpha})$ since it depends on $\boldsymbol{\eta}$, which suggests that deriving an unbiased estimator for $\partial{L}/\partial\boldsymbol{\eta}$ has become non-trivial without using the parameterization of $q(\boldsymbol{\alpha})$ in Lemma~\ref{lem:l2}.
\vspace{1mm} 

\iffalse
Furthermore, it can be shown that the above derivation of $\partial\widehat{{L}}/\partial\boldsymbol{\eta}$ in Eq.~\eqref{eqn:a5c}, unlike that of the exact gradient $\partial{L}/\partial\boldsymbol{\eta}$ in Eq.~\eqref{eqn:a5}, is in fact tractable, thereby guaranteeing the tractability of our stochastic gradient update in Eq.~\eqref{eqn:a5d}. To understand this, it suffices to show that both the derivative terms, $\partial\log (q(\boldsymbol{\alpha})/p(\boldsymbol{\alpha}))/\partial\boldsymbol{\eta}$ and $\partial\log p(\mathbf{y}|\boldsymbol{\alpha})/\partial\boldsymbol{\eta}$, which appear in the expression of $\partial\widehat{{L}}/\partial\boldsymbol{\eta}$ can be tractably computed. This is particularly true for $\partial\log (q(\boldsymbol{\alpha})/p(\boldsymbol{\alpha}))/\partial\boldsymbol{\eta}$ since the closed-form expressions of $p(\boldsymbol{\alpha})$ and $q(\boldsymbol{\alpha})$ have been previously expressed in terms of $\mathbf{M}$ and $\mathbf{b}$ (Section 3.1), which consequently allows us to compute $\partial\log (q(\boldsymbol{\alpha})/p(\boldsymbol{\alpha}))/\partial\mathbf{M}$ and $\partial\log (q(\boldsymbol{\alpha})/p(\boldsymbol{\alpha}))/\partial\mathbf{b}$ and aggregate the results to construct $\partial\log (q(\boldsymbol{\alpha})/p(\boldsymbol{\alpha}))/\partial\boldsymbol{\eta} = \mathrm{vec}(\partial\log (q(\boldsymbol{\alpha})/p(\boldsymbol{\alpha}))/\partial\mathbf{M}, \partial\log (q(\boldsymbol{\alpha})/p(\boldsymbol{\alpha}))/\partial\mathbf{b})$ analytically (see Appendix~\ref{app:d} for a mored detailed derivation). 
\fi

To understand why ${\partial\widehat{{L}}}/{\partial\boldsymbol{\eta}}$~\eqref{eqn:a5c} is analytically tractable, it suffices to show that
both its derivatives $\partial\log p(\mathbf{y}|\boldsymbol{\alpha})/\partial\boldsymbol{\eta}$ and $\partial\log (q(\boldsymbol{\alpha})/p(\boldsymbol{\alpha}))/\partial\boldsymbol{\eta}$ in~\eqref{eqn:a5c} are analytically tractable, the latter of which is detailed in\if\myproof1 Appendix~\ref{app:d}\fi\if\myproof0 \cite{MinhAAAI17}\fi.
For the former,
since the trigonometric basis functions $\{\phi_{\boldsymbol{\theta}}^i(\mathbf{x})\}_{i=1}^{2m}$ (Section~\ref{hgp}) are differentiable, it can be shown that $\partial\log p(\mathbf{y}|\boldsymbol{\alpha})/\partial\boldsymbol{\eta}$ is analytically tractable by exploiting the following result giving a closed-form expression of $\log p(\mathbf{y} |\boldsymbol{\alpha})$ in terms of ${\boldsymbol{\theta}}$ and $\mathbf{s}$:
%, as detailed in 
%Lemma~\ref{lem:l3} (see Appendix~\ref{app:e} for a detailed proof):
%
\begin{lemma}
\label{lem:l3}
%Let $p(\mathbf{s}) \triangleq \mathcal{N}(\mathbf{s}\ |\ 0, \boldsymbol{\Lambda})$ as induced from Eq.~\eqref{eq:imagine}, 
$\log p(\mathbf{y}|\boldsymbol{\alpha}) = -0.5 \sigma_n^{-2}\mathbf{v}^\top\mathbf{v} - 0.5n\log(2\pi\sigma_n^2)$
%\begin{eqnarray}
%\log p(\mathbf{y}|\boldsymbol{\alpha}) &=& -\frac{1}{2\sigma_n^2}\mathbf{v}^\top\mathbf{v} + \mathcal{C} \ , \label{eqn:a6}
%\end{eqnarray}
where $\mathbf{v} \triangleq \mathbf{y} - \boldsymbol{\Phi}^\top_{\boldsymbol{\theta}}(\mathbf{X})\mathbf{s}$. 
%$\boldsymbol{\alpha} \triangleq \mathrm{vec}({\boldsymbol{\theta}},\mathbf{s})$ 
%and $c\triangleq $.
\end{lemma} 
Its proof is in\if\myproof1 Appendix~\ref{app:e}\fi\if\myproof0 \cite{MinhAAAI17}\fi.
%Appendix~\ref{app:e}. 
The above choice of ${\partial\widehat{{L}}}/{\partial\boldsymbol{\eta}}$~\eqref{eqn:a5c} and Lemma~\ref{lem:l3} show that it is not only an unbiased estimator of the exact gradient but is also analytically tractable, which satisfies all the required conditions to guarantee the asymptotic convergence of 
%our iterative stochastic gradient ascent updates, thereby concluding the technical feasibility of 
our proposed stochastic optimization method. 
However, a critical issue remains that makes it scale poorly in practice: It can be derived from Lemma~\ref{lem:l3} that naively evaluating its derivative $\partial\log p(\mathbf{y}|\boldsymbol{\alpha})/\partial\boldsymbol{\eta}$ in~\eqref{eqn:a5c} in a straightforward manner incurs 
linear time in  data size $n$ per iteration of stochastic gradient ascent update, 
%$\mathcal{O}(nm)$ time per iteration where $n$ is the size of training data and $m$ is proportional to the number of basis functions (Section~\ref{hgp}), 
which is  expensive for massive datasets.\vspace{-1mm} 
\section{Stochastic Optimization}
\label{anytime}
To overcome the issue of scalability in evaluating the stochastic gradient ${\partial\widehat{{L}}}/{\partial\boldsymbol{\eta}}$~\eqref{eqn:a5c} (Section~\ref{optimisation}), we  will show in Theorem~\ref{theo:t1} below that $\partial\log p(\mathbf{y} | \boldsymbol{\alpha})/\partial\boldsymbol{\eta}$ is decomposable into a linear sum of analytically tractable terms, each of which depends on only a small subset of local data. 
Interestingly, since only the derivative $\partial\log p(\mathbf{y} | \boldsymbol{\alpha})/\partial\boldsymbol{\eta}$ in~\eqref{eqn:a5c} involves the training data $(\mathbf{X},\mathbf{y})$, Theorem~\ref{theo:t1} implies a similar decomposition of $\partial\widehat{{L}}/\partial\boldsymbol{\eta}$. As a result, we can derive a new stochastic estimate of the exact gradient $\partial{L}/\partial\boldsymbol{\eta}$ 
%(Appendix~\ref{app:g}) 
that can be computed efficiently and scalably using only one or a few randomly sampled subset(s) of local data of fixed size and still preserves the property of being its unbiased estimator (Section~\ref{revisit}). Computing this new stochastic gradient incurs only constant time in the data size $n$ per iteration of stochastic gradient ascent update, 
%the processing cost of our update equation with respect to this new stochastic gradient remains constant (per iteration) with respect to the data size, 
which, together with Proposition~\ref{lem:l1} in Section~\ref{hgp}, constitute the foundation of our generalized framework of stochastic variational Bayesian SSGP regression models for big data (Section~\ref{infer}).\vspace{-0.2mm} 
\subsection{Stochastic Gradient Revisited}
\label{revisit}
To derive a computationally scalable and unbiased stochastic gradient, we rely on our main result below showing the decomposability of $\partial\log p(\mathbf{y}|\boldsymbol{\alpha})/\partial\boldsymbol{\eta}$ in~\eqref{eqn:a5c} into a linear sum of analytically tractable terms, each of which depends on only a small subset of local training data of fixed size:
\begin{theorem}
\label{theo:t1}
Let
$\mathbf{v}_i \triangleq \mathbf{y}_i - \boldsymbol{\Phi}^\top_{\boldsymbol{\theta}}(\mathbf{X}_i)\mathbf{s}$ for $i = 1,\ldots,p$. 
%\mathbf{r}_k \triangleq [r_k^u]^\top_{u\in\boldsymbol{\eta}}$ where $r_k^u \triangleq \sigma_n^{-2}(\partial/\partial u)\left(\mathbf{v}^\top_k\mathbf{v}_k\right) \ .$
Then, \vspace{0.5mm} $\partial\log p(\mathbf{y}|\boldsymbol{\alpha})/\partial\boldsymbol{\eta} = \sum_{i=1}^{p} {F}_i(\boldsymbol{\eta},\boldsymbol{\alpha})$ where ${F}_i(\boldsymbol{\eta},\boldsymbol{\alpha}) \triangleq  -0.5\sigma_n^{-2}\nabla_{\boldsymbol{\eta}}(\mathbf{v}^\top_i\mathbf{v}_i)$ is analytically tractable.
\end{theorem}
Its proof in\if\myproof1 Appendix~\ref{app:f} \fi\if\myproof0 \cite{MinhAAAI17} \fi
utilizes Lemma~\ref{lem:l3}.
Using Theorem~\ref{theo:t1} , 
%Then, by substituting the derived expression of $\partial\log p(\mathbf{y}|\boldsymbol{\alpha})/\partial\boldsymbol{\eta}$ in Theorem~\ref{theo:t1} into that of , 
the following linearly decomposable structure of ${\partial\widehat{{L}}}/{\partial\boldsymbol{\eta}}$~\eqref{eqn:a5c} results:\vspace{-1mm}
\begin{equation}
\partial\widehat{{L}}/\partial\boldsymbol{\eta} = \mathbb{E}_{i\sim \mathcal{U}(1,p)}[\ p{F}_i(\boldsymbol{\eta},\boldsymbol{\alpha}) - {\partial}\log(q(\boldsymbol{\alpha})/p(\boldsymbol{\alpha}))/\partial\boldsymbol{\eta}\ ] 
\label{eqn:a9}\vspace{-1mm}
\end{equation}
where $i$ is treated as a discrete random variable uniformly distributed over the set of partition indices $\{1, 2, \ldots, p\}$.
% in the last equality of~\eqref{eqn:a9}. 
This interestingly reveals an unbiased estimator for the stochastic gradient $\partial\widehat{{L}}/\partial\boldsymbol{\eta}$ which can be constructed stochastically by sampling $i$:\vspace{-1mm}  
$${\partial\widetilde{{L}}}/{\partial\boldsymbol{\eta}} \triangleq p{F}_i(\boldsymbol{\eta},\boldsymbol{\alpha}) - {\partial}\log({q(\boldsymbol{\alpha})}/{p(\boldsymbol{\alpha})})/{\partial\boldsymbol{\eta}}\vspace{-1mm}$$ 
such that substituting it into~\eqref{eqn:a9} yields $\mathbb{E}_{i\sim \mathcal{U}(1,p)}[\partial\widetilde{{L}}/\partial\boldsymbol{\eta}] = \partial\widehat{{L}}/\partial\boldsymbol{\eta}$. 
Then, taking the expectation over $\mathbf{z} \sim \psi(\mathbf{z})$ on both sides of this equality gives $\mathbb{E}_{\mathbf{z} \sim \psi(\mathbf{z})}[\mathbb{E}_{i\sim \mathcal{U}(1,p)}[\partial\widetilde{{L}}/\partial\boldsymbol{\eta}]] = \mathbb{E}_{\mathbf{z} \sim \psi(\mathbf{z})}[\partial\widehat{{L}}/\partial\boldsymbol{\eta}] = \partial{L}/\partial\boldsymbol{\eta}$, which proves that $\partial\widetilde{{L}}/\partial\boldsymbol{\eta}$ is also an unbiased estimator of $\partial{L}/\partial\boldsymbol{\eta}$ that can be constructed by sampling both $i \sim \mathcal{U}(1,p)$ and $\mathbf{z} \sim \psi(\mathbf{z})$ (Section~\ref{parameterisation}) independently. 
Replacing $\partial\widehat{{L}}/\partial\boldsymbol{\eta}$ with $\partial\widetilde{{L}}/\partial\boldsymbol{\eta}$ produces a highly efficient stochastic gradient ascent update that incurs constant time in the data size $n$ per iteration since $\partial\widetilde{{L}}/\partial\boldsymbol{\eta}$ can be computed using only a single randomly sampled subset of local data $(\mathbf{X}_i,\mathbf{y}_i)$ of fixed size. 

Instead of utilizing just a single pair $(i, \mathbf{z})$ of samples, the above stochastic gradient ${\partial\widetilde{{L}}}/{\partial\boldsymbol{\eta}}$ can be generalized to simultaneously process multiple pairs of independent samples and their corresponding sampled subsets of local data in one stochastic gradient ascent update (detailed in\if\myproof1 Appendix~\ref{app:g}\fi\if\myproof0 \cite{MinhAAAI17}\fi), thereby improving the rate of convergence while preserving its property of an unbiased estimator of the exact gradient ${\partial{{L}}}/{\partial\boldsymbol{\eta}}$. %This enables us to evaluate the stochastic update more efficiently by replacing $\partial\widehat{{L}}/\partial\boldsymbol{\eta}$ with $\partial\tilde{{L}}/\partial\boldsymbol{\eta}$ using  ${\partial\tilde{{L}}}/{\partial\boldsymbol{\eta}} \triangleq p\mathbf{F}_k(\boldsymbol{\eta},\boldsymbol{\alpha}) - {\partial}/{\partial\boldsymbol{\eta}}(\log({q(\boldsymbol{\alpha})}/{p(\boldsymbol{\alpha})}))$. 
Asymptotic convergence of the estimate of $\boldsymbol{\eta}$ (and hence the estimate of $q(\boldsymbol{\alpha})$) is guaranteed if the step sizes are scheduled appropriately.\vspace{-1mm} 
\subsection{Approximate Predictive Inference}
\label{infer}\vspace{-0.3mm}
%Using the deterministic relationship between $f_{\mathbf{x}_\ast}$ and $\boldsymbol{\alpha} \triangleq \mathrm{vec}({\boldsymbol{\theta}},\mathbf{s})$ in Lemma~\ref{lem:l1}, 
%
%We can also provide approximate anytime inference by sampling $\boldsymbol{\alpha}$ from the current estimation of $q(\boldsymbol{\alpha})$ after each update iteration. 
%
In iteration $t$ of stochastic gradient ascent update, an estimate $q_t(\boldsymbol{\alpha})$ of the variationally optimal approximation $q(\boldsymbol{\alpha})$ can be induced from the current estimate $\boldsymbol{\eta}_t = \mathrm{vec}(\mathbf{M}_t, \mathbf{b}_t)$ of its variational parameters $\boldsymbol{\eta}$ using the parameterization in Lemma~\ref{lem:l2}. As a result, using the law of iterated expectations, the predictive mean can be approximated by $\widehat{\mu}_{\mathbf{x}_\ast} = \mathbb{E}_{\boldsymbol{\alpha} \sim q_t(\boldsymbol{\alpha})}[\mathbb{E}[f_{\mathbf{x}_\ast} | \mathbf{y}_k, \boldsymbol{\alpha}]]= \mathbb{E}_{\boldsymbol{\alpha} \sim q_t(\boldsymbol{\alpha})}[ \mu_{\mathbf{x}_\ast}\hspace{-0.7mm}(\boldsymbol{\alpha}) ] =\int_{\boldsymbol{\alpha}}q_t(\boldsymbol{\alpha})\mu_{\mathbf{x}_\ast}\hspace{-0.7mm}(\boldsymbol{\alpha})\mathrm{d}\boldsymbol{\alpha}$
%
%\begin{eqnarray}
%\mathbb{E}_{\boldsymbol{\alpha} \sim q_t(\boldsymbol{\alpha})}\Big[f_{\mathbf{x}_\ast} | \mathbf{y}\Big] &=&\int_{\boldsymbol{\alpha}} \hspace{-1mm}q_t(\boldsymbol{\alpha})\mu_\ast(\boldsymbol{\alpha})\mathrm{d}\boldsymbol{\alpha} , \label{eqn:a11a}
%\end{eqnarray}
%
where $\boldsymbol{\alpha} = \mathrm{vec}({\boldsymbol{\theta}},\mathbf{s})$ and $\mu_{\mathbf{x}_\ast}\hspace{-0.7mm}(\boldsymbol{\alpha})$ is previously defined in Proposition~\ref{lem:l1}. 
%This results directly from integrating $q_t(\boldsymbol{\alpha})$ with the analytical expression of $p(f_{\mathbf{x}_\ast} | \boldsymbol{\alpha},\mathbf{y}_k)$ (see Proposition~\ref{lem:l1}). 
However, since the integration over $\boldsymbol{\alpha}$ is not always analytically tractable,
%Its tractability depends on the particular user-specified distribution $\psi(\mathbf{z})$ which underlines the parameterization of $q({\boldsymbol{\theta}},\mathbf{s})$ in Lemma~\ref{lem:l2}. 
%\footnote{For many choices of $\psi(\mathbf{z})$ parameterizing $q({\boldsymbol{\alpha}})$ in Lemma~\ref{lem:l2}, the integration over $\boldsymbol{\alpha}$ cannot be evaluated analytically.}, 
%${\partial\widetilde{{L}}}/{\partial\boldsymbol{\eta}}$ is not tractable.
we  approximate it by drawing i.i.d. samples $\boldsymbol{\alpha}_1,\ldots, \boldsymbol{\alpha}_r$ from $q_t(\boldsymbol{\alpha})$ to estimate $\widehat{\mu}_{\mathbf{x}_\ast} \simeq r^{-1}\sum_{i=1}^r\mu_{\mathbf{x}_\ast}\hspace{-0.7mm}(\boldsymbol{\alpha}_i)$. 
Similarly, using the variance decomposition formula and definition of variance, the predictive variance can be approximated by $\widehat{\sigma}^2_{\mathbf{x}_\ast}
= \mathbb{E}_{\boldsymbol{\alpha} \sim q_t(\boldsymbol{\alpha})}[ \sigma^2_{\mathbf{x}_\ast}\hspace{-0.7mm}(\boldsymbol{\alpha}) ] + \mathbb{V}_{\boldsymbol{\alpha} \sim q_t(\boldsymbol{\alpha})}[ \mu_{\mathbf{x}_\ast}\hspace{-0.7mm}(\boldsymbol{\alpha}) ]
= \mathbb{E}_{\boldsymbol{\alpha} \sim q_t(\boldsymbol{\alpha})}[ \sigma^2_{\mathbf{x}_\ast}\hspace{-0.7mm}(\boldsymbol{\alpha}) + \mu^2_{\mathbf{x}_\ast}\hspace{-0.7mm}(\boldsymbol{\alpha}) ] -\widehat{\mu}^2_{\mathbf{x}_\ast}
\simeq r^{-1}\sum_{i=1}^r(\sigma^2_{\mathbf{x}_\ast}\hspace{-0.7mm}(\boldsymbol{\alpha}_i)+\mu^2_{\mathbf{x}_\ast}\hspace{-0.7mm}(\boldsymbol{\alpha}_i) ) -\widehat{\mu}^2_{\mathbf{x}_\ast}$ where $\sigma^2_{\mathbf{x}_\ast}\hspace{-0.7mm}(\boldsymbol{\alpha})$ is previously defined in Proposition~\ref{lem:l1} and the $\mathbb{V}_{\boldsymbol{\alpha} \sim q_t(\boldsymbol{\alpha})}[ \mu_{\mathbf{x}_\ast}\hspace{-0.7mm}(\boldsymbol{\alpha}) ]$ term arises due to the uncertainty of $\boldsymbol{\alpha}$. 
%\begin{eqnarray}
%\mathbb{E}_{\boldsymbol{\alpha}\sim q_t(\boldsymbol{\alpha})}\Big[f_{\mathbf{x}_\ast} | \mathbf{y}\Big] &\simeq& \frac{1}{r}\sum_{i=1}^r\mu_\ast(\boldsymbol{\alpha}_i) . \label{eqn:a12}
%\end{eqnarray}
The samples $\boldsymbol{\alpha}_1,\ldots, \boldsymbol{\alpha}_r$ are in turn obtained by sampling $\mathbf{z}_1,\ldots, \mathbf{z}_r$ from $\psi(\mathbf{z})$ and applying the parametric transformation in Lemma~\ref{lem:l2} with respect to the current estimates $\mathbf{M}_t$ and $\mathbf{b}_t$, that is, $\mathrm{vec}({\boldsymbol{\theta}}_i, \mathbf{s}_i) = \boldsymbol{\alpha}_i = \mathbf{M}_t\mathbf{z}_i + \mathbf{b}_t$ for $i=1,\ldots,r$. 
Computing the predictive mean $\widehat{\mu}_{\mathbf{x}_\ast}$ and variance $\widehat{\sigma}^2_{\mathbf{x}_\ast}$ thus incurs constant time in the data size $n$, hence achieving efficient approximate predictive inference.\vspace{-0.8mm} 
%
%This suggests that the cost of computing the predictive mean does not depend on the data size, hence allowing us to efficiently produce low-cost predictions.
%
\section{Empirical Studies}
\label{experiment}%\vspace{-0.5mm}
This section empirically evaluates the predictive performance and time efficiency of our $s$VBSSGP model on three real-world datasets: (a) The AIMPEAK dataset \cite{Chen13} consists of $41800$ traffic speed observations (km/h) along $775$ urban road segments during the morning peak hours on April $20$, $2011$. Each observation features a $5$-dimensional input vector of a road segment's length, number of lanes, speed limit, direction, and its recording time (i.e., discretized into $54$ five-minute time slots), and a corresponding output measuring the traffic speed (km/h);   
(b) the benchmark AIRLINE dataset \cite{Hensman13,NghiaICML15} contains $2000000$ information records of US commercial flights in $2008$. Each record features a $8$-dimensional input vector of the aircraft's age (year), travel distance (km), the flight's total airtime, departure time, arrival time (min), and the date given by day of week, day of month, and month, and a corresponding output of the flight's delay time (min); and
(c) the BLOG feedback dataset \cite{Buza14} contains $60000$ instances of blog posts. Each blog post features a fairly large $60$-dimensional input vector associated with its first $60$ attributes described at https://archive.ics.uci.edu/ml/datasets/BlogFeedback, and a corresponding output measuring the number of comments in the next $24$ hours.
The BLOG dataset is used to evaluate the robustness of $s$VBSSGP to overfitting which usually occurs in training with datasets of high input dimensions.
%Then, we will evaluate the robustness of $s$VBSSGP to overfitting which usually happens on high-dimensional datasets, we also evaluate its performance on (c) the BLOG feedback dataset from UCI Machine Learning Repository \cite{Buza14} with 60000 data points, each represented by 10 features and their average, standard deviation, min, max, and median (60 attributes in total)\footnote{These attributes correspond to features 1-60 described in the UCI repository.}. The predictive output is the number of comments in the next $24$ hours.
All datasets are modeled using GPs with prior covariance defined in Section~\ref{hgp} and split into $95\%$ training data and $5\%$ test data.
All experimental results are averaged over $5$ random splits.  
For the AIMPEAK, AIRLINE, and BLOG datasets, we use, respectively, $2m = 40, 40, \text{and}\ 10$ trigonometric basis functions to approximate the GP kernel~\eqref{eq:kernel} and sample $\mathbf{z}$ from $\mathcal{N}(\mathbf{0}, \mathbf{I})$ (Section~\ref{parameterisation}). All experiments are run on a Linux system with Intel$\circledR$ Xeon$\circledR$ E$5$-$2670$ at $2.6$GHz with $96$ GB memory.

%We model all datasets using VBGPR with covariance defined in~\eqref{eq:kernel}, which is trained on 95$\%$ of data points and tested with the remaining 5$\%$. 
%Each tested model is evaluated 
The performance of $s$VBSSGP is compared against the state-of-the-art VSSGP \cite{Yarin15} and stochastic implementations of sparse GP models based on inducing variables such as DTC$+$ and PIC$+$ \cite{Hensman13,NghiaICML15}
(i.e., run with their GitHub codes) using the following metrics: (a) \emph{Root mean square error} (RMSE) $\{|\overline{\mathbf{X}}|^{-1}\sum_{\mathbf{x}_*\in\overline{\mathbf{X}}}(y_{\mathbf{x}_*} - \widehat{\mu}_{\mathbf{x}_*})^2\}^{1/2}$ 
over the set $\overline{\mathbf{X}}$ of test inputs,
%of the test predictions $\{\widehat{y}_\mathbf{x}\}_{\mathbf{x}\in\mathcal{U}}$, 
(b) \emph{mean negative log probability} (MNLP) $0.5|\overline{\mathbf{X}}|^{-1}\sum_{{\mathbf{x}_*}\in\overline{\mathbf{X}}}\{(y_{\mathbf{x}_*} - \widehat{\mu}_{\mathbf{x}_*})^2/\widehat{\sigma}^2_{\mathbf{x}_*} + \mathrm{log}(2\pi\widehat{\sigma}^2_{\mathbf{x}_*})\}$,
% with $\{\widehat{\sigma}^2_\mathbf{x}\}_{\mathbf{x}\in\mathcal{U}}$,
% being the variances of predictions, 
 and (c) training time.\vspace{1mm}
% vs. no. of training iterations. 
\begin{figure}[b]\vspace{-2.6mm}
%	\begin{small}
		\begin{tabular}{cccc}
			\hspace{1mm}\includegraphics[width=1.85cm]{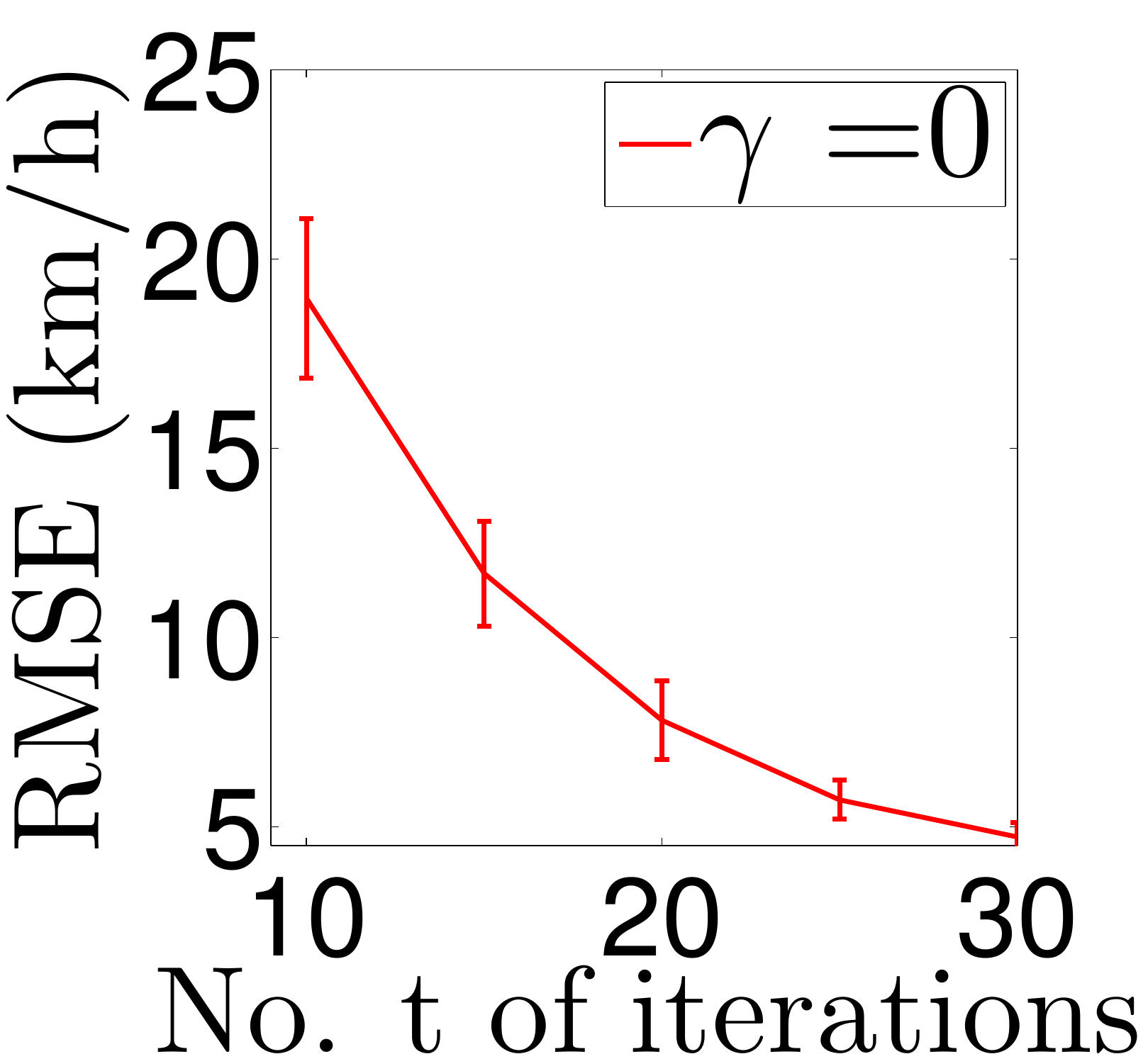} & \hspace{-3mm}\includegraphics[width=1.85cm]{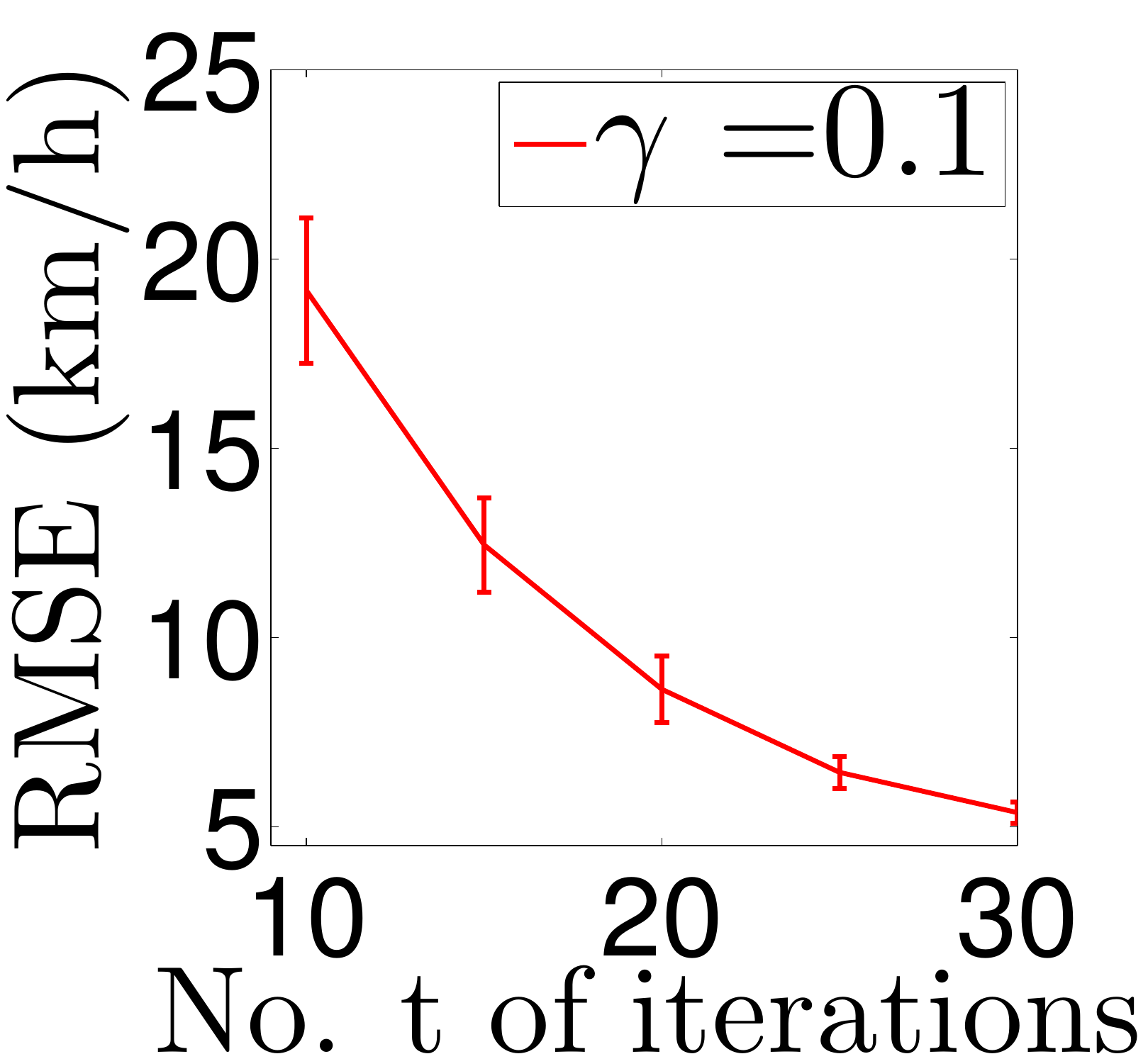} & \hspace{-3mm}\includegraphics[width=1.85cm]{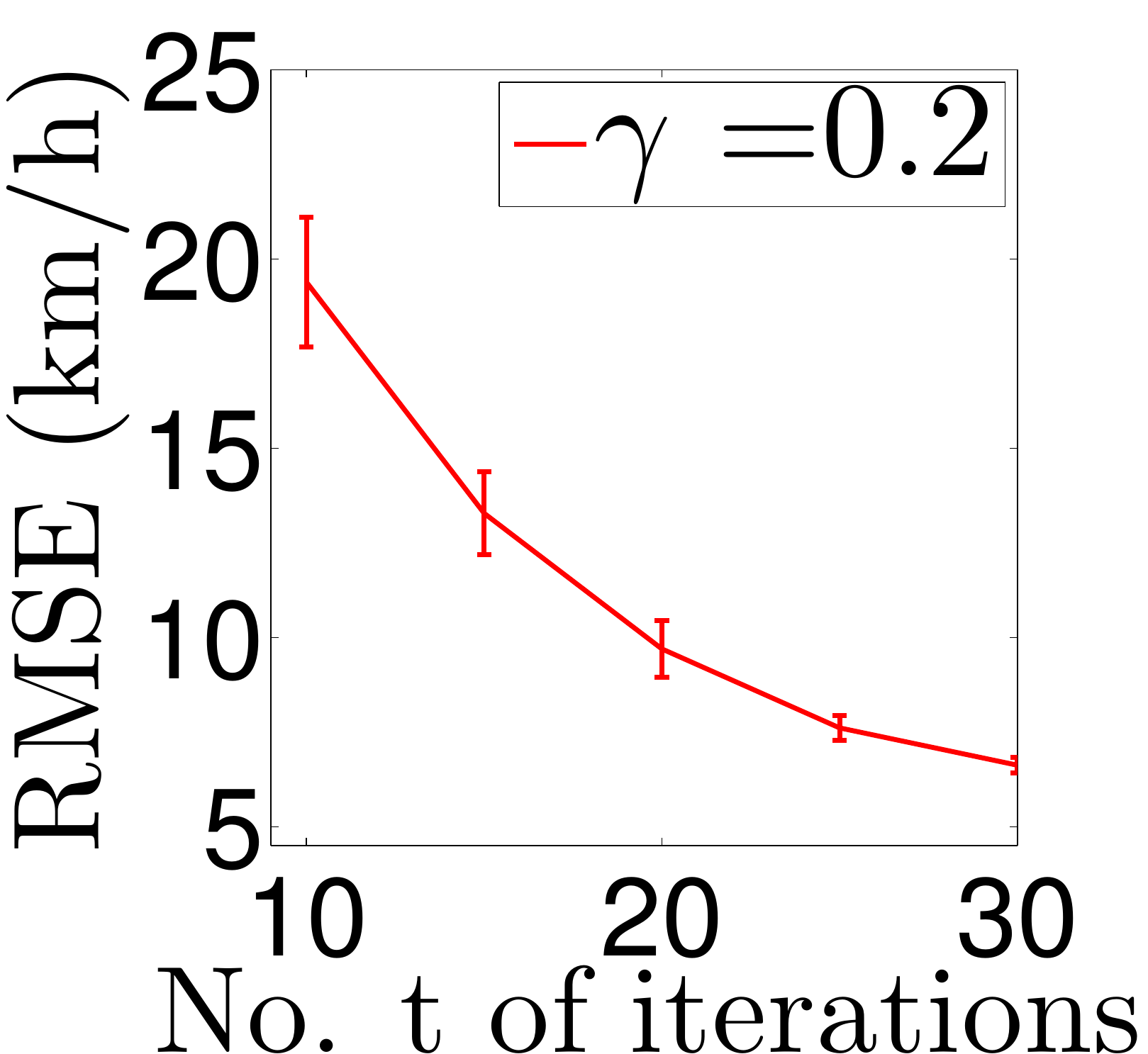} & \hspace{-3mm}\includegraphics[width=1.85cm]{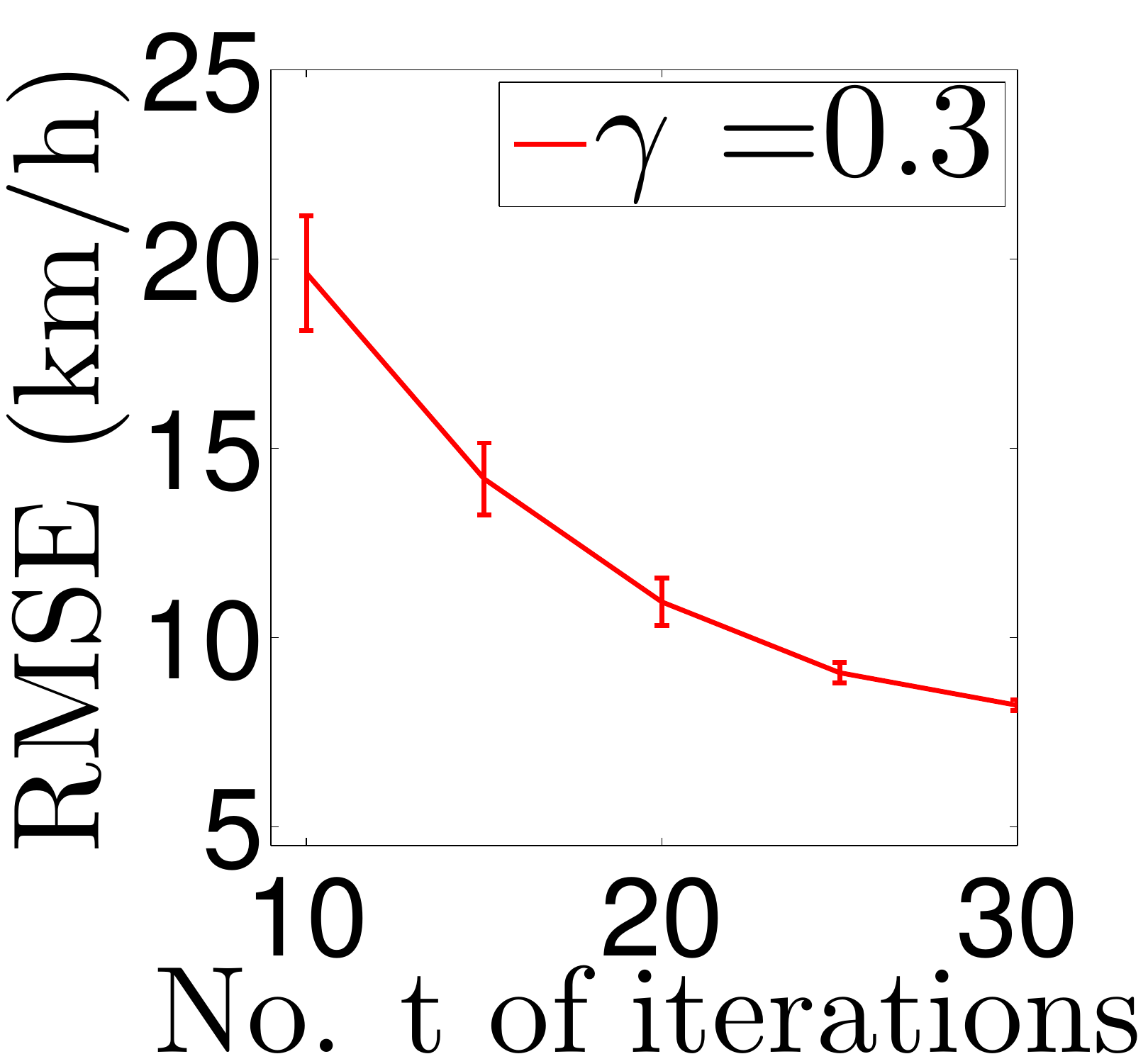} \vspace{-3mm}
%			\vspace{-1mm}\\
%			\hspace{-2mm}(a) & \hspace{-3mm}(b) & \hspace{-3mm}(c) & \hspace{-3mm}(d)
		\end{tabular}
%	\end{small}
	\caption{Graphs of RMSEs (with standard deviations) achieved by $s$VBSSGP vs. number $t$ of iterations with varying values of $\gamma$
%$\gamma =$ (a) $0.0$, (b) $\gamma = 0.1$, (c) $\gamma = 0.2$, and (d) $\gamma = 0.3$ 
for the AIMPEAK dataset.\vspace{-1mm}}
	\label{fig:aimpeak}
\end{figure}

\begin{figure*}
%	\begin{small}
		\begin{tabular}{cccc}
			\hspace{4mm}\includegraphics[height=3.1cm]{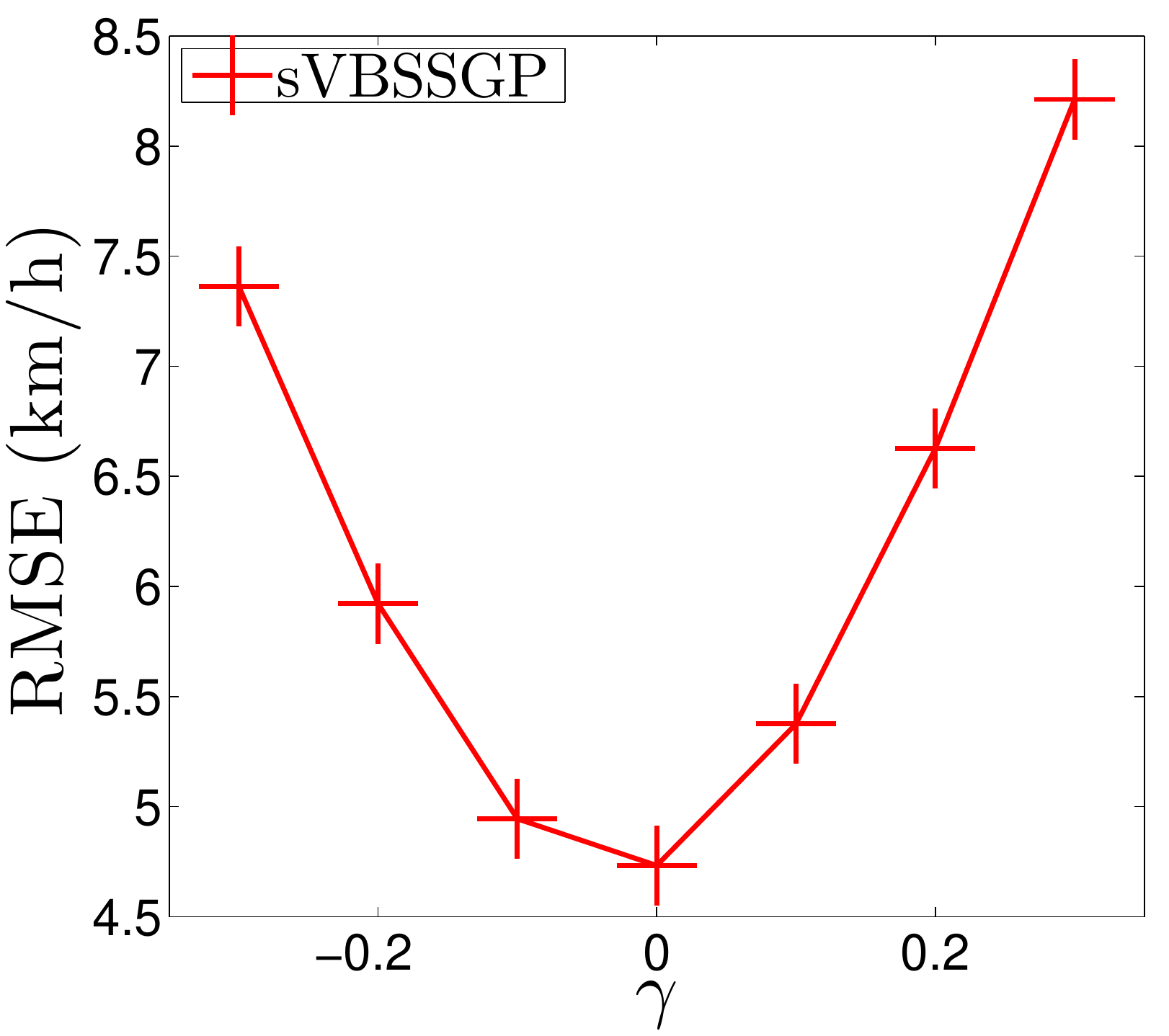} &
			\hspace{2mm}\includegraphics[height=3.1cm]{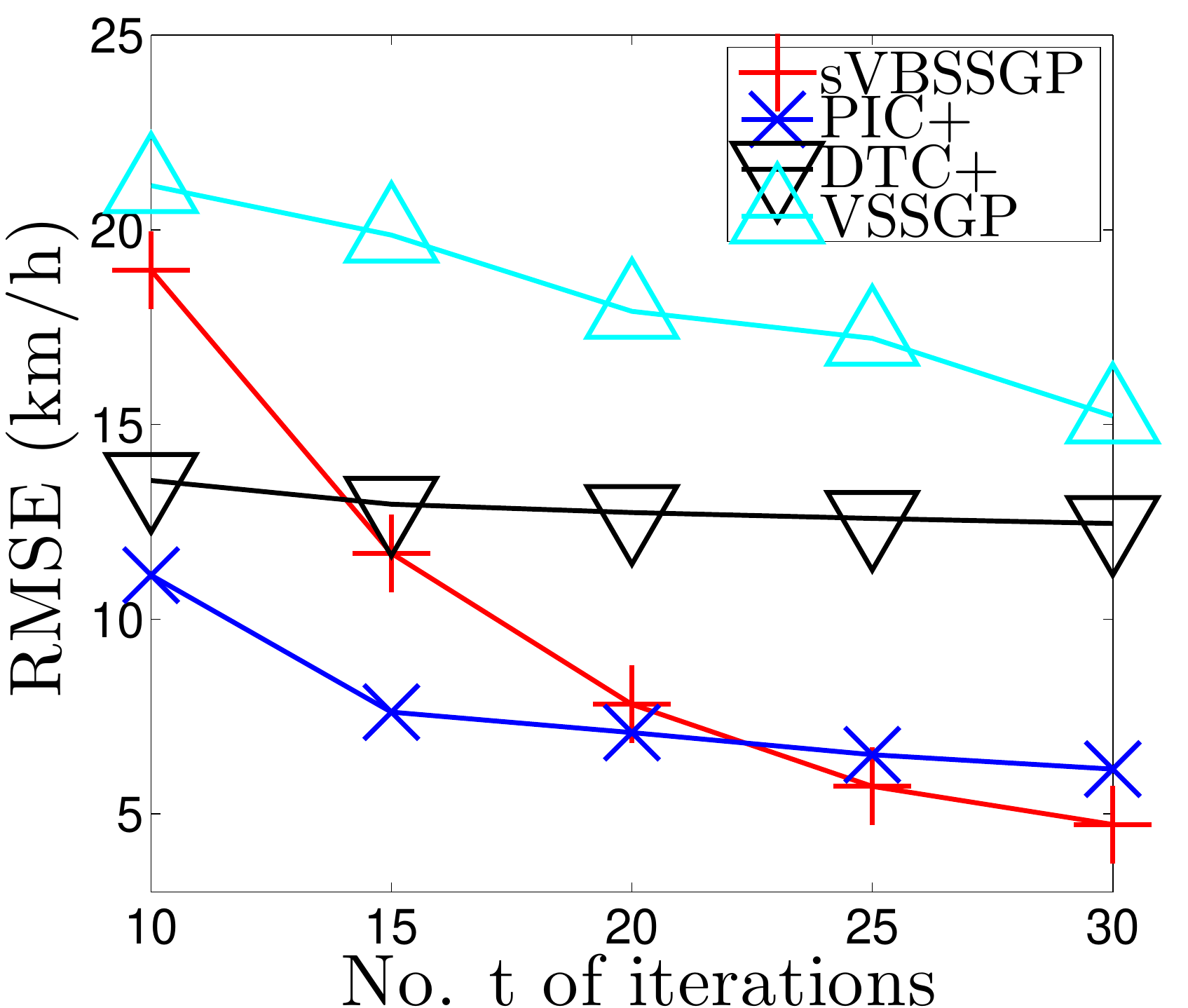} &
			\hspace{2mm}\includegraphics[height=3.1cm]{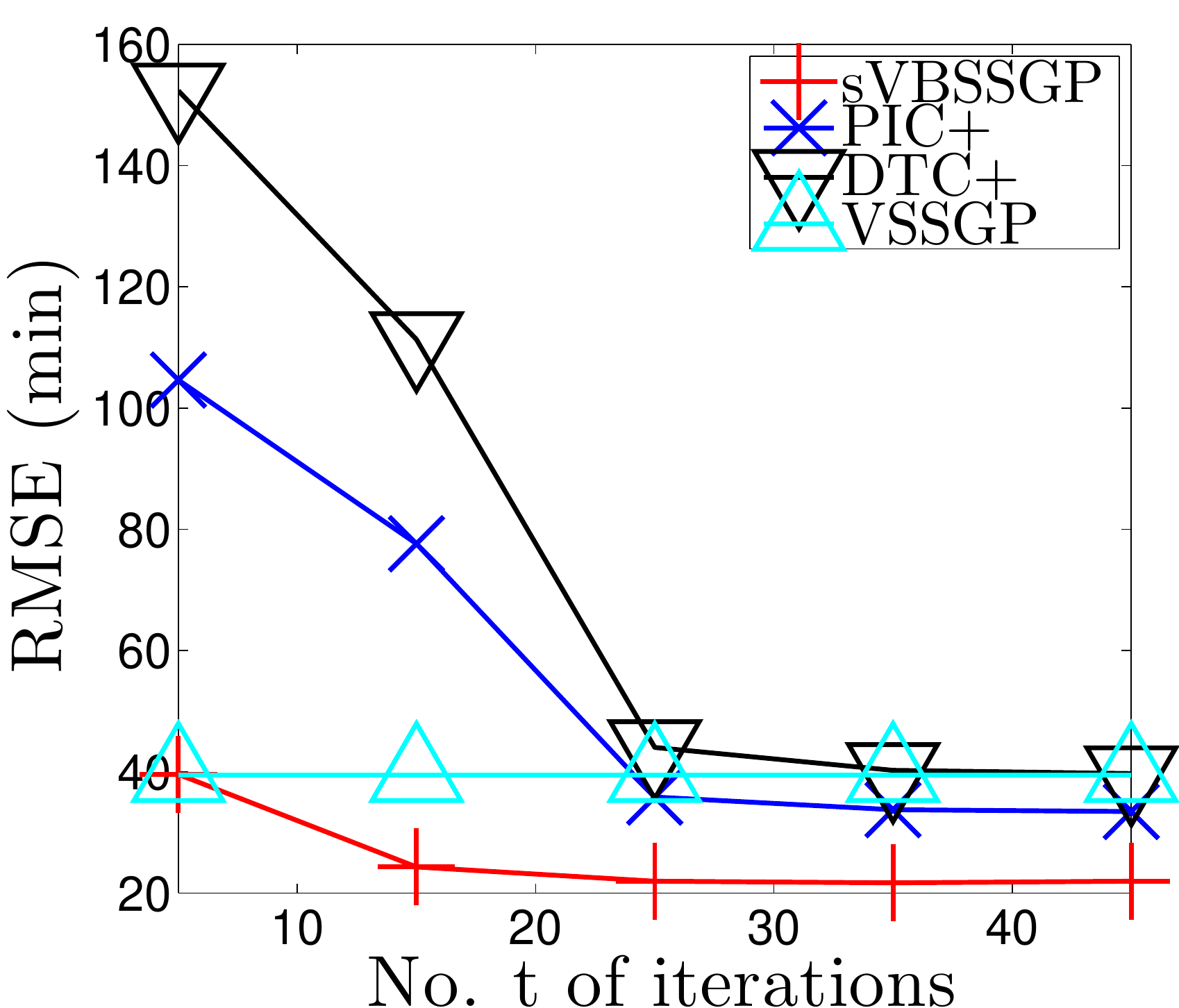} &
			\hspace{2mm}\includegraphics[height=3.1cm]{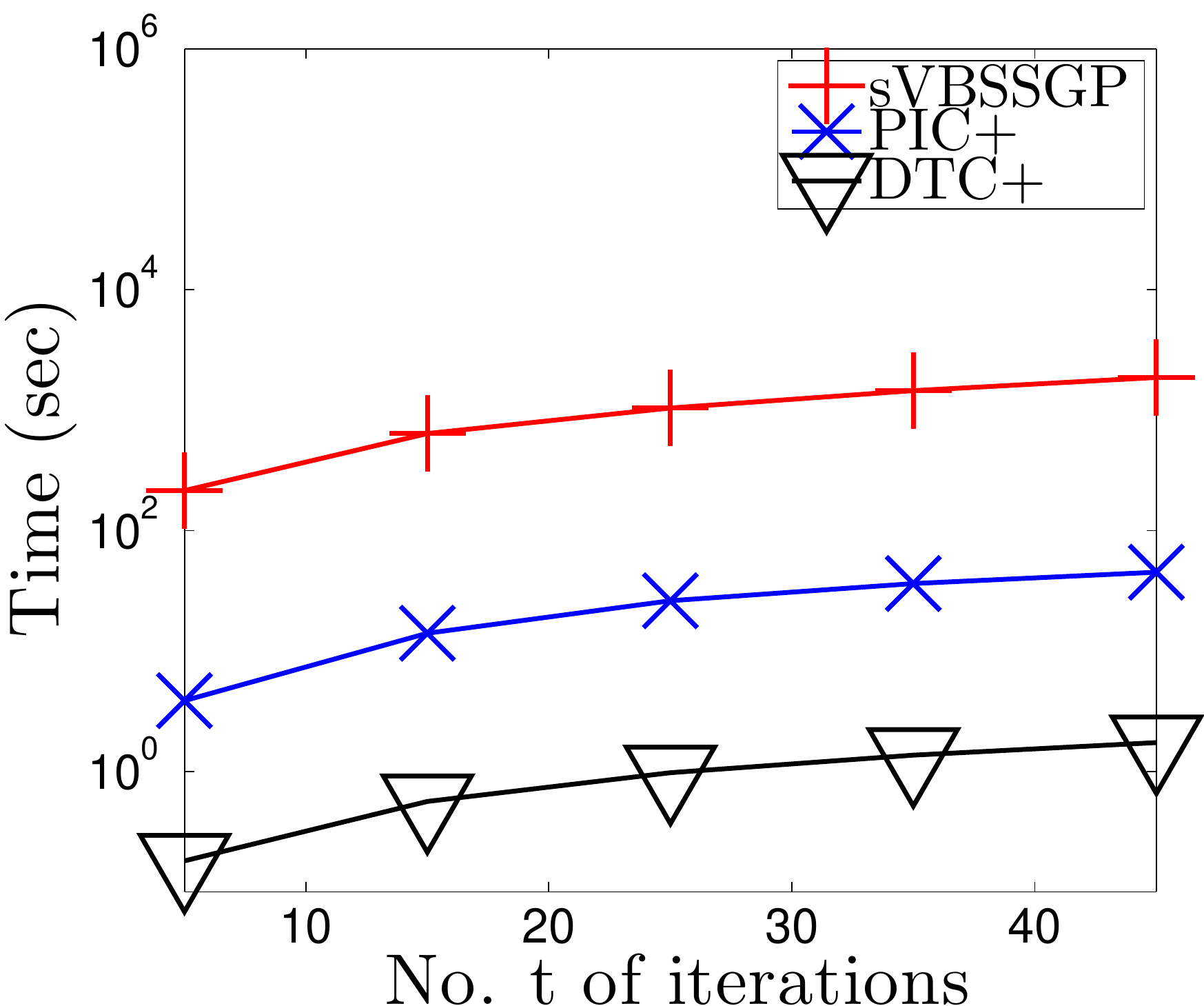} \vspace{-1mm}
			\\ 
			\hspace{4mm}(a) & \hspace{2mm}(b) & \hspace{2mm}(c) & \hspace{2mm}(d) \vspace{-3.5mm}
		\end{tabular}		
%	\end{small}
	\caption{(a) Graph of RMSEs achieved by $s$VBSSGP vs. $\gamma$ for the AIMPEAK dataset; graphs of RMSEs achieved by $s$VBSSGP, PIC$+$, DTC$+$, and VSSGP vs. number $t$ of iterations for the (b) AIMPEAK and (c) AIRLINE datasets; and (d) graph of total training time incurred by $s$VBSSGP, PIC$+$, and DTC$+$ vs. number $t$ of iterations for the AIRLINE dataset.\vspace{-4.5mm}}
	\label{fig:compare}
\end{figure*}

\noindent
{\bf AIMPEAK Dataset.} This dataset is evenly partitioned into $p = 200$ disjoint subsets using $k$-means ($k=p$). 
Fig.~\ref{fig:aimpeak} shows results of RMSEs of $s$VBSSGP for $\gamma = 0, 0.1, 0.2, 0.3$ that
%on the AIMPEAK dataset (measured by RMSE and MNLP) 
rapidly decrease by $4$- to $5$-fold over $30$ iterations. 
It can also be observed that increasing $\gamma$ from $0$ results in higher converged RMSEs.
%
%As $\gamma$ moves away from 0.0, we also observe that the RMSE and MNLP curves tend to converge at higher values (poorer performance). 
To further investigate this, 
%we also test $s$VBSSGP with various values of $\gamma$ along the spectrum $[-1.0,1.0]$. 
Fig.~\ref{fig:compare}a reveals that $s$VBSSGP indeed achieves the lowest converged RMSE at $\gamma = 0$ among all tested values of $\gamma$. 
%showing converged RMSEs plotted against $\gamma$ suggests that at $\gamma = 0.0$,  indeed yields the best performance among all evaluated values of $\gamma$. 
This confirms our hypothesis stated earlier in Remark $1$ that given the learned spectral frequencies $\boldsymbol{\theta}$, the information carried by $\mathbf{s}$ becomes a ``nuisance'' to prediction despite its interaction with $\boldsymbol{\theta}$ in $q(\boldsymbol{\theta}, \mathbf{s}) = q(\boldsymbol{\alpha})$ during stochastic optimization. 
%crucial role in scalable learning of $\boldsymbol{\theta}$ via stochastic optimization 
%redundant in characterizing the testing outputs. 
That is, when the influence of $\mathbf{s}$ on the test output is completely removed from the test conditional by setting $\gamma = 0$ in Proposition~\ref{lem:l1},
% (though it still interacts with the spectral frequencies during the learning phase), 
the predictions are no longer interfered by the nuisance information of $s$, hence explaining the lowest RMSE achieved by $s$VBSSGP at $\gamma = 0$. 

%n_alpha = 5 for AIRLINE and 20 for AIMPEAK
Fig.~\ref{fig:compare}b and Table~\ref{tab:exp} show that $s$VBSSGP ($\gamma = 0$ and $r=20$) significantly outperforms VSSGP, DTC$+$, and PIC$+$ ($250$ inducing variables) in terms of RMSE after $30$ iterations. 
To explain this, DTC$+$ and PIC$+$ find point estimates of the kernel hyperparameters, which may have resulted in their poorer performance.
Though VSSGP also adopts a Bayesian treatment of the spectral frequencies, it uses a degenerate test conditional corresponding to the case of $\gamma = 1$ in Proposition~\ref{lem:l1}. As a result, VSSGP imposes a highly restrictive deterministic relationship between the test output and spectral frequencies and also fails to exploit the local data for prediction (see Remark $1$).
%
%: This fails to utilize local information during prediction and further imposes a restrictive deterministic relationship between the testing outputs and spectral frequencies.
%
%This is expected because: (a) unlike DTC$+$ and PIC$+$ which train their hyperparameters beforehand using a small subset of the training data, our approach instead learns their spectral analogues (e.g., $\mathbf{M},\mathbf{b}$) using the entire dataset.  
The results of the MNLP metric are similar and reported in\if\myproof1 Appendix~\ref{supp}\fi\if\myproof0 \cite{MinhAAAI17}\fi.\vspace{1mm} 
%\vspace{0mm}
\iffalse
\begin{table}[t]
	\vspace{0mm}
	\begin{small}
	%\hspace{1mm}
	\begin{tabular}{r!{\vrule width 1pt}ccccc}\noalign{\hrule height 1pt \vspace{0.5mm}}
		& VBGPR \hspace{-1mm}& DTC+ \hspace{-1mm}& PIC+ \hspace{-1mm}& SSGP \hspace{-1mm}& VSSGP \\ \hline \vspace{-2mm} \\ %\noalign{\hrule height 0.8pt} 
		\hspace{-2mm}AIMPEAK \hspace{-1mm} & {\bf 4.73} \hspace{-1mm}& $12.46$ \hspace{-1mm}& $6.15$ \hspace{-1mm}& $11.30$ \hspace{-1mm}&  $9.46$ \\ 
		\hspace{-2mm}AIRLINE \hspace{-1mm}& {\bf 22.18} \hspace{-1mm}& $39.60$ \hspace{-1mm}& $33.40$ \hspace{-1mm}& N/A \hspace{-1mm}& $39.49$ \\
		\hspace{-2mm}ONLINE \hspace{-1mm}& {\bf 0.95} \hspace{-1mm}& $0.97$ \hspace{-1mm}& $1.08$ \hspace{-1mm}& 0.95 \hspace{-1mm}& $0.96$ \\ \noalign{\hrule height 1pt} 
	\end{tabular}
	\end{small}
	%\vspace{-2mm}
	%\label{tab:exp}
	\caption{RMSEs achieved by VBGPR, DTC+, PIC+, SSGP and VSSGP on all datasets.}
	%\end{tiny}
	\label{tab:exp}
	\vspace{-1mm}
\end{table}
\fi
\begin{figure}[b]
	\vspace{-4mm}
%	\begin{small}
		\begin{tabular}{cc}
			\hspace{-0mm}\includegraphics[height=3.1cm]{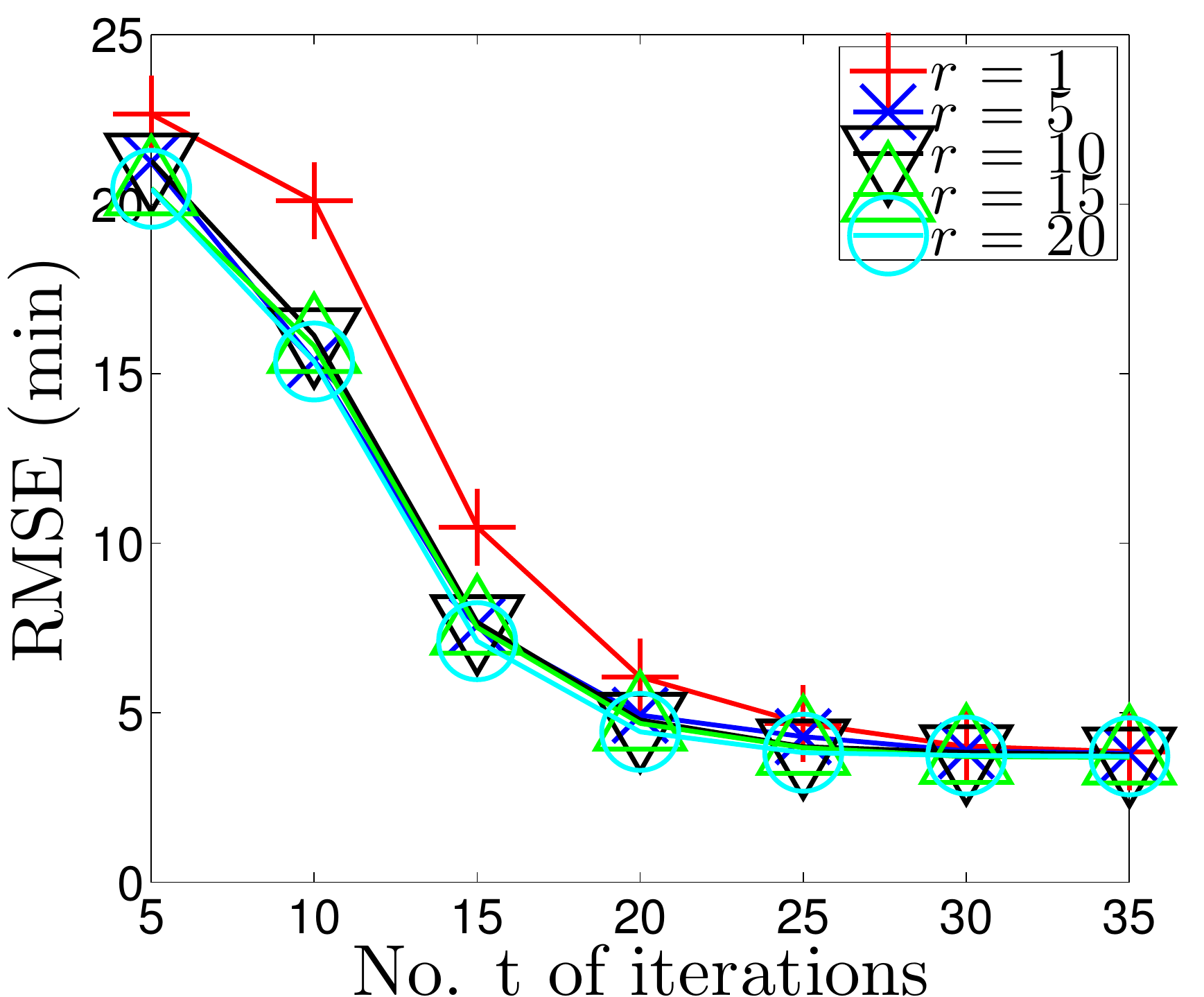} &
			\hspace{-2mm}\includegraphics[height=3.1cm]{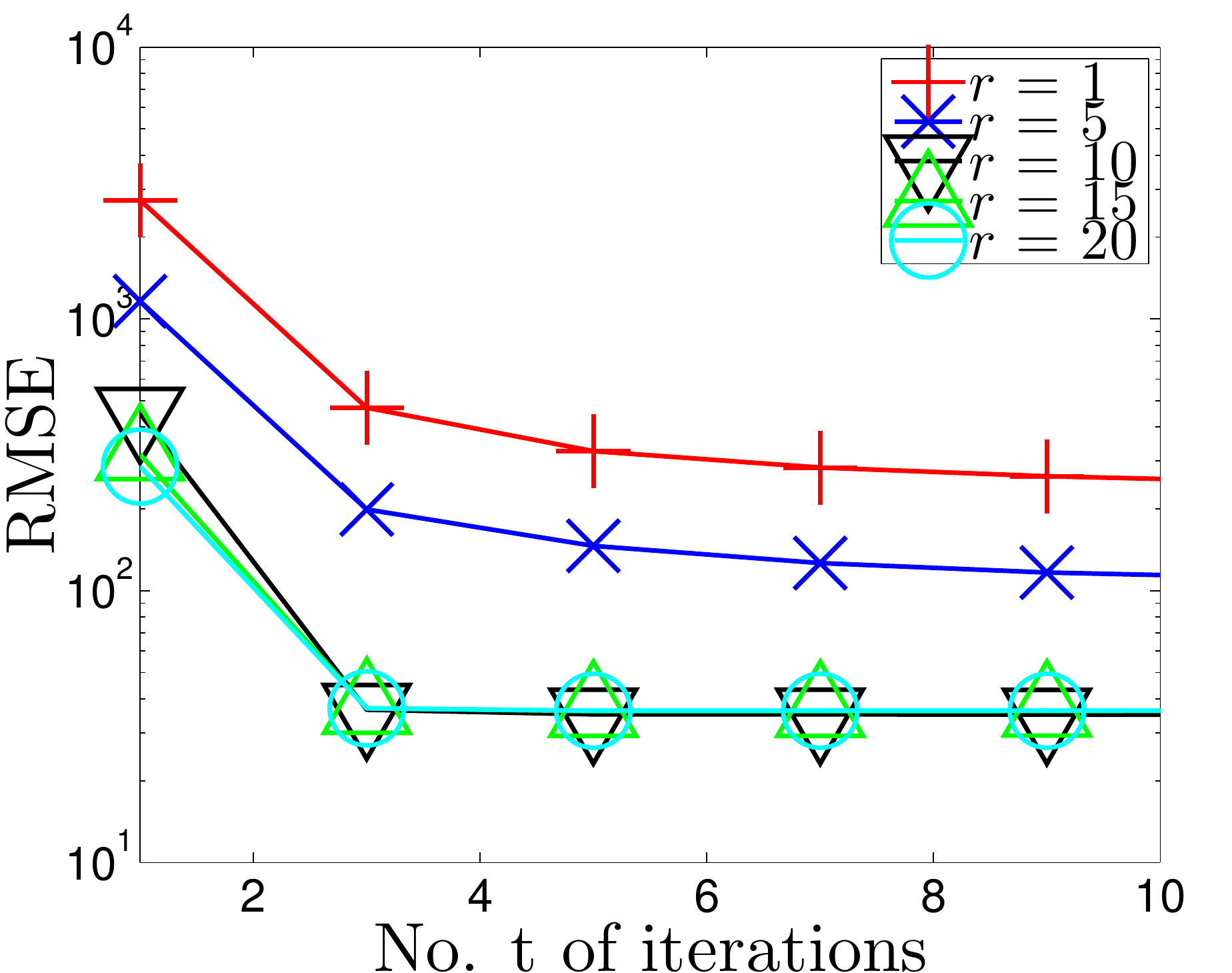} \vspace{-0.5mm}\\
			(a) & \hspace{-2mm}(b) \vspace{-3.5mm} 
		\end{tabular}
%	\end{small}
	\caption{Graphs of RMSEs achieved by $s$VBSSGP vs. number $t$ of iterations with varying number $r$ of samples drawn from $q_t(\boldsymbol{\alpha})$ to compute predictive mean $\widehat{\mu}_{\mathbf{x}_*}$ (Section~\ref{infer}) for the (a) AIMPEAK and (b) BLOG datasets.\vspace{-1mm}}
	\label{fig:bayesian}
\end{figure}

\noindent
{\bf AIRLINE Dataset.} This dataset is partitioned into $p = 2000$ disjoint subsets using $k$-means. Fig.~\ref{fig:compare}c and Table~\ref{tab:exp} show that $s$VBSSGP ($\gamma = 0$ and $r=5$) significantly outperforms VSSGP, DTC$+$, and PIC$+$ ($512$ inducing variables) in terms of RMSE after $45$ iterations, as explained previously.
Fig.~\ref{fig:compare}d shows that the total training time of $s$VBSSGP increases linearly with the number $t$ of iterations, which highlights a principled trade-off between its predictive performance and time efficiency. 	
The training time of $s$VBSSGP, though longer than DTC$+$ and PIC$+$, is only $43$~sec. per iteration; the training time of VSSGP is not included since its GitHub code runs on GPU instead of CPU.
The results of MNLP metric are similar to that for the AIMPEAK dataset.\vspace{1mm} 
%In terms of MNLP, we observe similar trend to the experiments on AIMPEAK dataset (Appendix~\ref{supp}).
\begin{table}
	%\vspace{1.5mm}
	\begin{small}
		%\hspace{1mm}
		\begin{tabular}{l|ccccc}
		\hline\vspace{-3mm}\\
			\hspace{-2.7mm} Dataset & \hspace{-1mm} $s$VBSSGP & \hspace{-1mm} DTC$+$ & \hspace{-1mm} PIC$+$ & \hspace{-1mm} SSGP & \hspace{-1mm}VSSGP \hspace{-2.7mm}\\ 
			\hline \vspace{-3mm} \\ %\noalign{\hrule height 0.8pt} 
			\hspace{-2mm}AIMPEAK &  \hspace{-1mm} {\bf 4.73} & \hspace{-1mm} $12.46$ & \hspace{-1mm} $6.15$ & \hspace{-1mm} $11.30$ & \hspace{-1mm} $9.46$ \\ 
			\hspace{-2mm}AIRLINE &  \hspace{-1mm} {\bf 22.18} & \hspace{-1mm} $39.60$ & \hspace{-1mm} $33.40$ & \hspace{-1mm} n/a & \hspace{-1mm} $39.49$ \\
		\hline 
		\end{tabular}
	\end{small}
	\vspace{-3mm}
	%\label{tab:exp}
	\caption{RMSEs achieved by $s$VBSSGP, DTC$+$, PIC$+$, SSGP \cite{Miguel10}, and VSSGP after final convergence for AIMPEAK and AIRLINE datasets.}\vspace{-5.1mm}
	%\end{tiny}http://www.tsc.uc3m.es/~miguel/code/ssgpr_code.zip
	\label{tab:exp}	
\end{table}

\noindent
{\bf BLOG Dataset.} This dataset is evenly partitioned into $p = 200$ disjoint subsets using $k$-means. Fig.~\ref{fig:bayesian} shows that with more samples drawn from $q_t(\boldsymbol{\alpha})$ to compute predictive mean $\widehat{\mu}_{\mathbf{x}_*}$ (Section~\ref{infer}), 
%Our experiments show that as we increase the number $r$ of samples drawn from $q(\boldsymbol{\alpha})$ for prediction (Section 4.2), 
$s$VBSSGP tends to converge faster and to a lower RMSE, which suggests a greater robustness to overfitting by exploiting a higher degree of Bayesian model averaging.
%This suggests that as we move away from a point-based approach ($n_{\boldsymbol{\alpha}}$ = 1) towards being Bayesian ($n_{\boldsymbol{\alpha}}$ = 20), the degree of overfitting will be reduced. 
More interestingly, the effect of overfitting appears to be more pronounced for the BLOG dataset with a much larger input dimension of $60$:
%when the data is high-dimensional: On the high dimensional BLOG dataset (60 attributes), 
When $r=1$, $s$VBSSGP effectively reduces to a local SSGP model utilizing the sampled spectral frequencies as its point estimate (see Proposition~\ref{lem:l1}, Remark $1$, and Section~\ref{infer}) and converges to the poorest performance that stops improving after $10$ iterations.
It can also be observed that the performance gap between $s$VBSSGP's with different number $r$ of samples is much wider at early iterations, thus highlighting the practicality of our Bayesian treatment in the case of limited data.
%the point-based variant of $s$VBSSGP ($n_{\boldsymbol{\alpha}}$ = 1) converges to a poor RMSE as compared to the rest, whereas on the low dimensional AIMPEAK dataset (6 attributes), all variants converge to the same RMSEs albeit with different rates. 
%It is also observable that the performace gap between $s$VBSSGP variants with different sample sizes is significantly more distinctive at early iterations, which highlights the fact that when data is limited, a Bayesian approach is more practical. 
\vspace{-2mm}
\section{Conclusion}
\label{conclude}\vspace{-0.5mm}
This paper describes a novel generalized framework of $s$VBSSGP regression models that addresses the shortcomings of existing sparse spectrum GP models like SSGP \cite{Miguel10} and VSSGP \cite{Yarin15} by adopting a Bayesian treatment of the spectral frequencies to avoid overfitting, modeling the spectral frequencies jointly in its variational distribution to enable their interaction \emph{a posteriori}, and exploiting local data for improving the predictive performance while still being able to preserve its scalability to big data through stochastic optimization.
As a result, empirical evaluation on real-world datasets (i.e., including the million-sized  benchmark AIRLINE dataset) shows that our proposed $s$VBSSGP regression model can significantly outperform the existing sparse spectrum GP models like SSGP and VSSGP as well as the stochastic implementations of the sparse GP models based on inducing variables like DTC$+$ and PIC$+$ \cite{Hensman13,NghiaICML15}.\vspace{0.5mm}

\noindent
{\bf Acknowledgments.}
This research is supported by the National Research Foundation, Prime Minister's Office, Singapore under its International Research Centre in Singapore Funding Initiative and Campus for Research Excellence and Technological Enterprise (CREATE) programme.\vspace{-1mm}Ê

\bibliographystyle{aaai}
\bibliography{aaai17}

\begin{thebibliography}{}

\bibitem[\protect\citeauthoryear{Buza}{2014}]{Buza14}
Buza, K.
\newblock 2014.
\newblock Feedback prediction for blogs.
\newblock In Spiliopoulou, M.; Schmidt-Thieme, L.; and Janning, R., eds., {\em
  Data Analysis, Machine Learning and Knowledge Discovery}. Springer
  International Publishing.
\newblock  145--152.

\bibitem[\protect\citeauthoryear{Cao, Low, and Dolan}{2013}]{LowAAMAS13}
Cao, N.; Low, K.~H.; and Dolan, J.~M.
\newblock 2013.
\newblock Multi-robot informative path planning for active sensing of
  environmental phenomena: A tale of two algorithms.
\newblock In {\em Proc. {AAMAS}}.

\bibitem[\protect\citeauthoryear{Chen \bgroup et al\mbox.\egroup
  }{2012}]{LowUAI12}
Chen, J.; Low, K.~H.; Tan, C. K.-Y.; Oran, A.; Jaillet, P.; Dolan, J.~M.; and
  Sukhatme, G.~S.
\newblock 2012.
\newblock Decentralized data fusion and active sensing with mobile sensors for
  modeling and predicting spatiotemporal traffic phenomena.
\newblock In {\em Proc. UAI},  163--173.

\bibitem[\protect\citeauthoryear{Chen \bgroup et al\mbox.\egroup
  }{2013}]{Chen13}
Chen, J.; Cao, N.; Low, K.~H.; Ouyang, R.; Tan, C. K.-Y.; and Jaillet, P.
\newblock 2013.
\newblock Parallel {G}aussian process regression with low-rank covariance
  matrix approximations.
\newblock In {\em Proc. UAI},  152--161.

\bibitem[\protect\citeauthoryear{Chen \bgroup et al\mbox.\egroup
  }{2015}]{TASE15}
Chen, J.; Low, K.~H.; Jaillet, P.; and Yao, Y.
\newblock 2015.
\newblock Gaussian process decentralized data fusion and active sensing for
  spatiotemporal traffic modeling and prediction in mobility-on-demand systems.
\newblock {\em {IEEE} T-ASE} 12(3):901--921.

\bibitem[\protect\citeauthoryear{Chen, Low, and Tan}{2013}]{LowRSS13}
Chen, J.; Low, K.~H.; and Tan, C. K.-Y.
\newblock 2013.
\newblock {Gaussian} process-based decentralized data fusion and active sensing
  for mobility-on-demand system.
\newblock In {\em Proc. {RSS}}.

\bibitem[\protect\citeauthoryear{Dolan \bgroup et al\mbox.\egroup
  }{2009}]{LowSPIE09}
Dolan, J.~M.; Podnar, G.; Stancliff, S.; Low, K.~H.; Elfes, A.; Higinbotham,
  J.; Hosler, J.~C.; Moisan, T.~A.; and Moisan, J.
\newblock 2009.
\newblock Cooperative aquatic sensing using the telesupervised adaptive ocean
  sensor fleet.
\newblock In {\em Proc. {SPIE} Conference on Remote Sensing of the Ocean, Sea
  Ice, and Large Water Regions}, volume 7473.

\bibitem[\protect\citeauthoryear{Gal and Turner}{2015}]{Yarin15}
Gal, Y., and Turner, R.
\newblock 2015.
\newblock Improving the {Gaussian} process sparse spectrum approximation by
  representing uncertainty in frequency inputs.
\newblock In {\em Proc. {ICML}},  655--664.

\bibitem[\protect\citeauthoryear{Hensman, Fusi, and Lawrence}{2013}]{Hensman13}
Hensman, J.; Fusi, N.; and Lawrence, N.~D.
\newblock 2013.
\newblock Gaussian processes for big data.
\newblock In {\em Proc. UAI},  282--290.

\bibitem[\protect\citeauthoryear{Hoang \bgroup et al\mbox.\egroup
  }{2014a}]{LowECML14b}
Hoang, T.~N.; Low, K.~H.; Jaillet, P.; and Kankanhalli, M.
\newblock 2014a.
\newblock Active learning is planning: Nonmyopic $\epsilon$-{Bayes}-optimal
  active learning of {Gaussian} processes.
\newblock In {\em Proc. {ECML/PKDD Nectar Track}},  494--498.

\bibitem[\protect\citeauthoryear{Hoang \bgroup et al\mbox.\egroup
  }{2014b}]{NghiaICML14}
Hoang, T.~N.; Low, K.~H.; Jaillet, P.; and Kankanhalli, M.
\newblock 2014b.
\newblock Nonmyopic $\epsilon$-{B}ayes-{O}ptimal {A}ctive {L}earning of
  {G}aussian {P}rocesses.
\newblock In {\em Proc. ICML},  739--747.

\bibitem[\protect\citeauthoryear{Hoang, Hoang, and Low}{2015}]{NghiaICML15}
Hoang, T.~N.; Hoang, Q.~M.; and Low, K.~H.
\newblock 2015.
\newblock A unifying framework of anytime sparse {Gaussian} process regression
  models with stochastic variational inference for big data.
\newblock In {\em Proc. {ICML}},  569--578.

\bibitem[\protect\citeauthoryear{Hoang, Hoang, and Low}{2016}]{HoangICML16}
Hoang, T.~N.; Hoang, Q.~M.; and Low, K.~H.
\newblock 2016.
\newblock A distributed variational inference framework for unifying parallel
  sparse {Gaussian} process regression models.
\newblock In {\em Proc. ICML},  382--391.

\bibitem[\protect\citeauthoryear{{L\'{a}zaro}-Gredilla \bgroup et
  al\mbox.\egroup }{2010}]{Miguel10}
{L\'{a}zaro}-Gredilla, M.; {Qui\~{n}onero}-Candela, J.; Rasmussen, C.~E.; and
  Figueiras-Vidal, A.~R.
\newblock 2010.
\newblock Sparse spectrum {G}aussian process regression.
\newblock {\em JMLR}  1865--1881.

\bibitem[\protect\citeauthoryear{Ling, Low, and Jaillet}{2016}]{LowAAAI16b}
Ling, C.~K.; Low, K.~H.; and Jaillet, P.
\newblock 2016.
\newblock {Gaussian} process planning with {Lipschitz} continuous reward
  functions: Towards unifying {Bayesian} optimization, active learning, and
  beyond.
\newblock In {\em Proc. {AAAI}},  1860--1866.

\bibitem[\protect\citeauthoryear{Low \bgroup et al\mbox.\egroup
  }{2012}]{LowAAMAS12}
Low, K.~H.; Chen, J.; Dolan, J.~M.; Chien, S.; and Thompson, D.~R.
\newblock 2012.
\newblock Decentralized active robotic exploration and mapping for
  probabilistic field classification in environmental sensing.
\newblock In {\em Proc. {AAMAS}},  105--112.

\bibitem[\protect\citeauthoryear{Low \bgroup et al\mbox.\egroup
  }{2014a}]{LowDyDESS14}
Low, K.~H.; Chen, J.; Hoang, T.~N.; Xu, N.; and Jaillet, P.
\newblock 2014a.
\newblock Recent advances in scaling up {Gaussian} process predictive models
  for large spatiotemporal data.
\newblock In {\em Proc. {DyDESS}}.

\bibitem[\protect\citeauthoryear{Low \bgroup et al\mbox.\egroup
  }{2014b}]{LowECML14a}
Low, K.~H.; Xu, N.; Chen, J.; Lim, K.~K.; and {\"{O}zg\"{u}l}, E.~B.
\newblock 2014b.
\newblock Generalized online sparse {Gaussian} processes with application to
  persistent mobile robot localization.
\newblock In {\em Proc. {ECML/PKDD Nectar Track}},  499--503.

\bibitem[\protect\citeauthoryear{Low \bgroup et al\mbox.\egroup
  }{2015}]{LowAAAI15}
Low, K.~H.; Yu, J.; Chen, J.; and Jaillet, P.
\newblock 2015.
\newblock Parallel {Gaussian} process regression for big data: Low-rank
  representation meets {M}arkov approximation.
\newblock In {\em Proc. {AAAI}}.

\bibitem[\protect\citeauthoryear{Low, Dolan, and Khosla}{2008}]{LowAAMAS08}
Low, K.~H.; Dolan, J.~M.; and Khosla, P.
\newblock 2008.
\newblock Adaptive multi-robot wide-area exploration and mapping.
\newblock In {\em Proc. {AAMAS}},  23--30.

\bibitem[\protect\citeauthoryear{Low, Dolan, and Khosla}{2009}]{LowICAPS09}
Low, K.~H.; Dolan, J.~M.; and Khosla, P.
\newblock 2009.
\newblock Information-theoretic approach to efficient adaptive path planning
  for mobile robotic environmental sensing.
\newblock In {\em Proc. {ICAPS}}.

\bibitem[\protect\citeauthoryear{Low, Dolan, and Khosla}{2011}]{LowAAMAS11}
Low, K.~H.; Dolan, J.~M.; and Khosla, P.
\newblock 2011.
\newblock Active {Markov} information-theoretic path planning for robotic
  environmental sensing.
\newblock In {\em Proc. {AAMAS}},  753--760.

\bibitem[\protect\citeauthoryear{Ouyang \bgroup et al\mbox.\egroup
  }{2014}]{LowAAMAS14}
Ouyang, R.; Low, K.~H.; Chen, J.; and Jaillet, P.
\newblock 2014.
\newblock Multi-robot active sensing of non-stationary {Gaussian} process-based
  environmental phenomena.
\newblock In {\em Proc. {AAMAS}}.

\bibitem[\protect\citeauthoryear{Podnar \bgroup et al\mbox.\egroup
  }{2010}]{LowAeroconf10}
Podnar, G.; Dolan, J.~M.; Low, K.~H.; and Elfes, A.
\newblock 2010.
\newblock Telesupervised remote surface water quality sensing.
\newblock In {\em Proc. {IEEE} Aerospace Conference}.

\bibitem[\protect\citeauthoryear{{Qui\~{n}onero}-Candela and
  Rasmussen}{2005}]{Candela05}
{Qui\~{n}onero}-Candela, J., and Rasmussen, C.~E.
\newblock 2005.
\newblock A unifying view of sparse approximate {Gaussian} process regression.
\newblock {\em Journal of Machine Learning Research} 6:1939--1959.

\bibitem[\protect\citeauthoryear{Titsias and
  {L\'{a}zaro}-Gredilla}{2014}]{Titsias14}
Titsias, M.~K., and {L\'{a}zaro}-Gredilla, M.
\newblock 2014.
\newblock Doubly stochastic variational {Bayes} for non-conjugate inference.
\newblock In {\em Proc. {ICML}},  1971--1979.

\bibitem[\protect\citeauthoryear{Titsias}{2009}]{Titsias09}
Titsias, M.~K.
\newblock 2009.
\newblock Variational learning of inducing variables in sparse {G}aussian
  processes.
\newblock In {\em Proc. {AISTATS}}.

\bibitem[\protect\citeauthoryear{Xu \bgroup et al\mbox.\egroup
  }{2014}]{LowAAAI14}
Xu, N.; Low, K.~H.; Chen, J.; Lim, K.~K.; and {\"{O}zg\"{u}l}, E.~B.
\newblock 2014.
\newblock {GP-Localize}: Persistent mobile robot localization using online
  sparse {Gaussian} process observation model.
\newblock In {\em Proc. {AAAI}},  2585--2592.

\bibitem[\protect\citeauthoryear{Yu \bgroup et al\mbox.\egroup
  }{2012}]{LowIAT12}
Yu, J.; Low, K.~H.; Oran, A.; and Jaillet, P.
\newblock 2012.
\newblock Hierarchical {Bayesian} nonparametric approach to modeling and
  learning the wisdom of crowds of urban traffic route planning agents.
\newblock In {\em Proc. {IAT}},  478--485.

\bibitem[\protect\citeauthoryear{Zhang \bgroup et al\mbox.\egroup
  }{2016}]{LowAAAI16a}
Zhang, Y.; Hoang, T.~N.; Low, K.~H.; and Kankanhalli, M.
\newblock 2016.
\newblock Near-optimal active learning of multi-output {Gaussian} processes.
\newblock In {\em Proc. {AAAI}},  2351--2357.

\end{thebibliography}

\if \myproof1	
\clearpage

\appendix
\section{Proof of Proposition~\ref{lem:l1}}
\label{app:a}
Since $\{f_\mathbf{x}\}_{\mathbf{x} \in \mathcal{X}}$ denotes a zero-mean Gaussian process with kernel $k(\mathbf{x},\mathbf{x}')$ and the noisy output $y_\mathbf{x} = f_\mathbf{x} + \epsilon$ is generated by perturbing $f_\mathbf{x}$ with a random noise $\epsilon \sim \mathcal{N}(0, \sigma_n^2)$ (Section~\ref{hgp}), $p(f_{\mathbf{x}_\ast} | \mathbf{y}_k, {\boldsymbol{\theta}}) = \mathcal{N}(\mathbb{E}[f_{\mathbf{x}_\ast} | \mathbf{y}_k,{\boldsymbol{\theta}}], \mathbb{V}[f_{\mathbf{x}_\ast} | \mathbf{y}_k,{\boldsymbol{\theta}}])$ for $\mathbf{x}_\ast \in \mathcal{X}_k$ where %\vspace{1mm}{\mathbf{x}_\ast}\hspace{-0.7mm}
\begin{equation}
\hspace{-1.7mm}\begin{array}{rcl}
\mathbb{E}[f_{\mathbf{x}_\ast} | \mathbf{y}_k,{\boldsymbol{\theta}}] &\hspace{-2.4mm}\triangleq &\hspace{-2.4mm} \displaystyle\mathbf{K}(\mathbf{x}_\ast, \mathbf{X}_k)\boldsymbol{\Xi}_k\mathbf{y}_k \ ,\vspace{0.5mm} \\
\mathbb{V}[f_{\mathbf{x}_\ast} | \mathbf{y}_k,{\boldsymbol{\theta}}] &\hspace{-2.4mm}\triangleq &\hspace{-2.4mm} \displaystyle k(\mathbf{x_\ast},\mathbf{x_\ast}) -  \mathbf{K}(\mathbf{x_\ast},\mathbf{X}_k)\boldsymbol{\Xi}_k\mathbf{K}(\mathbf{X}_k,\mathbf{x}_\ast)\ , 
%\vspace{2mm}
\end{array}\hspace{-5mm}
\label{a0}
\end{equation}
%\begin{eqnarray}
%p\left(f_{\mathbf{x}_\ast} | \mathbf{y}_k, {\boldsymbol{\theta}}\right) = \mathcal{N}\left(f_{\mathbf{x}_\ast} \Big| \mathbf{K}(\mathbf{x}_\ast, \mathbf{X}_k)\boldsymbol{\Gamma}_k\mathbf{y}_k, k(\mathbf{x_\ast},\mathbf{x_\ast}) -  \mathbf{K}(\mathbf{x_\ast},\mathbf{X}_k)\boldsymbol{\Gamma}_k\mathbf{K}(\mathbf{X}_k,\mathbf{x}_\ast)\right) \ , \label{a0}
%\end{eqnarray}
and $\boldsymbol{\Xi}_k \triangleq (\mathbf{K}(\mathbf{X}_k,\mathbf{X}_k) + \sigma_n^2\mathbf{I})^{-1}$, which is essentially the standard GP predictive distribution of $f_{\mathbf{x}_\ast}$ given the noisy outputs $\mathbf{y}_k$ for the training inputs $\mathbf{X}_k$ and the hyperparameters ${\boldsymbol{\theta}}$.
% \cite{Rasmussen06}. 
Using~\eqref{eq:kernel},
$$
\hspace{-1.7mm}
\begin{array}{l}
\boldsymbol{\Xi}_k \\
\displaystyle = (\boldsymbol{\Phi}_{\boldsymbol{\theta}}^\top(\mathbf{X}_k)\boldsymbol{\Lambda}\boldsymbol{\Phi}_{\boldsymbol{\theta}}(\mathbf{X}_k) + \sigma_n^2\mathbf{I})^{-1}\vspace{0.5mm}\\
\displaystyle = \sigma_n^{-2}\mathbf{I}\hspace{-0.5mm} -\hspace{-0.5mm} \sigma_n^{-4}\boldsymbol{\Phi}_{\boldsymbol{\theta}}^\top(\mathbf{X}_k)(\sigma_n^{-2}\boldsymbol{\Phi}_{\boldsymbol{\theta}}(\mathbf{X}_k)\boldsymbol{\Phi}_{\boldsymbol{\theta}}^\top(\mathbf{X}_k)\hspace{-0.5mm}+\hspace{-0.5mm}\boldsymbol{\Lambda}^{-1})^{-1}\boldsymbol{\Phi}_{\boldsymbol{\theta}}(\mathbf{X}_k)\\
\displaystyle = \sigma_n^{-2}(\mathbf{I} - \boldsymbol{\Phi}_{\boldsymbol{\theta}}^\top(\mathbf{X}_k)\boldsymbol{\Gamma}^{-1}_k\boldsymbol{\Phi}_{\boldsymbol{\theta}}(\mathbf{X}_k))
\end{array}
$$
where the second and third equalities are due to the matrix inversion lemma and the definition of $\boldsymbol{\Gamma}_k$, respectively. 
Using~\eqref{eq:kernel},~\eqref{a0} can also be rewritten as
%Then, exploiting the fact that $k(\mathbf{x},\mathbf{x}') = \boldsymbol{\phi}^\top_{\boldsymbol{\theta}}(\mathbf{x})\boldsymbol{\Lambda}\boldsymbol{\phi}(\mathbf{x})$ (see Section~\ref{hgp}), we can rewrite Eq.~\eqref{a0} as 
$$
\begin{array}{rcl}
\mathbb{E}\left[f_{\mathbf{x}_\ast} | \mathbf{y}_k,{\boldsymbol{\theta}}\right] &\hspace{-2.4mm} =&\hspace{-2.4mm} \displaystyle \gamma\mathbf{H}\boldsymbol{\mu} + (1 - \gamma)\mathbf{H}\boldsymbol{\mu}\\
\mathbb{V}\left[f_{\mathbf{x}_\ast} | \mathbf{y}_k,{\boldsymbol{\theta}}\right] &\hspace{-2.4mm} =&\hspace{-2.4mm} \displaystyle (1 - \gamma^2)\mathbf{H}\boldsymbol{\Sigma}\mathbf{H}^\top + \gamma^2\mathbf{H}\boldsymbol{\Sigma}\mathbf{H}^\top
\end{array}
$$
 for all $|\gamma| \leq 1$ where $\mathbf{H}\triangleq\boldsymbol{\phi}_{\boldsymbol{\theta}}^\top(\mathbf{x}_\ast)$, 
$$
\begin{array}{rcl}
\boldsymbol{\mu}&\hspace{-2.4mm}\triangleq&\hspace{-2.4mm}\displaystyle\boldsymbol{\Lambda}\boldsymbol{\Phi}_{\boldsymbol{\theta}}(\mathbf{X}_k)\boldsymbol{\Xi}_k\mathbf{y}_k\\
&\hspace{-2.4mm} = &\hspace{-2.4mm} \displaystyle\sigma_n^{-2}\boldsymbol{\Lambda}(\mathbf{I} - \boldsymbol{\Phi}_{\boldsymbol{\theta}}(\mathbf{X}_k)\boldsymbol{\Phi}_{\boldsymbol{\theta}}^\top(\mathbf{X}_k)\boldsymbol{\Gamma}^{-1}_k)\boldsymbol{\Phi}_{\boldsymbol{\theta}}(\mathbf{X}_k) \mathbf{y}_k\\
&\hspace{-2.4mm} = &\hspace{-2.4mm} \displaystyle\sigma_n^{-2}\boldsymbol{\Lambda}(\mathbf{I} - (\boldsymbol{\Gamma}_k-\sigma^2_n\boldsymbol{\Lambda}^{-1})\boldsymbol{\Gamma}^{-1}_k)\boldsymbol{\Phi}_{\boldsymbol{\theta}}(\mathbf{X}_k) \mathbf{y}_k\\
&\hspace{-2.4mm} = &\hspace{-2.4mm} \displaystyle\boldsymbol{\Gamma}^{-1}_k \boldsymbol{\Phi}_{\boldsymbol{\theta}}(\mathbf{X}_k) \mathbf{y}_k
\end{array}
$$ 
where the third equality is due to the definition of $\boldsymbol{\Gamma}_k$, 
and 
$$
\begin{array}{rcl}
\mathbf{\Sigma}  &\hspace{-2.4mm}\triangleq&\hspace{-2.4mm}\displaystyle \boldsymbol{\Lambda} - \boldsymbol{\Lambda}\boldsymbol{\Phi}_{\boldsymbol{\theta}}(\mathbf{X}_k)\boldsymbol{\Xi}_k\boldsymbol{\Phi}_{\boldsymbol{\theta}}^\top(\mathbf{X}_k)\boldsymbol{\Lambda}\\
&\hspace{-2.4mm} = &\hspace{-2.4mm} \displaystyle\boldsymbol{\Lambda} - 
\boldsymbol{\Gamma}^{-1}_k \boldsymbol{\Phi}_{\boldsymbol{\theta}}(\mathbf{X}_k)\boldsymbol{\Phi}_{\boldsymbol{\theta}}^\top(\mathbf{X}_k)\boldsymbol{\Lambda}\\
&\hspace{-2.4mm} = &\hspace{-2.4mm} \displaystyle\boldsymbol{\Lambda} - 
\boldsymbol{\Gamma}^{-1}_k (\boldsymbol{\Gamma}_k-\sigma^2_n\boldsymbol{\Lambda}^{-1})\boldsymbol{\Lambda}\\
&\hspace{-2.4mm} = &\hspace{-2.4mm} \displaystyle\sigma^2_n \boldsymbol{\Gamma}^{-1}_k\ .
\end{array}
$$ 
Using the Gaussian identity of affine transformation for marginalization, %it follows that
\begin{equation}
p\left(f_{\mathbf{x}_\ast} | \mathbf{y}_k, {\boldsymbol{\theta}}\right) =\hspace{-1mm} \int_\mathbf{s} \mathcal{N}\hspace{-0.7mm}\left(f_{\mathbf{x}_\ast} | \mu_{\mathbf{x}_\ast}\hspace{-0.7mm}(\boldsymbol{\alpha}), \sigma^2_{\mathbf{x}_\ast}\hspace{-0.7mm}(\boldsymbol{\alpha})\right)\mathcal{N}(\mathbf{s}|\boldsymbol{\mu},\boldsymbol{\Sigma})\ \mathrm{d}\mathbf{s} 
\label{a:a1}
%\mathbb{E}_{\mathbf{s}}\left[\mathcal{N}\left(f_{\mathbf{x}_\ast}\Big | \gamma\mathbf{H}\mathbf{s} + (1 - \gamma)\mathbf{H}\boldsymbol{\mu}, (1 - \gamma^2)\mathbf{H}\boldsymbol{\Sigma}\mathbf{H}^\top\right)\right] \label{a:a1}
\end{equation}
where $\mu_{\mathbf{x}_\ast}\hspace{-0.7mm}(\boldsymbol{\alpha}) \triangleq \gamma\mathbf{H}\mathbf{s} + (1 - \gamma)\mathbf{H}\boldsymbol{\mu}$ and $\sigma^2_{\mathbf{x}_\ast}\hspace{-0.7mm}(\boldsymbol{\alpha}) \triangleq (1 - \gamma^2)\mathbf{H}\boldsymbol{\Sigma}\mathbf{H}^\top$. Also, marginalizing out 
$[f_{\mathbf{x}}]^{\top}_{\mathbf{x}\in\mathbf{X}\setminus\mathbf{X}_k}$
%$\mathbf{f} \setminus \mathbf{f}_k$ 
from $p(\mathbf{f}, \mathbf{s} |{\boldsymbol{\theta}})$ in~\eqref{eq:imagine} (Section~\ref{hgp}) yields
$$
\left[\hspace{-1mm}
\begin{array}{l}
\mathbf{s} \\
\mathbf{f}_k
\end{array}\hspace{-1mm}
\right]
\sim
\mathcal{N}\hspace{-0.7mm}\left(
\left[\hspace{-1mm}
\begin{array}{l}
\mathbf{0}\\
\mathbf{0}
\end{array}\hspace{-1mm}
\right],
\left[\hspace{-1mm}
\begin{array}{cc}
\boldsymbol{\Lambda} & \boldsymbol{\Lambda}\boldsymbol{\Phi}_{\boldsymbol{\theta}}(\mathbf{X}_k) \\
\boldsymbol{\Phi}^\top_{\boldsymbol{\theta}}(\mathbf{X}_k)\boldsymbol{\Lambda} & \mathbf{K}(\mathbf{X}_k, \mathbf{X}_k) 
\end{array}\hspace{-1mm}
\right]
\right) 
$$
where $\mathbf{f}_k \triangleq[f_{\mathbf{x}}]^{\top}_{\mathbf{x}\in\mathbf{X}_k}$.
Consequently,
$$
\left[\hspace{-1mm}
\begin{array}{l}
\mathbf{s} \\
\mathbf{y}_k
\end{array}\hspace{-1mm}
\right]
\sim
\mathcal{N}\hspace{-0.7mm}\left(
\left[\hspace{-1mm}
\begin{array}{l}
\mathbf{0}\\
\mathbf{0}
\end{array}\hspace{-1mm}
\right],
\left[\hspace{-1mm}
\begin{array}{cc}
\boldsymbol{\Lambda} & \boldsymbol{\Lambda}\boldsymbol{\Phi}_{\boldsymbol{\theta}}(\mathbf{X}_k) \\
\boldsymbol{\Phi}^\top_{\boldsymbol{\theta}}(\mathbf{X}_k)\boldsymbol{\Lambda} & \boldsymbol{\Xi}^{-1}_k
\end{array}\hspace{-1mm}
\right]
\right) ,
$$
which allows us to derive an analytic expression for $p(\mathbf{s}| \mathbf{y}_k, {\boldsymbol{\theta}}) = \mathcal{N}(\mathbf{s} | \boldsymbol{\mu},\boldsymbol{\Sigma})$ where $\boldsymbol{\mu}$ and $\boldsymbol{\Sigma}$ are defined as above. Plugging this expression into~\eqref{a:a1} yields 
\begin{equation}
p\left(f_{\mathbf{x}_\ast} | \mathbf{y}_k, {\boldsymbol{\theta}}\right) = \int_\mathbf{s} \mathcal{N}\hspace{-0.7mm}\left(f_{\mathbf{x}_\ast} | \mu_{\mathbf{x}_\ast}\hspace{-0.7mm}(\boldsymbol{\alpha}), \sigma^2_{\mathbf{x}_\ast}\hspace{-0.7mm}(\boldsymbol{\alpha})\right)p(\mathbf{s}| \mathbf{y}_k, {\boldsymbol{\theta}})\ \mathrm{d}\mathbf{s}\ . 
\label{a2}
\end{equation}
%\begin{eqnarray}
%\hspace{-11mm}p\left(f_\mathbf{x_\ast} | \mathbf{y}_k,{\boldsymbol{\theta}}\right) &=& 
%\int_\mathbf{s} \mathcal{N}\left(f_{\mathbf{x}_\ast} | \gamma\mathbf{H}\mathbf{s} + (1-\gamma)\mathbf{H}\boldsymbol{\mu}, (1-\gamma^2)\mathbf{H}\boldsymbol{\Sigma}\mathbf{H}^\top \right)
%p(\mathbf{s}|\mathbf{y}_k,{\boldsymbol{\theta}})\mathrm{d}\mathbf{s} \ . \label{a2}
%\end{eqnarray}
Alternatively, $p(f_{\mathbf{x}_\ast} | \mathbf{y}_k,{\boldsymbol{\theta}})$ can be expressed in terms of $p(f_{\mathbf{x}_\ast} | \mathbf{y}_k, \mathbf{s}, {\boldsymbol{\theta}})$ and $p(\mathbf{s}|\mathbf{y}_k, {\boldsymbol{\theta}})$ using marginalization:
\begin{equation}
p\left(f_{\mathbf{x}_\ast} | \mathbf{y}_k, {\boldsymbol{\theta}}\right) = \int_\mathbf{s} p\left(f_{\mathbf{x}_\ast} | \mathbf{y}_k, \mathbf{s}, {\boldsymbol{\theta}}\right) p(\mathbf{s} | \mathbf{y}_k, {\boldsymbol{\theta}})\ \mathrm{d}\mathbf{s} \ . 
\label{a3}
\end{equation}
Then, subtracting~\eqref{a2} from~\eqref{a3} gives
$$
0 = \mathbb{E}_{\mathbf{s} \sim p(\mathbf{s} | \mathbf{y}_k, {\boldsymbol{\theta}})} 
\hspace{-0.7mm}\Big[
p(f_\mathbf{x_\ast} | \mathbf{y}_k, \mathbf{s}, {\boldsymbol{\theta}}) - \mathcal{N}\hspace{-0.7mm}\left(f_{\mathbf{x}_\ast} | \mu_{\mathbf{x}_\ast}\hspace{-0.7mm}(\boldsymbol{\alpha}), \sigma^2_{\mathbf{x}_\ast}\hspace{-0.7mm}(\boldsymbol{\alpha})\right)\Big]. 
%\nonumber
%\label{a4}
$$
Since $p(\mathbf{s}|\mathbf{y}_k, {\boldsymbol{\theta}}) > 0$ for all $\mathbf{s}$, the above equation suggests that $p(f_\mathbf{x_\ast} | \mathbf{y}_k, \mathbf{s}, {\boldsymbol{\theta}}) = \mathcal{N}\hspace{-0.7mm}\left(f_{\mathbf{x}_\ast} | \mu_{\mathbf{x}_\ast}\hspace{-0.7mm}(\boldsymbol{\alpha}), \sigma^2_{\mathbf{x}_\ast}\hspace{-0.7mm}(\boldsymbol{\alpha})\right)$ is a valid test conditional which is consistent with the structural assumption in~\eqref{eq:imagine}. 
Finally, plugging in the above definitions of $\mathbf{H}$, $\boldsymbol{\mu}$, and $\mathbf{\Sigma}$ reproduces the definitions of $\mu_{\mathbf{x}_\ast}\hspace{-0.7mm}(\boldsymbol{\alpha})$ and $\sigma^2_{\mathbf{x}_\ast}\hspace{-0.7mm}(\boldsymbol{\alpha})$ in Proposition~\ref{lem:l1}, thus completing our proof.

\section{Derivation of $L(q)$}
\label{app:b}
%Let $\boldsymbol{\alpha} \triangleq ({\boldsymbol{\theta}},\mathbf{s})$ as defined in Lemma~\ref{lem:l2}. 
For all $\boldsymbol{\alpha}$, $p(\mathbf{y}) = p(\boldsymbol{\alpha},\mathbf{y}) / p(\boldsymbol{\alpha}|\mathbf{y})$ which implies
\begin{equation}
\log p(\mathbf{y}) = \log \frac{p(\boldsymbol{\alpha},\mathbf{y})}{p(\boldsymbol{\alpha}|\mathbf{y})} \ .
\label{b1}
\end{equation} 
Then, let $q(\boldsymbol{\alpha})$ denotes an arbitrary probability density function of $\boldsymbol{\alpha}$ (i.e., $\int_{\boldsymbol{\alpha}} q(\boldsymbol{\alpha})\mathrm{d}\boldsymbol{\alpha} = 1$). Integrating both sides of~\eqref{b1} with $q(\boldsymbol{\alpha})$ gives
\begin{equation}
\log p(\mathbf{y}) = \int_{\boldsymbol{\alpha}} q(\boldsymbol{\alpha})\log\frac{p(\boldsymbol{\alpha}, \mathbf{y})}{p(\boldsymbol{\alpha}|\mathbf{y})}\mathrm{d}\boldsymbol{\alpha} \ . 
\label{b2}
\end{equation}
Then, plugging 
$$\log\frac{p(\boldsymbol{\alpha},\mathbf{y})}{p(\boldsymbol{\alpha}|\mathbf{y})} = \log\frac{p(\boldsymbol{\alpha},\mathbf{y})}{q(\boldsymbol{\alpha})} + \log\frac{q(\boldsymbol{\alpha})}{p(\boldsymbol{\alpha}|\mathbf{y})}$$ 
into the RHS of~\eqref{b2} yields
\begin{equation}
\begin{array}{rcl}
\log p(\mathbf{y}) &\hspace{-2.4mm}=&\hspace{-2.4mm}\displaystyle \int_{\boldsymbol{\alpha}}q(\boldsymbol{\alpha})\log\frac{p(\boldsymbol{\alpha},\mathbf{y})}{q(\boldsymbol{\alpha})}\mathrm{d}\boldsymbol{\alpha} + \mathrm{KL}\left(q(\boldsymbol{\alpha})\|p(\boldsymbol{\alpha}|\mathbf{y})\right) \\
&\hspace{-2.4mm}=&\hspace{-2.4mm}\displaystyle \mathbb{E}_{\boldsymbol{\alpha}\sim q(\boldsymbol{\alpha})}\hspace{-1mm}\left[\log\frac{p(\boldsymbol{\alpha},\mathbf{y})}{q(\boldsymbol{\alpha})}\right] + {D}_\mathrm{KL}(q) 
\end{array}
\label{b3}
\end{equation}
where the second equality is due to the definition of ${D}_\mathrm{KL}(q)$ in Section~\ref{vbgpr}. Finally, plugging $p(\boldsymbol{\alpha},\mathbf{y}) = p(\mathbf{y}|\boldsymbol{\alpha})\ p(\boldsymbol{\alpha})$ into~\eqref{b3} yields
$$
\begin{array}{rcl}
\log p(\mathbf{y}) &\hspace{-2.4mm}=&\hspace{-2.4mm}\displaystyle \mathbb{E}_{\boldsymbol{\alpha} \sim q(\boldsymbol{\alpha})}\hspace{-1mm}\left[\log p(\mathbf{y} |\boldsymbol{\alpha}) - \log\frac{q(\boldsymbol{\alpha})}{p(\boldsymbol{\alpha})}\right] +{D}_\mathrm{KL}(q) \\
&\hspace{-2.4mm}=&\hspace{-2.4mm}\displaystyle {L}(q) + {D}_\mathrm{KL}(q) \ ,
%\label{b4}
\end{array}
$$
which concludes our proof.
\section{Reparameterizing $L(q)$ via $\psi(\mathbf{z})$}
\label{app:c}
To do this, let us first define the following function:
\begin{equation}
g(\boldsymbol{\alpha}) \triangleq q(\boldsymbol{\alpha})\left[\log p(\mathbf{y} |\boldsymbol{\alpha}) - \log \frac{q(\boldsymbol{\alpha})}{p(\boldsymbol{\alpha})}\right]  , \label{c1}
\end{equation}
which allows us to re-express ${L}(q)$ as
\begin{equation}
{L}(q) = \int_{\boldsymbol{\alpha}} g(\boldsymbol{\alpha})\ \mathrm{d}\boldsymbol{\alpha}\ .\label{c2}
\end{equation}
Then, applying the \emph{change of variables} theorem to the RHS of~\eqref{c2} with respect to a sufficiently well-behaved function $\boldsymbol{\alpha} \triangleq \varphi(\mathbf{z})$, which transforms a variable $\mathbf{z}$ into $\boldsymbol{\alpha}$, gives
\begin{equation}
{L}(q) = \int_\mathbf{z} g(\varphi(\mathbf{z}))\left|{J}_\varphi(\mathbf{z})\right|\mathrm{d}\mathbf{z} = \int_\mathbf{z} g(\boldsymbol{\alpha})\left|{J}_\varphi(\mathbf{z})\right|\mathrm{d}\mathbf{z} 
\label{c3}
\end{equation} 
where 
%$\boldsymbol{\alpha} = \varphi(\mathbf{z})$ and 
$|{J}_\varphi(\mathbf{z})|$ denotes the absolute value of $\mathrm{det}({J}_\varphi(\mathbf{z})) \triangleq \mathrm{det}(\partial\varphi(\mathbf{z})/\partial\mathbf{z})$. 

So, choosing $\varphi(\mathbf{z}) \triangleq \mathbf{M}\mathbf{z} + \mathbf{b}$ yields $|{J}_\varphi(\mathbf{z})| = |\mathbf{M}|$ which can be plugged into~\eqref{c3} to give
\begin{equation}
\begin{array}{rcl}
{L}(q) &\hspace{-2.4mm}=&\hspace{-2.4mm} \displaystyle\int_\mathbf{z} g(\boldsymbol{\alpha})\left|\mathbf{M}\right|\mathrm{d}\mathbf{z} \\
&\hspace{-2.4mm}=&\hspace{-2.4mm} \displaystyle\int_\mathbf{z}q(\boldsymbol{\alpha})\left|\mathbf{M}\right|\left[\log p(\mathbf{y} |\boldsymbol{\alpha}) - \log \frac{q(\boldsymbol{\alpha})}{p(\boldsymbol{\alpha})}\right]\mathrm{d}\mathbf{z} 
\end{array}
\label{c4}
\end{equation}
where the last equality follows from our definition of $g(\boldsymbol{\alpha})$ in~\eqref{c1}. Now, using the parameterization in Lemma~\ref{lem:l2}, $q(\boldsymbol{\alpha}) =  \psi(\mathbf{z})/|\mathbf{M}|$ or, equivalently, $\psi(\mathbf{z}) = q(\boldsymbol{\alpha})|\mathbf{M}|$. Thus, plugging this into~\eqref{c4} yields
$$
\begin{array}{rcl}
{L}(q) &\hspace{-2.4mm}=& \hspace{-2.4mm} \displaystyle\int_\mathbf{z} \psi\left(\mathbf{z}\right) \left[\log p(\mathbf{y} |\boldsymbol{\alpha}) - \log \frac{q(\boldsymbol{\alpha})}{p(\boldsymbol{\alpha})}\right]\mathrm{d}\mathbf{z} \vspace{1mm}\\
&\hspace{-2.4mm}=& \hspace{-2.4mm} \displaystyle\mathbb{E}_{\mathbf{z}\sim\psi(\mathbf{z})}\hspace{-1mm}\left[\log p(\mathbf{y} |\boldsymbol{\alpha}) - \log \frac{q(\boldsymbol{\alpha})}{p(\boldsymbol{\alpha})}\right] \ , 
%\label{c5}
\end{array}
$$
which completes our proof.
\section{Closed-Form Evaluation of $\partial\widehat{{L}}/\partial{\boldsymbol{\eta}}$}
%$\partial(\log q(\boldsymbol{\alpha})/p(\boldsymbol{\alpha}))/\partial\boldsymbol{\eta}$
\label{app:d}
To show that $\partial\widehat{{L}}/\partial\boldsymbol{\eta}$ is analytically tractable, it suffices to show that both $\partial\log p(\mathbf{y}|\boldsymbol{\alpha})/\partial\boldsymbol{\eta}$ and $\partial\log (q(\boldsymbol{\alpha})/p(\boldsymbol{\alpha}))/\partial\boldsymbol{\eta}$ are analytically tractable, the former of which is shown in Appendix~\ref{app:f}. To show the latter,
%that $\partial(\log q(\boldsymbol{\alpha})/p(\boldsymbol{\alpha}))/\partial\boldsymbol{\eta}$ is tractable, 
note that $\boldsymbol{\eta} = \mathrm{vec}(\mathbf{M},\mathbf{b})$ (Section~\ref{parameterisation}) implies
$$
\frac{\partial}{\partial\boldsymbol{\eta}}\log\frac{q(\boldsymbol{\alpha})}{p(\boldsymbol{\alpha})} = \mathrm{vec}\hspace{-0.7mm}\left(\frac{\partial}{\partial\mathbf{M}}\log\frac{q(\boldsymbol{\alpha})}{p(\boldsymbol{\alpha})}, \frac{\partial}{\partial\mathbf{b}}\log\frac{q(\boldsymbol{\alpha})}{p(\boldsymbol{\alpha})}\right) ,
$$
which reveals that $\partial\log (q(\boldsymbol{\alpha})/p(\boldsymbol{\alpha}))/\partial\boldsymbol{\eta}$ can be analytically evaluated if $\partial\log (q(\boldsymbol{\alpha})/p(\boldsymbol{\alpha}))/\partial\mathbf{M}$ and $\partial\log (q(\boldsymbol{\alpha})/p(\boldsymbol{\alpha}))/\partial\mathbf{b}$ can be analytically evaluated, as detailed below.

To evaluate the derivative $\partial\log (q(\boldsymbol{\alpha})/p(\boldsymbol{\alpha}))/\partial\mathbf{M}$, note that $\partial\log (q(\boldsymbol{\alpha})/p(\boldsymbol{\alpha}))/\partial\mathbf{M} = \partial\log q(\boldsymbol{\alpha})/\partial\mathbf{M} - \partial\log p(\boldsymbol{\alpha})/\partial\mathbf{M}$. Then, using the parameterization of $q(\boldsymbol{\alpha}) \triangleq\psi(\mathbf{z})/ |\mathbf{M}|$ in Lemma~\ref{lem:l2}, it follows that $\log q(\boldsymbol{\alpha}) = \log \psi(\mathbf{z}) - \log |\mathbf{M}|$ which immediately implies $\partial \log q(\boldsymbol{\alpha})/\partial\mathbf{M} = -\partial\log |\mathbf{M}|/\partial\mathbf{M} = -(\mathbf{M}^{-1})^\top$. Note that $\partial\log \psi(\mathbf{z})/\partial\mathbf{M} = 0$ since the above derivatives are evaluated with respect to a sampled value of $\mathbf{z}$, which effectively makes $\log\psi(\mathbf{z})$ a constant that does not depend on either $\mathbf{M}$ or $\mathbf{b}$. 

Also, since $p(\boldsymbol{\alpha}) = \mathcal{N}(\boldsymbol{\alpha} | \mathbf{0}, \mathrm{blkdiag}[{\boldsymbol{\Theta}},\boldsymbol{\Lambda}])$ (Section~\ref{parameterisation}), $\partial\log p(\boldsymbol{\alpha})/\partial\boldsymbol{\alpha} = -\mathrm{blkdiag}[{\boldsymbol{\Theta}}^{-1},\boldsymbol{\Lambda}^{-1}]\boldsymbol{\alpha}$.
Let $\boldsymbol{\alpha}\triangleq[\alpha_k]^{\top}_k$, $\mathbf{M}\triangleq[M_{ij}]_{i,j}$, $\mathbf{z}\triangleq[z_j]^{\top}_j$, and $\mathbf{b}=[b_k]^{\top}_k$.
Then, by applying the chain rule of derivatives, it follows that $\partial\log p(\boldsymbol{\alpha})/\partial\mathbf{M} = \sum_{k} (\partial\alpha_k/\partial\mathbf{M})(\partial\log p(\boldsymbol{\alpha})/\partial\alpha_k)$ where $\partial\log p(\boldsymbol{\alpha})/\partial\alpha_k$ corresponds to the $k$-th component of the gradient vector $\partial\log p(\boldsymbol{\alpha})/\partial\boldsymbol{\alpha}$ and $\partial\alpha_k/\partial\mathbf{M} \triangleq [\partial\alpha_k/\partial M_{ij}]_{i,j}$. 
Since $\boldsymbol{\alpha} = \mathbf{M}\mathbf{z} + \mathbf{b}$ (Section~\ref{parameterisation}), $\alpha_k = \sum_j M_{kj}z_j + b_k$ which implies $\partial\alpha_k/\partial M_{ij} = \mathbb{I}(k=i)z_j$. 

Putting all of the above together therefore yields the following analytic expression for $\partial\log (q(\boldsymbol{\alpha})/p(\boldsymbol{\alpha}))/\partial\mathbf{M}$:
$$
\frac{\partial}{\partial\mathbf{M}}\log\frac{q(\boldsymbol{\alpha})}{p(\boldsymbol{\alpha})} = -\left(\mathbf{M}^{-1}\right)^\top - \sum_k \frac{\partial\alpha_k}{\partial\mathbf{M}}\frac{\partial\log p(\boldsymbol{\alpha})}{\partial\alpha_k}
$$
where $\partial\log p(\boldsymbol{\alpha})/\partial\boldsymbol{\alpha} = -\mathrm{blkdiag}[{\boldsymbol{\Theta}}^{-1},\boldsymbol{\Lambda}^{-1}]\boldsymbol{\alpha}$ and $\partial\alpha_k/\partial\mathbf{M} = [\mathbb{I}(k = i)z_j]_{i,j}$ are derived previously. So, $\partial\log (q(\boldsymbol{\alpha})/p(\boldsymbol{\alpha}))/\partial\mathbf{M}$ is analytically tractable. 

Likewise, to derive the derivative $\partial\log (q(\boldsymbol{\alpha})/p(\boldsymbol{\alpha}))/\partial\mathbf{b}$, note that $\partial\log q(\boldsymbol{\alpha})/\partial\mathbf{b} = 0$ as $\log q(\boldsymbol{\alpha}) = \log\psi(\mathbf{z})-\log |\mathbf{M}|$ does not depend on $\mathbf{b}$ since $\mathbf{z}$ is a sampled value that makes $\log\psi(\mathbf{z})$ a constant. 
Also, $\partial\log p(\boldsymbol{\alpha})/\partial\mathbf{b} = (\partial\boldsymbol{\alpha}/\partial\mathbf{b})(\partial\log p(\boldsymbol{\alpha})/\partial\boldsymbol{\alpha}) = \partial\log p(\boldsymbol{\alpha})/\partial\boldsymbol{\alpha}$ since $\partial\boldsymbol{\alpha}/\partial\mathbf{b} = \mathbf{I}$ which is implied by the fact that $\boldsymbol{\alpha} = \mathbf{M}\mathbf{z} + \mathbf{b}$ (Section~\ref{parameterisation}). Hence, $\partial\log (q(\boldsymbol{\alpha})/p(\boldsymbol{\alpha}))/\partial\mathbf{b} = -\partial\log p(\boldsymbol{\alpha})/\partial\boldsymbol{\alpha}$ is analytically tractable, as shown previously. 
\section{Proof of Lemma~\ref{lem:l3}}
\label{app:e}
Using a derivation similar to that in Appendix~\ref{app:a},
$$
\begin{array}{rcl}
p(\mathbf{y} | {\boldsymbol{\theta}}) &\hspace{-2.4mm}=&\hspace{-2.4mm}\displaystyle \mathcal{N}\hspace{-0.7mm}\left(\mathbf{y} | \mathbf{0}, \boldsymbol{\Phi}_{\boldsymbol{\theta}}^\top(\mathbf{X})\boldsymbol{\Lambda}\boldsymbol{\Phi}_{\boldsymbol{\theta}}(\mathbf{X}) + \sigma_n^2\mathbf{I}\right) \\
&\hspace{-2.4mm}=&\hspace{-2.4mm} \displaystyle\int_\mathbf{s} \mathcal{N}\hspace{-0.7mm}\left(\mathbf{y} | \boldsymbol{\Phi}_{\boldsymbol{\theta}}^\top(\mathbf{X})\mathbf{s}, \sigma_n^2\mathbf{I}\right) \mathcal{N}\hspace{-0.7mm}\left(\mathbf{s} | \mathbf{0}, \boldsymbol{\Lambda}\right)\mathrm{d}\mathbf{s} 
%\label{e1}
\end{array}
$$
where the last equality follows from the Gaussian identity of affine transformation for marginalization. Also, from~\eqref{eq:imagine}, $p(\mathbf{s}) = \mathcal{N}(\mathbf{s} | \mathbf{0}, \boldsymbol{\Lambda})$ and hence
\begin{equation}
p(\mathbf{y} | {\boldsymbol{\theta}}) = \int_\mathbf{s} \mathcal{N}\hspace{-0.7mm}\left(\mathbf{y}  | \boldsymbol{\Phi}_{\boldsymbol{\theta}}^\top(\mathbf{X})\mathbf{s}, \sigma^2_n\mathbf{I}\right) p(\mathbf{s})\ \mathrm{d}\mathbf{s} \ .
\label{e2}
\end{equation}
On the other hand, $p(\mathbf{y}|{\boldsymbol{\theta}})$ can also be expressed in terms of $p(\mathbf{y} |{\boldsymbol{\theta}},\mathbf{s})$ and $p(\mathbf{s})$ using marginalization:
\begin{equation}
\begin{array}{rcl}
p(\mathbf{y} | {\boldsymbol{\theta}}) &\hspace{-2.4mm}=&\hspace{-2.4mm} \displaystyle\int_\mathbf{s} p(\mathbf{y}, \mathbf{s} |{\boldsymbol{\theta}})\mathrm{d}\mathbf{s} \ \ =\ \int_\mathbf{s} p(\mathbf{y} | {\boldsymbol{\theta}}, \mathbf{s}) p(\mathbf{s} |{\boldsymbol{\theta}})\mathrm{d}\mathbf{s} \vspace{0.5mm}\\
&\hspace{-2.4mm}=&\hspace{-2.4mm}  \displaystyle\int_\mathbf{s} p(\mathbf{y} |{\boldsymbol{\theta}},\mathbf{s}) p(\mathbf{s})\mathrm{d}\mathbf{s} 
\end{array}
\label{e3}
\end{equation}
where the last equality of~\eqref{e3} follows from our setting in Section~\ref{hgp} and~\eqref{eq:imagine} that ${\boldsymbol{\theta}}$ and $\mathbf{s}$ are statistically independent \emph{a priori}. Subtracting both sides of~\eqref{e2} from that of~\eqref{e3} consequently yields
$$
0 = \int_\mathbf{s} \left(p(\mathbf{y} |{\boldsymbol{\theta}},\mathbf{s}) - \mathcal{N}\hspace{-0.7mm}\left(\mathbf{y} | \boldsymbol{\Phi}^\top_{\boldsymbol{\theta}}(\mathbf{X})\mathbf{s}, \sigma^2_n\mathbf{I}\right)\right) p(\mathbf{s})\ \mathrm{d}\mathbf{s} \ ,
$$
which directly implies $p(\mathbf{y} |{\boldsymbol{\theta}},\mathbf{s}) = \mathcal{N}(\mathbf{y} | \boldsymbol{\Phi}_{\boldsymbol{\theta}}^\top(\mathbf{X})\mathbf{s}, \sigma^2_n\mathbf{I})$ since $p(\mathbf{s}) > 0$ for all $\mathbf{s}$. Then, since $\boldsymbol{\alpha} \triangleq \mathrm{vec}({\boldsymbol{\theta}},\mathbf{s})$ (Section~\ref{parameterisation}), this result can be rewritten more concisely as $p(\mathbf{y}|\boldsymbol{\alpha}) = \mathcal{N}(\mathbf{y} |\boldsymbol{\Phi}_{\boldsymbol{\theta}}^\top(\mathbf{X})\mathbf{s}, \sigma^2_n\mathbf{I})$. Finally, taking the logarithm on both sides of this equation completes our proof of Lemma~\ref{lem:l3}.
\section{Proof of Theorem~\ref{theo:t1}}
\label{app:f}
Note that $\mathbf{v}^\top\mathbf{v} = \sum_{i=1}^p \mathbf{v}_i^\top\mathbf{v}_i$.
% where $\mathbf{v}_i \triangleq \mathbf{y}_i - \boldsymbol{\Phi}_{\boldsymbol{\theta}}^\top(\mathbf{X}_i)\mathbf{s}$. 
Plugging this into Lemma~\ref{lem:l3} yields
\begin{equation}
\log p(\mathbf{y}|\boldsymbol{\alpha}) = -0.5 \sigma_n^{-2} \sum_{i=1}^p\mathbf{v}_i^\top\mathbf{v}_i - 0.5n\log(2\pi\sigma_n^2)\ .
\label{f1}
\end{equation}
%where $\mathcal{C} \triangleq -(n/2)\log(2\pi\sigma_n^2)$. 
Taking derivatives with respect to $\boldsymbol{\eta}$ on both sides of~\eqref{f1} gives 
$$
\begin{array}{rcl}
\displaystyle\frac{\partial}{\partial\boldsymbol{\eta}}
\log p(\mathbf{y}|\boldsymbol{\alpha}) &\hspace{-2.4mm}=&\hspace{-2.4mm}\displaystyle -0.5  \sigma_n^{-2}\sum_{i=1}^p \nabla_{\boldsymbol{\eta}} (\mathbf{v}_i^\top\mathbf{v}_i)\\
&\hspace{-2.4mm}=&\hspace{-2.4mm}\displaystyle -0.5\sum_{i=1}^p \mathbf{r}_i = \sum_{i=1}^p {F}_i(\boldsymbol{\eta,\alpha})
\end{array}
$$
where the last two equalities follow from the definitions of $\mathbf{r}_i$ and ${F}_i(\boldsymbol{\eta,\alpha})$ in Theorem~\ref{theo:t1}. This completes our proof.

\subsubsection{Closed-Form Evaluation of ${F}_i(\boldsymbol{\eta},\boldsymbol{\alpha})$.} 
To show that ${F}_i(\boldsymbol{\eta},\boldsymbol{\alpha})$ and hence $\partial\log p(\mathbf{y} |\boldsymbol{\alpha})/\partial\boldsymbol{\eta}$ can be analytically evaluated, it suffices to show that  $\nabla_{\boldsymbol{\eta}}(\mathbf{v}_i^\top\mathbf{v}_i)$ is analytically tractable.

To understand this, note that $\partial\mathbf{v}_i/\partial\mathbf{s}$ and $\partial\mathbf{v}_i/\partial{\boldsymbol{\theta}}$ are both analytically tractable since $\mathbf{v}_i$ is linear in $\mathbf{s}$ while the trigonometric basis functions constituting $\boldsymbol{\Phi}_{\boldsymbol{\theta}}(\mathbf{X}_i)$ are analytically differentiable with respect to ${\boldsymbol{\theta}}$. 
This therefore implies $\partial\mathbf{v}_i/\partial\boldsymbol{\alpha}$ is also analytically tractable since $\boldsymbol{\alpha} = \mathrm{vec}({\boldsymbol{\theta}},\mathbf{s})$. 

Recall from Appendix~\ref{app:d} that $\boldsymbol{\alpha}\triangleq[\alpha_k]^{\top}_k$.
Then, by applying the chain rule of derivatives,
% for $\partial\mathbf{v}_i/\partial u$ for $u \in \boldsymbol{\eta}$ (i.e., $u = \mathbf{M}_{ij}$ or $u = \mathbf{b}_i$ for some indices $i$ and $j$) gives 
$$
\frac{\partial \mathbf{v}_i}{\partial \mathbf{M}} = \sum_{k}  \frac{\partial \alpha_k}{\partial \mathbf{M}}  \frac{\partial \mathbf{v}_i}{\partial \alpha_k}\ ,\quad\frac{\partial \mathbf{v}_i}{\partial \mathbf{b}} = \frac{\partial \boldsymbol{\alpha}}{\partial \mathbf{b}} \frac{\partial \mathbf{v}_i}{\partial \boldsymbol{\alpha}}
$$
where $\partial \alpha_k/\partial \mathbf{M}$ and $\partial \boldsymbol{\alpha}/\partial \mathbf{b} = \mathbf{I}$ have previously been derived in Appendix~\ref{app:d}.
%$\alpha_t \in \boldsymbol{\alpha}$ and $\partial \alpha_t/\partial u$ . 
Thus, $\partial\mathbf{v}_i/\partial\boldsymbol{\eta}$ is analytically tractable, which shows that ${F}_i(\boldsymbol{\eta,\alpha})$ is indeed analytically tractable.

\section{Generalized Stochastic Gradient}
\label{app:g}
Let $\mathcal{S} \triangleq \{\mathcal{I}, \mathcal{Z}\}$ where $\mathcal{I} \triangleq \{i_k\}_{k=1}^a$ and $\mathcal{Z} \triangleq \{\mathbf{z}_j\}_{j=1}^b$ denote the sets of i.i.d. random samples drawn from $\mathcal{U}(1,p)$ and $\psi(\mathbf{z})$ (Section~\ref{parameterisation}), respectively. Define the stochastic gradient of ${L}(q)$ with respect to $\boldsymbol{\eta} = \mathrm{vec}(\mathbf{M},\mathbf{b})$ as\vspace{-1mm}
\begin{equation}
\frac{\partial\widetilde{{L}}}{\partial\boldsymbol{\eta}} \triangleq \frac{1}{ab}\sum_{k = 1}^a\sum_{j = 1}^b\left(p{F}_{i_k}(\boldsymbol{\eta},\boldsymbol{\alpha}_j) - \frac{\partial}{\partial\boldsymbol{\eta}}\log\frac{q(\boldsymbol{\alpha}_j)}{p(\boldsymbol{\alpha}_j)} \right) 
\label{eqn:a11}
%\vspace{-2mm}
\end{equation}
where $\boldsymbol{\alpha}_j = \mathbf{M}\mathbf{z}_j + \mathbf{b}$.
The result below shows that the generalized stochastic gradient $\partial\widetilde{{L}}/\partial\boldsymbol{\eta}$ \eqref{eqn:a11} is an unbiased estimator of the exact gradient $\partial{L}/\partial\boldsymbol{\eta}$:
%The generalized stochastic gradient can be constructed, as detailed in the result below proving its :
%
\begin{proposition}
\label{theo:t2}
%$\partial\tilde{{L}}/\partial\boldsymbol{\eta}$ is an unbiased and analytically tractable estimate of the exact gradient $\partial{L}/\partial\boldsymbol{\eta}$, that is, 
$\mathbb{E}_{\mathcal{S}}\hspace{-0.7mm}\left[\partial\widetilde{{L}}/\partial\boldsymbol{\eta}\right] = \partial{L}/\partial\boldsymbol{\eta}$. %See Appendix G for a detailed proof. \vspace{1mm}
\end{proposition}
{\bf Proof} Since $\mathcal{I} = \{i_k\}_{k=1}^a$ and $\mathcal{Z} = \{\mathbf{z}_j\}_{j=1}^b$ are sampled independently, 
\begin{equation}
\mathbb{E}_\mathcal{S}\hspace{-0.7mm}\left[\frac{\partial\widetilde{{L}}}{\partial\boldsymbol{\eta}}\right] \triangleq \mathbb{E}_{\mathcal{K},\mathcal{Z}}\hspace{-0.7mm}\left[\frac{\partial\widetilde{{L}}}{\partial\boldsymbol{\eta}}\right] = \mathbb{E}_\mathcal{Z}\hspace{-0.7mm}\left[\mathbb{E}_\mathcal{K}\hspace{-0.7mm}\left[\frac{\partial\widetilde{{L}}}{\partial\boldsymbol{\eta}}\right]\right] . 
\label{g1}
\end{equation}
Also, since $i_1,\ldots,i_a$ are sampled independently from the same uniform distribution $\mathcal{U}(1,p)$ over a discrete set of partition indices $\{1, 2,\ldots,p\}$, 
\begin{equation}
\hspace{-1.7mm}
\begin{array}{l}
\displaystyle\mathbb{E}_\mathcal{K}\hspace{-0.7mm}\left[\frac{\partial\widetilde{{L}}}{\partial\boldsymbol{\eta}}\right] \\
=\displaystyle\frac{1}{ab}\sum_{k=1}^a\sum_{j=1}^b\left(p\mathbb{E}_{i_k\sim\mathcal{U}(1,p)}[{F}_{i_k}(\boldsymbol{\eta},\boldsymbol{\alpha}_j)] - \frac{\partial}{\partial\boldsymbol{\eta}}\log\frac{q(\boldsymbol{\alpha}_j)}{p(\boldsymbol{\alpha}_j)}\right) \\
= \displaystyle\frac{1}{ab}\sum_{k=1}^a\sum_{j=1}^b\left(p\mathbb{E}_{i\sim\mathcal{U}(1,p)}[{F}_i(\boldsymbol{\eta}, \boldsymbol{\alpha}_j)] - \frac{\partial}{\partial\boldsymbol{\eta}}\log\frac{q(\boldsymbol{\alpha}_j)}{p(\boldsymbol{\alpha}_j)}\right) \\
= \displaystyle\frac{1}{b}\sum_{j=1}^b\left(p\mathbb{E}_{i\sim\mathcal{U}(1,p)}[{F}_i(\boldsymbol{\eta}, \boldsymbol{\alpha}_j)] - \frac{\partial}{\partial\boldsymbol{\eta}}\log\frac{q(\boldsymbol{\alpha}_j)}{p(\boldsymbol{\alpha}_j)}\right).
\end{array}
\label{g2}
\end{equation}
%where $\boldsymbol{\alpha}_j = \mathbf{M}\mathbf{z}_j + \mathbf{b}$. 
To simplify~\eqref{g2}, $\mathbb{E}_{i\sim\mathcal{U}(1,p)}[{F}_i(\boldsymbol{\eta},\boldsymbol{\alpha}_j)]$ can be re-expressed as
\begin{equation}
%\hspace{-1.7mm}
\begin{array}{rcl}
\displaystyle\mathbb{E}_{i\sim\mathcal{U}(1,p)}[{F}_i(\boldsymbol{\eta},\boldsymbol{\alpha}_j)] &\hspace{-2.4mm}=&\displaystyle\hspace{-2.4mm} \sum_{i=1}^p \mathcal{U}(1,p) {F}_i(\boldsymbol{\eta},\boldsymbol{\alpha}_j)\vspace{1mm} \\
&\hspace{-2.4mm}=&\hspace{-2.4mm} \displaystyle\frac{1}{p}\sum_{i=1}^p{F}_i(\boldsymbol{\eta}, \boldsymbol{\alpha}_j) \vspace{1mm}\\
&\hspace{-2.4mm}=&\hspace{-2.4mm} \displaystyle\frac{1}{p}\frac{\partial}{\partial\boldsymbol{\eta}}\log p(\mathbf{y}|\boldsymbol{\alpha}_j)
\end{array}
\label{g3}
\end{equation}
where the last equality follows directly from Theorem~\ref{theo:t1}. Then, plugging~\eqref{g3} into~\eqref{g2}, 
\begin{equation}
%\begin{array}{rcl}
\displaystyle\mathbb{E}_\mathcal{K}\hspace{-0.7mm}\left[\frac{\partial\widetilde{{L}}}{\partial\boldsymbol{\eta}}\right] =\displaystyle \frac{1}{b}\sum_{j=1}^b\left(\frac{\partial}{\partial\boldsymbol{\eta}}\log p(\mathbf{y}|\boldsymbol{\alpha}_j) - \frac{\partial}{\partial\boldsymbol{\eta}}\log\frac{q(\boldsymbol{\alpha}_j)}{p(\boldsymbol{\alpha}_j)}\right).%\\
%&\hspace{-2.4mm}=&\hspace{-2.4mm}\displaystyle \frac{1}{b}\sum_{j=1}^b\left(h(\boldsymbol{\alpha}_j) - \frac{\partial}{\partial\boldsymbol{\eta}}\log\frac{q(\boldsymbol{\alpha}_j)}{p(\boldsymbol{\alpha}_j)}\right)
%\end{array}
\label{g4}
\end{equation}
%where $h(\boldsymbol{\alpha}_j) \triangleq \partial\log p(\mathbf{y}|\boldsymbol{\alpha}_j)/\partial\boldsymbol{\eta}$. 
Hence, taking expectation over $\mathcal{Z}$ on both sides of~\eqref{g4} yields 
\begin{equation}
%\hspace{-1.7mm}
\begin{array}{l}
\displaystyle\mathbb{E}_\mathcal{K,Z}\hspace{-0.7mm}\left[\frac{\partial\widetilde{{L}}}{\partial\boldsymbol{\eta}}\right]\vspace{1mm}\\
 =\displaystyle\frac{1}{b}\sum_{j=1}^b\mathbb{E}_{\mathbf{z}_j\sim\psi(\mathbf{z})}\hspace{-0.7mm}\left[\frac{\partial}{\partial\boldsymbol{\eta}}\log p(\mathbf{y}|\boldsymbol{\alpha}_j) - \frac{\partial}{\partial\boldsymbol{\eta}}\log\frac{q(\boldsymbol{\alpha}_j)}{p(\boldsymbol{\alpha}_j)}\right] \\
=\displaystyle \frac{1}{b}\sum_{j=1}^b \frac{\partial{L}}{\partial\boldsymbol{\eta}}\vspace{1mm} \\
=\displaystyle\frac{\partial{L}}{\partial\boldsymbol{\eta}} 
\end{array}
\label{g5}
\end{equation}
where the first equality is due to the fact that $\mathbf{z}_1,\ldots,\mathbf{z}_b$ are sampled independently from the same distribution $\psi(\mathbf{z})$ while the second equality follows from~\eqref{eqn:a5b}. Finally, plugging~\eqref{g5} into~\eqref{g1} shows that $\partial\widetilde{{L}}/\partial\boldsymbol{\eta}$ is an unbiased estimator of $\partial{L}/\partial\boldsymbol{\eta}$, thereby completing our proof.\pagebreak
\section{Supplementary Experiments}
\label{supp}
\begin{figure}[h!]
		\begin{tabular}{cc}
			\hspace{-2mm}\includegraphics[width=4cm]{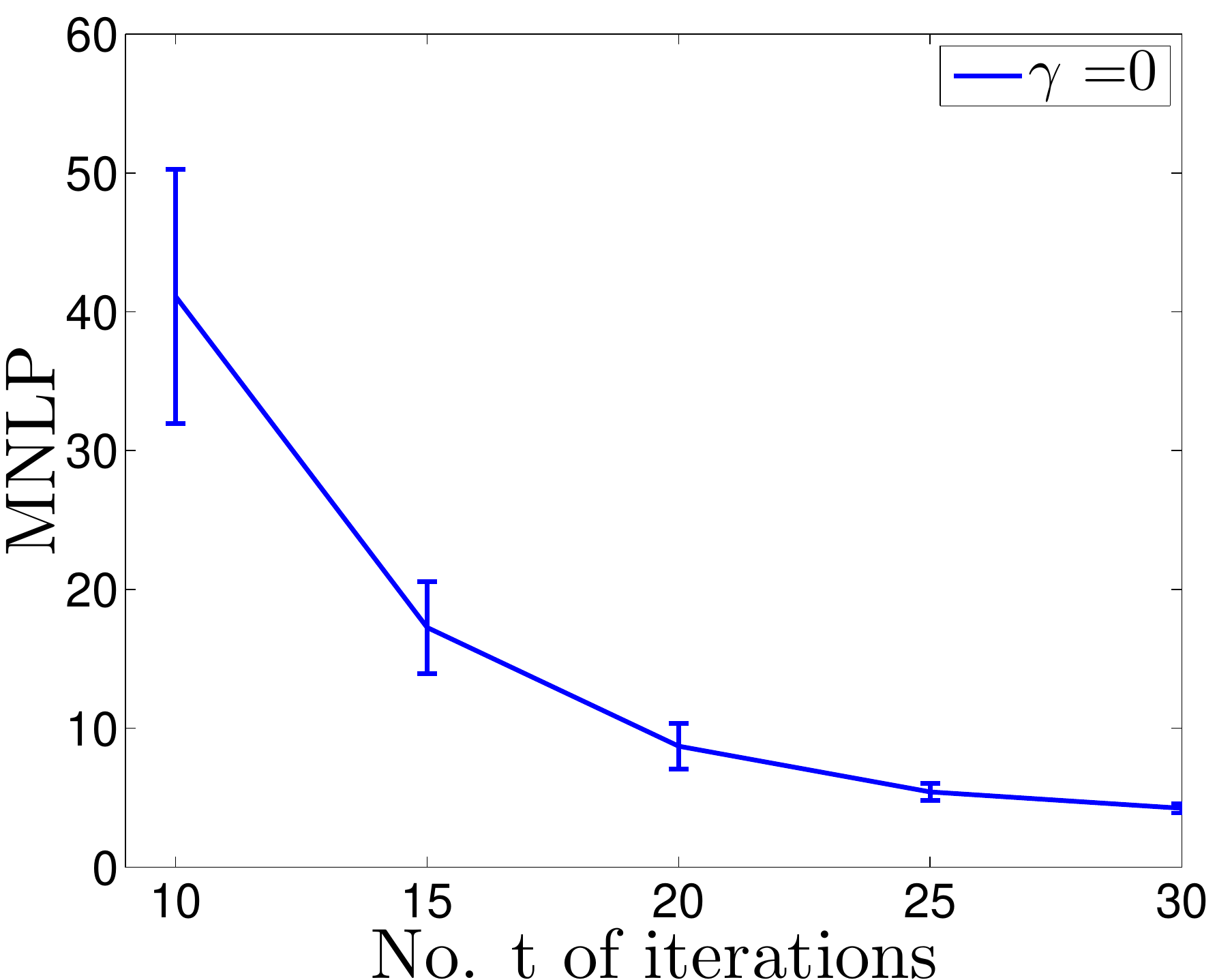} & \hspace{0mm}\includegraphics[width=4cm]{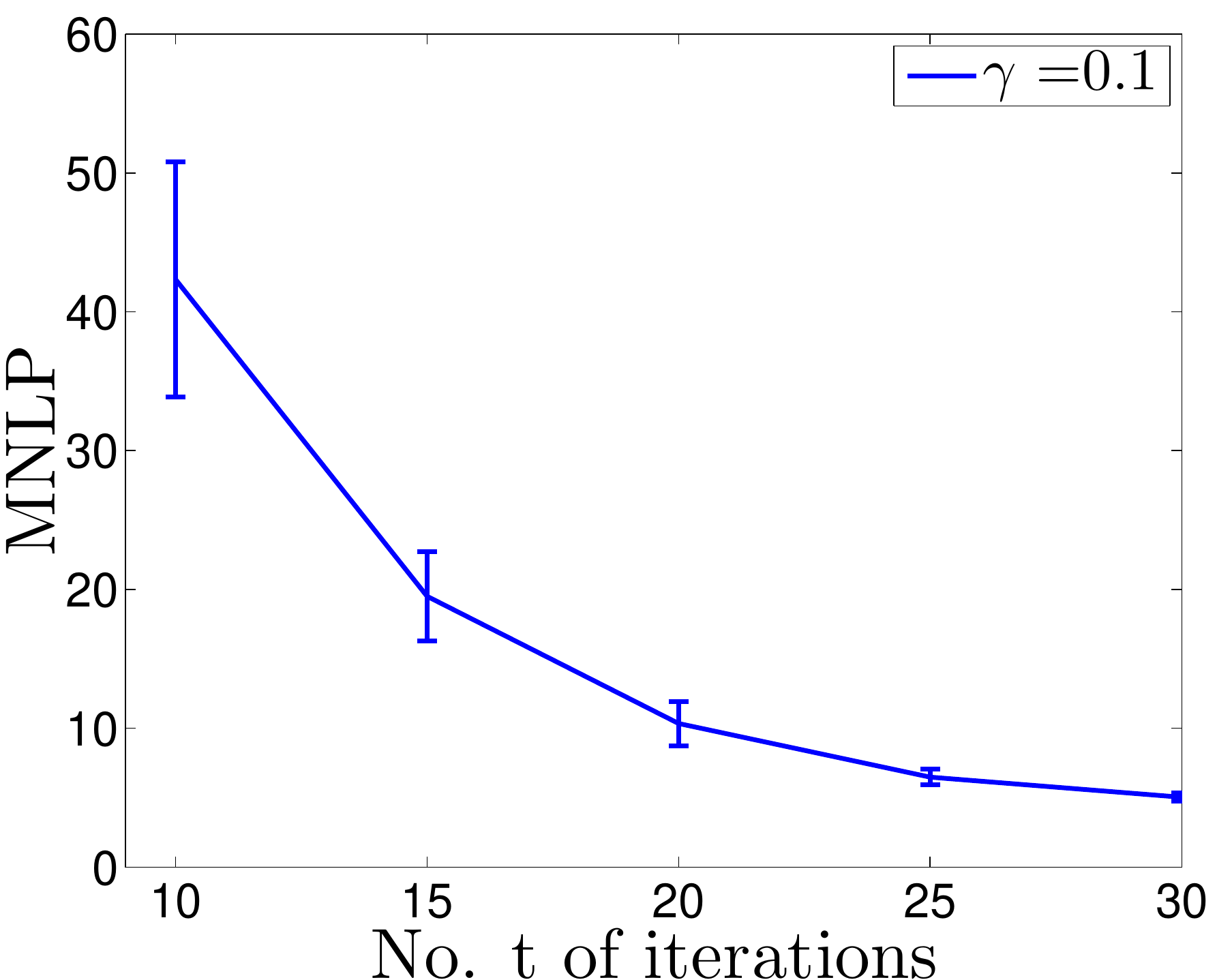}\vspace{-0.5mm}\\
			\hspace{-2mm}(a) & \hspace{0mm}(b) \\
			\hspace{-2mm}\includegraphics[width=4cm]{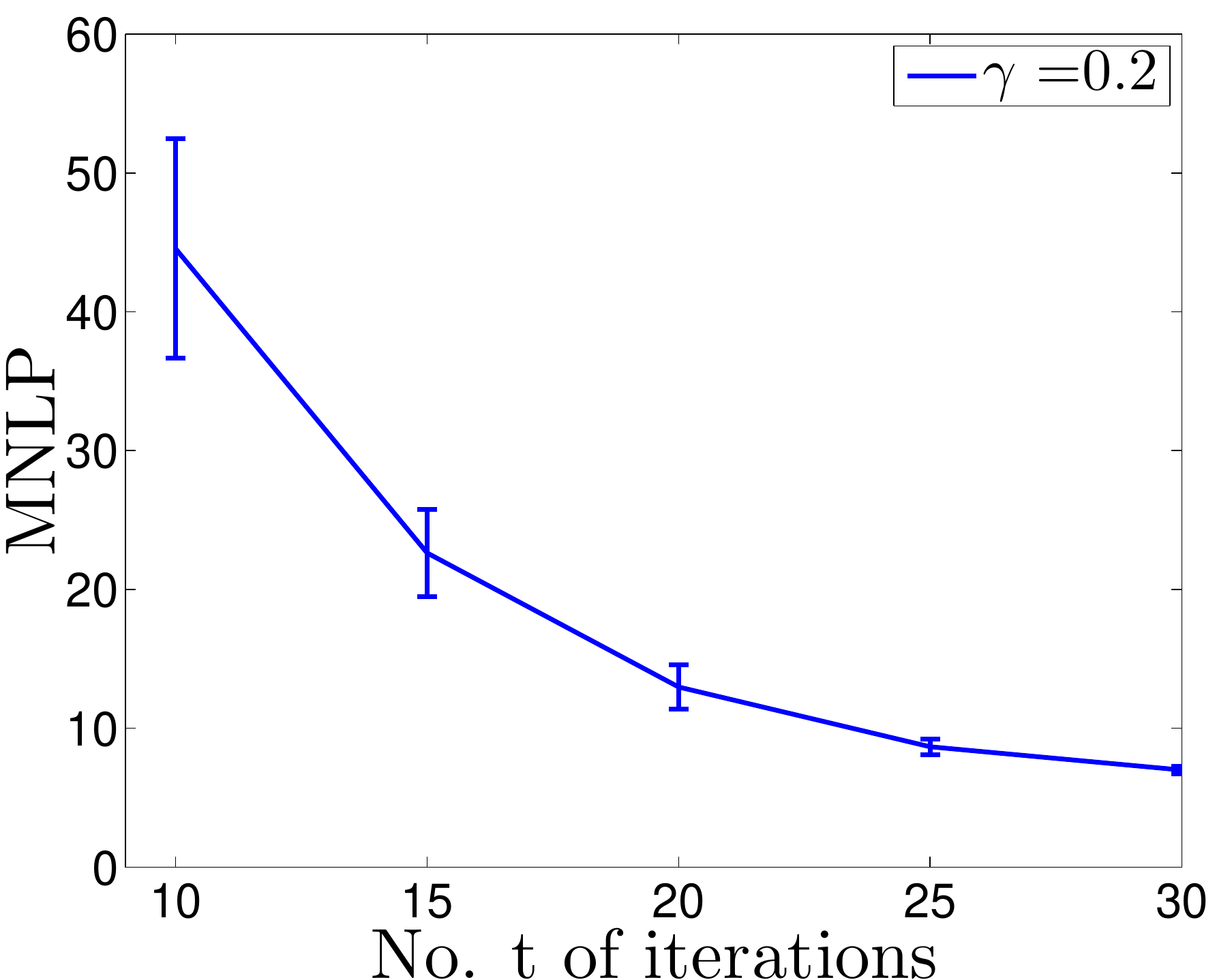} & \hspace{0mm}\includegraphics[width=4cm]{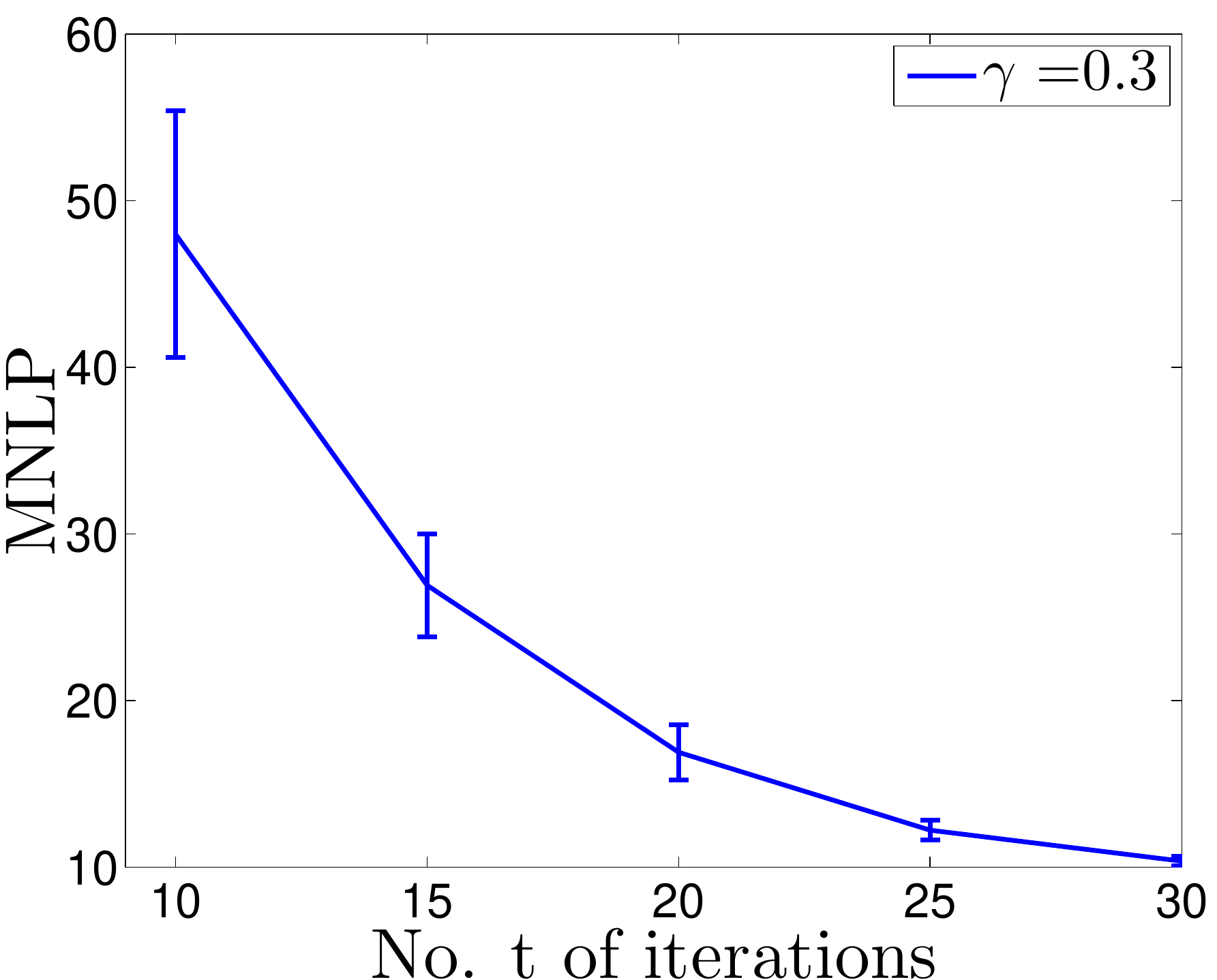}\vspace{-0.5mm}\\
			\hspace{-2mm}(c) & \hspace{0mm}(d)\vspace{-3.5mm}
		\end{tabular}			
	\caption{Graphs of MNLPs (with standard deviations) achieved by $s$VBSSGP vs. number $t$ of iterations with varying values of $\gamma$ for the AIMPEAK dataset.}
\end{figure}

\begin{figure}[h!]
%	\begin{small}
		\begin{tabular}{ccc}
			\hspace{-2mm}\includegraphics[height=2.25cm]{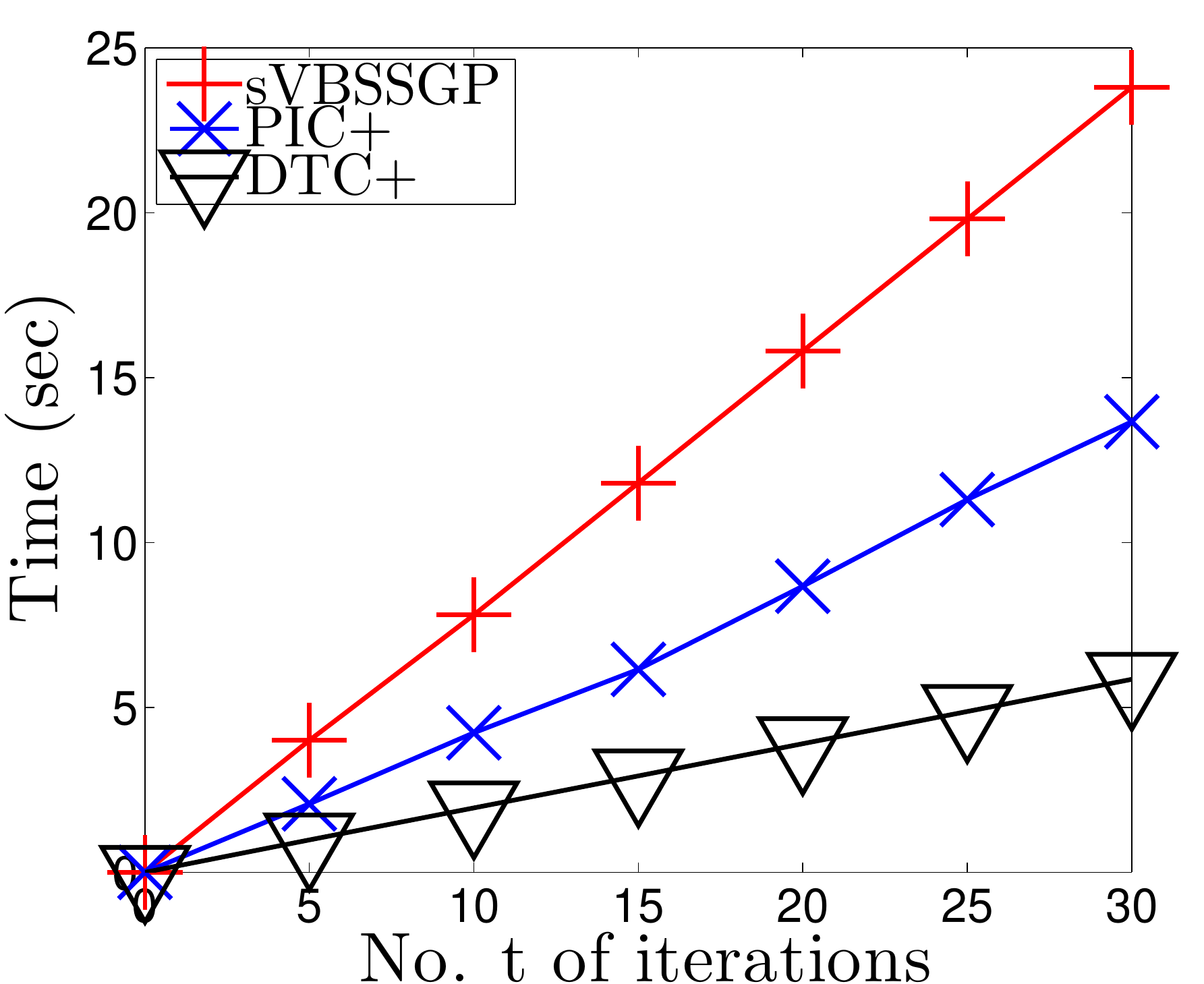} & 
			\hspace{-3.9mm}\includegraphics[height=2.2cm]{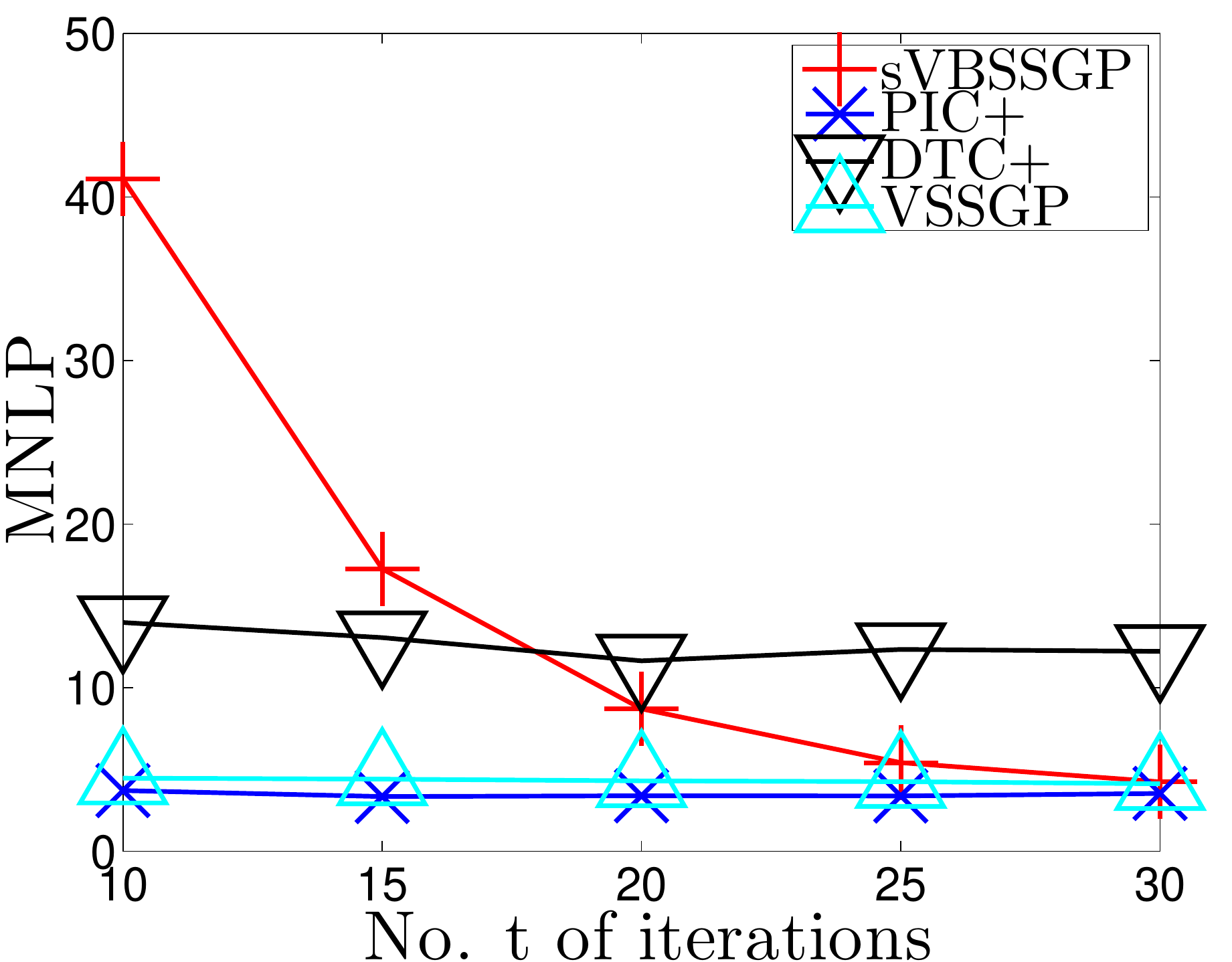} &
			\hspace{-3.9mm}\includegraphics[height=2.25cm]{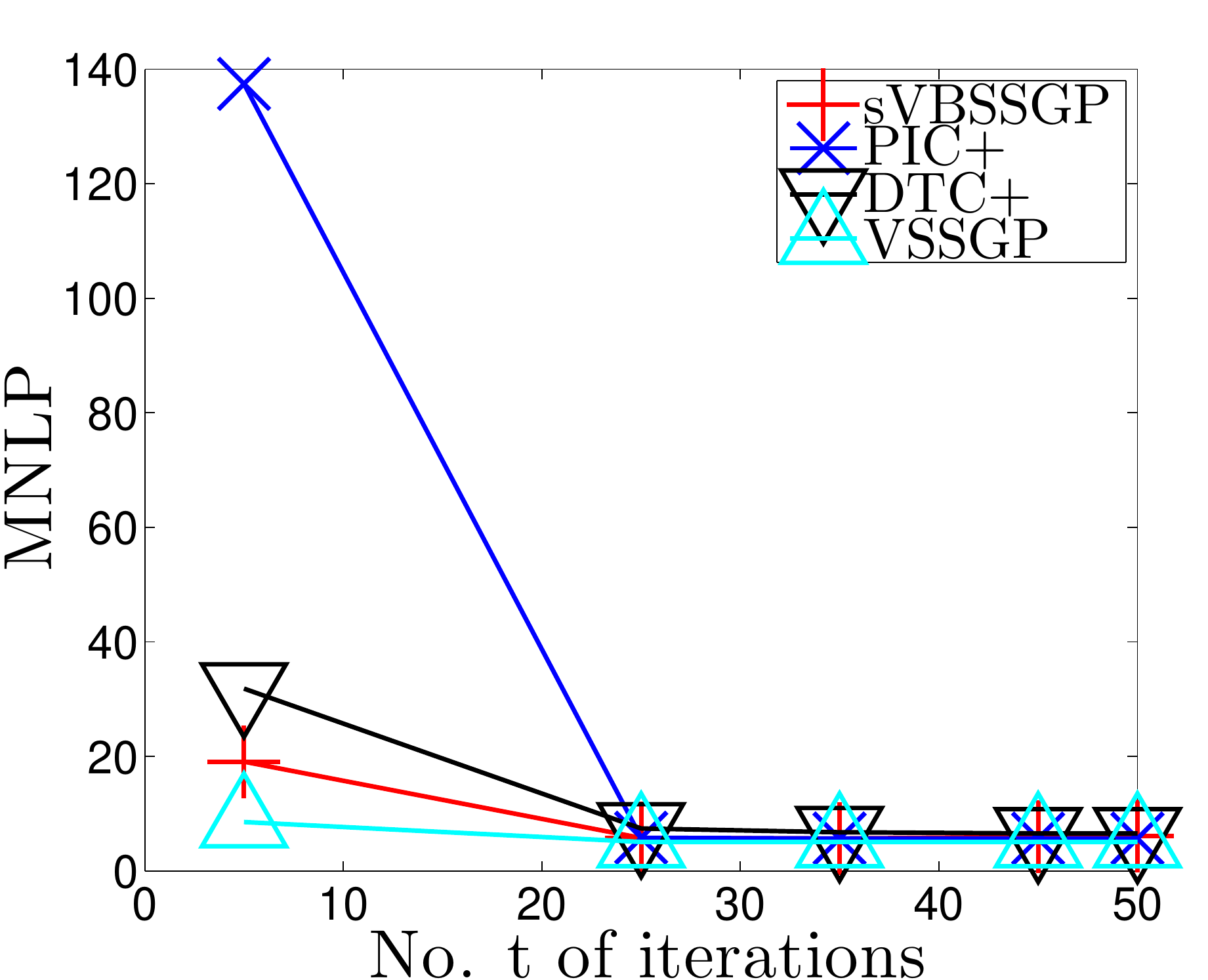} \vspace{-0.5mm}\\
			\hspace{-2mm}(a) & \hspace{-4mm}(b) & \hspace{-4mm}(c)\vspace{-3.5mm}
		\end{tabular}
%	\end{small}
	\caption{(a) Graph of total training time incurred by $s$VBSSGP, PIC$+$, and DTC$+$ vs. number $t$ of iterations for the AIMPEAK dataset; and
%Graphs of MNLP achieved by VBGPR for (a) $\gamma = $ 0.0, (b) $\gamma$ = 0.1, (c) $\gamma$ = 0.2 and (d) $\gamma$ = 0.3  vs. number $t$ of training iterations for the AIMPEAK dataset; 
graphs of MNLPs  achieved by $s$VBSSGP, PIC$+$, DTC$+$, and VSSGP vs. number $t$ of iterations for the (b) AIMPEAK and (c) AIRLINE datasets.}
\end{figure}
\fi

\end{document}